\documentclass[10pt,journal,compsoc]{IEEEtran}

\usepackage{amsmath}
\usepackage{algorithmic}
\usepackage{graphicx}
\usepackage{color}
\usepackage{amssymb}
\usepackage{mathrsfs}
\usepackage{multirow}
\usepackage{array}
\usepackage{url}
\usepackage{wrapfig}
\usepackage{xspace}
\usepackage{booktabs}

\usepackage[pagebackref=true,breaklinks=true,letterpaper=true,colorlinks,linkcolor=blue,citecolor=blue,bookmarks=false]{hyperref}

\makeatletter
\DeclareRobustCommand\onedot{\futurelet\@let@token\@onedot}
\def\@onedot{\ifx\@let@token.\else.\null\fi\xspace}

\def\eg{\emph{e.g}\onedot, } 
\def\ie{\emph{i.e}\onedot, }

\def\etal{\emph{et al}\onedot}
\makeatother

\makeatletter

\newcommand{\Rmnum}[1]{\expandafter\@slowromancap\romannumeral #1@}
\makeatother

\renewcommand{\mathbf}[1]{\boldsymbol{#1}}
\newcommand{\rev}[1]{\textcolor{black}{#1}}

\begin{document}
\title{Exploiting Raw Images for Real-Scene Super-Resolution}
\author{Xiangyu~Xu, Yongrui~Ma, Wenxiu~Sun, Ming-Hsuan~Yang %
\IEEEcompsocitemizethanks{
	\IEEEcompsocthanksitem
	X. Xu is with Robotics Institute, Carnegie Mellon University, Pittsburgh, PA 15213, USA. E-mail: xuxiangyu2014@gmail.com.
	\IEEEcompsocthanksitem
	Y. Ma is with SenseTime Research, Beijing, 100084, China. E-mail: yongrayma@gmail.com.
	\IEEEcompsocthanksitem
	W. Sun is with SenseTime Research, HongKong, 999077. E-mail: irene.wenxiu.sun@gmail.com.
	\IEEEcompsocthanksitem
	M.-H. Yang is with School of Engineering, University of California,	Merced, CA 95343, USA. E-mail: mhyang@ucmerced.edu.
	}%
}

\markboth{IEEE Transactions on Pattern Analysis and Machine Intelligence}{}

\IEEEtitleabstractindextext{%
\begin{abstract}
Super-resolution is a fundamental problem in computer vision which aims to overcome the spatial limitation of camera sensors. 
While significant progress has been made in single image super-resolution,
most algorithms only perform well on synthetic data, which limits their applications in real scenarios.
In this paper, we study the problem of real-scene single image super-resolution to bridge the gap between synthetic data and real captured images.
We focus on two issues of existing super-resolution algorithms: lack of realistic training data and 
insufficient utilization of visual information obtained from cameras.
To address the first issue, we propose a method to generate more realistic training data by mimicking the imaging process of digital cameras.
For the second issue, we develop a two-branch convolutional neural network to exploit the 
radiance information originally-recorded in raw images.
In addition, we propose a dense channel-attention block for better image restoration as well as a learning-based guided filter network for effective color correction.
Our model is able to generalize to different cameras without deliberately training on images from specific camera types.
Extensive experiments demonstrate that the proposed algorithm can recover fine details and clear structures, and  achieve high-quality results for single image super-resolution in real scenes.
\end{abstract}

}

\maketitle

\IEEEdisplaynontitleabstractindextext
\IEEEpeerreviewmaketitle

\IEEEraisesectionheading{
\section{Introduction}\label{sec:introduction}}
\IEEEPARstart{I}{n} digital photography, image resolution, \ie the number of pixels representing an object, is directly proportional to the square of the focal length of a camera~\cite{geo_optics}.
While one can use a long-focus lens to obtain a high-resolution image, the range of the captured scene is usually limited by the size of the sensor array at the imaging plane.
Thus, it is often desirable to capture the wide-range scene at a lower resolution with a short-focus camera (\eg a wide-angle lens), 
and then apply a single image super-resolution algorithm to recover a high-resolution image from the corresponding low-resolution frame.

\begin{figure*}[t]
	\newcommand{\thisheight}{0.16\linewidth}
	\footnotesize
	\begin{center}
		\begin{tabular}{ccccc}
			\hspace{-3mm} 
			\includegraphics[height = \thisheight]{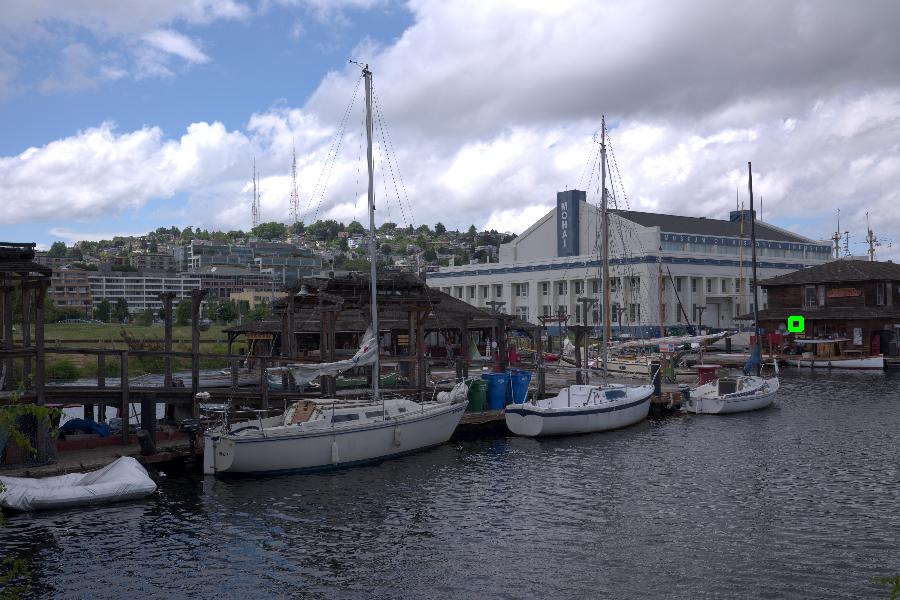} &
			\hspace{-3mm} 
			\includegraphics[height = \thisheight]{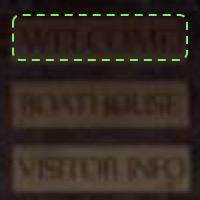} & 
			\hspace{-3mm} 
			\includegraphics[height = \thisheight]{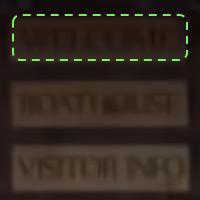} & 
			\hspace{-3mm}
			\includegraphics[height = \thisheight]{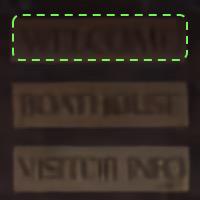} & 
			\hspace{-3mm}
			\includegraphics[height = \thisheight]{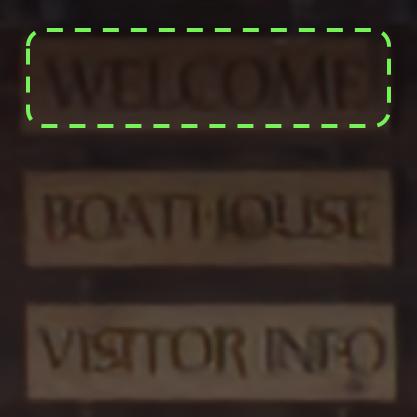} \\
			\hspace{-3mm} 
			\includegraphics[height = \thisheight]{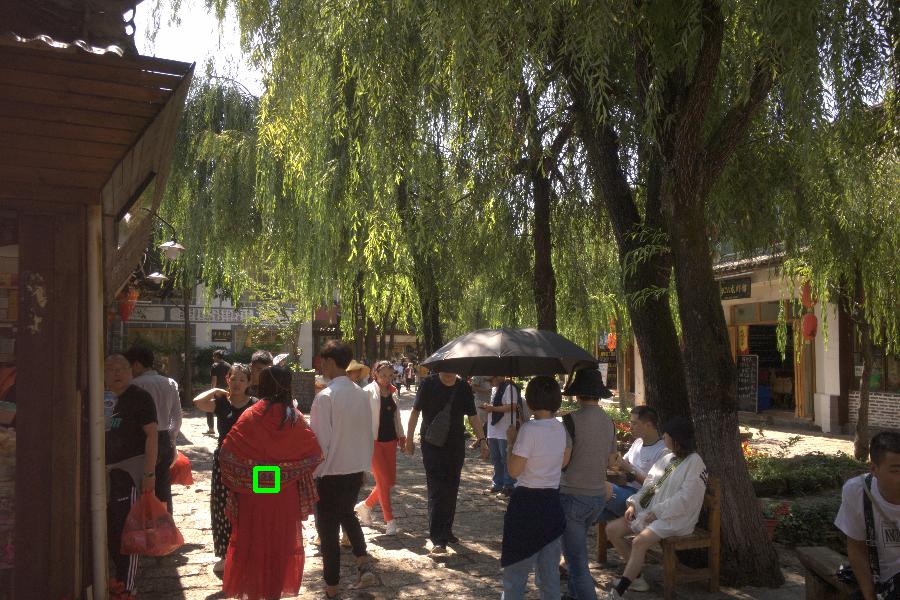} &
			\hspace{-3mm} 
			\includegraphics[height = \thisheight]{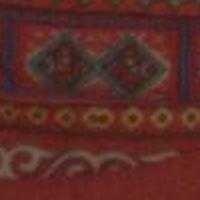} & 
			\hspace{-3mm} 
			\includegraphics[height = \thisheight]{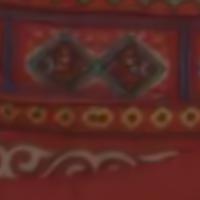} & 
			\hspace{-3mm}
			\includegraphics[height = \thisheight]{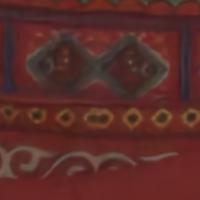} & 
			\hspace{-3mm}
			\includegraphics[height = \thisheight]{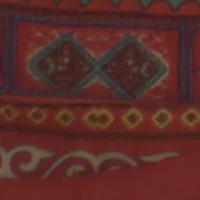} \\ 
			\hspace{-3mm} \scriptsize
			(a) Whole input image &
			\hspace{-2mm} \scriptsize
			(b) Input &
			\hspace{-2mm} \scriptsize
			(c) RDN with previous data &
			\hspace{-2mm} \scriptsize
			(d) RDN with our data &
			\hspace{-2mm} \scriptsize
			(e) Ours \\
		\end{tabular}
	\end{center}
	\vspace{-2mm}
\caption{Existing super-resolution method (\eg RDN~\cite{SR-residual-dense}) does not perform well on real captured images as shown in (c).
		We can obtain sharper results (d) by re-training the existing model~\cite{SR-residual-dense} with more realistic training data generated by our method.
		Furthermore, we recover more structures and details (e) by exploiting the radiance information recorded in raw images.
		The two input images in (a) are captured by Leica SL Typ-601
		and iPhone 6s Plus respectively, and both cameras are not seen by the models during training.
		Best enlarge and view on screen.
	}
	\label{fig:teaser}
	\vspace{-4mm}
\end{figure*}

Most state-of-the-art super-resolution methods~\cite{SRCNN,SR-VDSR,SR-regression-accv14,SR-SCN,SR-laplacian,SR-dense,SR-residual-dense,xxy-iccv17} are based on data-driven models, and in particular, deep convolutional neural networks (CNNs).
While these methods are effective on synthetic data, they do not perform well for images real captured by cameras or cellphones (see examples in Figure~\ref{fig:teaser}(c)) mainly due to two important problems. 
First, existing data generation pipelines are not able to synthesize realistic training data, and models trained with such data do not perform well on most real images. 
Second, the radiance information recorded by the cameras has not been fully utilized.
To address these problems for real-scene super-resolution, we propose a pipeline for generating training data as well as a two-branch CNN model for exploiting the raw information. 
We describe the above problems and our solutions in more detail.

Existing methods usually adopt an over-simplified generation model to synthesize low-resolution images, including fixed downsampling blur kernels (\eg bicubic kernel) and homoscedastic Gaussian noise~\cite{SR-residual-dense,xxy-tip18-edgedeblur}.
However, the blur kernel in practice may vary with zoom, focus, and camera shake during the image formation process. 
As such, the  fixed-kernel assumption usually does not hold in real scenes.
Furthermore,  image noise usually obeys a heteroscedastic Gaussian distribution~\cite{benchmarking_denoising,brooks2019unprocessing} in which variance depends on the pixel intensity.
More importantly, while the blur kernel and noise should be applied to the linear raw data, existing approaches operate on the \rev{nonlinear color images}.
To address these issues, 
we propose a data generation pipeline by simulating the imaging process of digital cameras. 
We synthesize the image degradations in the linear space and apply different downsampling kernels as well as 
heteroscedastic Gaussian noise to approximate real scenarios.
As shown in Figure~\ref{fig:teaser}(d), training an existing model~\cite{SR-residual-dense} using our generated data leads to sharper results. 

\begin{figure}[t]
	\footnotesize
	\begin{center}
		\begin{tabular}{c}
			\includegraphics[width = 0.95\linewidth]{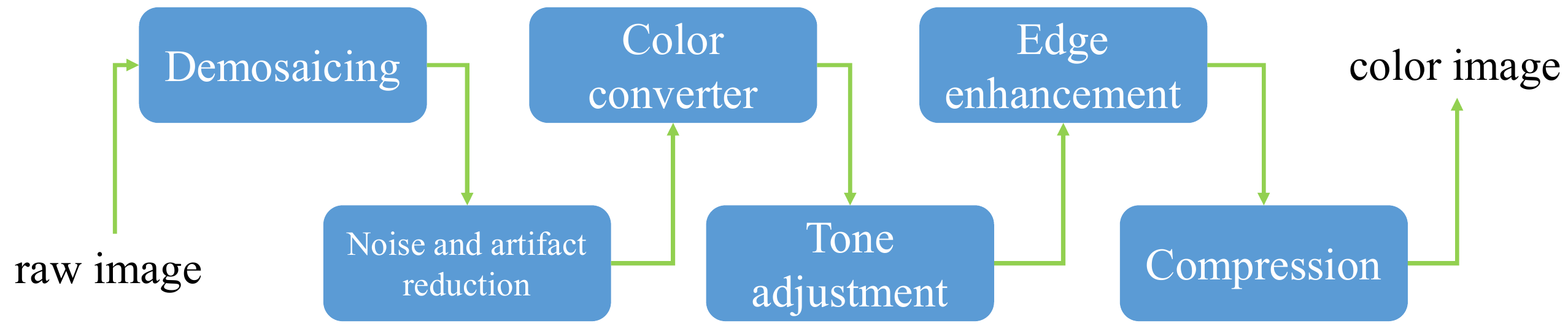} \\
		\end{tabular}
	\end{center}
	\vspace{-2mm}
\caption{A typical ISP pipeline.
	}
	\label{fig:introduction}
	\vspace{-4mm}
\end{figure}

While modern cameras provide users with both the raw data and color image processed by the image signal processing system (ISP)~\cite{hdr+},
most super-resolution algorithms only take the color image as input
without making full use of the radiance information contained in the raw data.
In contrast, we directly use the raw data for restoring high-resolution clear images, which conveys the following advantages.
First, more information can be exploited in raw pixels since they are typically 12 or 14 bits~\cite{hdr+,adobe-mit-5k}, %
whereas the color pixels produced by ISP are typically 8 bits~\cite{hdr+}. 
In addition to the bit depth, there is also information loss within a typical ISP (Figure~\ref{fig:introduction})
such as noise reduction and compression~\cite{pennebaker1992jpeg}.
Second, the
raw data is linearly proportional to scene radiance while the ISP system contains nonlinear operations such as tone mapping.
Thus, the linear degradations in the imaging process, including blur and noise, are linear for raw data but nonlinear in the processed RGB images.
As this nonlinear property makes image restoration more complicated, it is beneficial to address the problem on the raw data~\cite{raw-reconstruction,tai2013nonlinear}.
Third, the demosaicing step in ISP is closely related to image super-resolution because both problems are concerned with the spatial resolution limitations of cameras~\cite{SR_demosaic_deep}. 
Therefore, to solve the super-resolution problem with processed images is sub-optimal and less effective than a single unified model solving both problems simultaneously.

\begin{figure}[t]
	\footnotesize
	\begin{center}
		\begin{tabular}{c}
			\includegraphics[width = \linewidth]{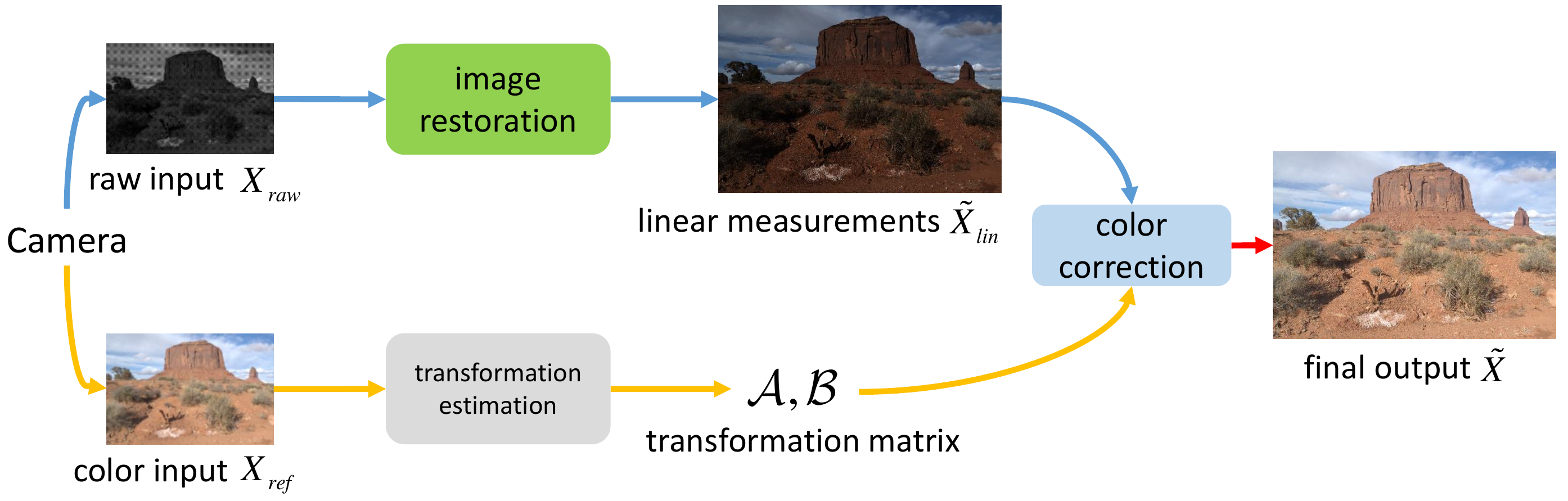} \\
		\end{tabular}
	\end{center}
	\vspace{-2mm}
	\caption{Overview of the proposed network. 
		Our model consists of two modules,
		where one exploits raw data $X_{raw}$ to restore high-resolution linear measurements $\tilde{X}_{lin}$ for all color channels with clear structures and fine details,
		and the other one estimates the transformation matrix to recover the final color result $\tilde{X}$ using the low-resolution color image $X_{ref}$ as reference.
	}
	\label{fig:network_overview}
	\vspace{-4mm}
\end{figure}

Nevertheless, directly mapping a degraded raw image to a desired full color output~\cite{learning_to_see_in_the_dark,deepisp_tip} results in networks only suitable for specific cameras,
as the raw data does not contain the relevant information for color corrections (\eg color space conversion and tone mapping) conducted within the ISP.
To address this issue,
we propose a dual CNN architecture (Figure~\ref{fig:network_overview}) which takes both the degraded raw and color images as input, such that our model can generalize well to different cameras.
The proposed model consists of two branches,
where one module restores clear structures and fine details from the raw data, 
and the other recovers high-fidelity colors using the low-resolution RGB image as a reference.
For the image restoration module, we train a CNN to reconstruct the high-resolution linear color measurements from the raw input.
To increase the representation capacity of the image restoration model, we improve the densely-connected convolution layers~\cite{densenet,rawSR} and propose a dense channel-attention network.
For the color correction module, a straightforward approach is to apply the guided filter~\cite{guidedFilter} to transfer the color of the reference image to the restored linear image. 
However, simply adopting this method usually leads to artifacts and inaccurate color appearances 
due to limitations of  local constant assumptions of the guided filter~\cite{guidedFilter}.
Instead, we present a learning-based guided filter which can better adapt to high-frequency image structures and generate more appealing results. 
As shown in Figure~\ref{fig:teaser}(e), the proposed algorithm significantly improves the super-resolution results for real-scene images.

We make the following contributions in this work.
First, we propose a data generation pipeline which is able to synthesize realistic raw and color images for learning real-scene image super-resolution models.
Second, we develop a two-branch network to exploit both raw data and color images. 
In the first module, we design a dense channel-attention block to improve image restoration.
In the second module, we develop a learning-based guided filter to overcome the limitations of the original guided filter, which helps recover more appealing color appearances.
Finally, we propose a real image dataset to better evaluate different models. 
Extensive experiments demonstrate that the proposed method can recover fine details as well as clear structures, and generate high-quality results in real scenes.

\vspace{-2mm}
\section{Related Work}
We discuss the state-of-the-art super-resolution methods as well as learning-based raw image processing techniques, and put this work in proper context.

\vspace{-2mm}
\subsection{Super-resolution}
Classical super-resolution methods can be broadly categorized into
exemplar-based~\cite{SR-freeman2002,SR-NNembedding,yangSR2010,SR-sparsecoding-wang12}
and regression-based~\cite{SR-regression-iccv13,SR-regression-accv14}.
One typical exemplar-based method is developed based on the Markov model~\cite{SR-freeman2002} to recover high-resolution images from low-resolution inputs. 
As this algorithm directly combines and stitches high-resolution patches from an external dataset, it tends to introduce unrealistic details in the reconstructed results.
Regression-based approaches typically learn the priors from image patches with either linear functions~\cite{SR-regression-iccv13} or anchored neighborhood regressors~\cite{SR-regression-accv14}.
However, the reconstructed images are often over-smoothed due to the limited capacity of the regression models.

With the rapid advances of deep learning, state-of-the-art super-resolution methods based on deep CNNs have been developed to reconstruct high-resolution images~\cite{SR-SCN,SR-VDSR,SR-dense,SRCNN,SR-residual-dense,SR-laplacian,FSRCNN,xxy-iccv17}.
Dong \etal~\cite{SRCNN} propose a three-layer CNN for mapping the low-resolution representation to the high-resolution space, but do not obtain better results with deeper networks~\cite{SRCNN_pami}.
To solve this problem, Kim \etal~\cite{SR-VDSR} introduce residual connections to alleviate the gradient-vanishing issue during the training process, which achieves significantly better results.
Similarly, recursive layers with skip connections are introduced for more efficient network structures~\cite{kim2016deeply}.
As the dense connections~\cite{densenet} are shown to be effective for strengthening feature propagation, they are also used for super-resolution to facilitate the reconstruction process~\cite{SR-dense,SR-residual-dense}.
However, these algorithms only perform well on synthetic data and do not generate high-quality results for real-scene inputs.

Another line of work aims to increase the perceptual  quality of the results~\cite{loss_perceptual,ganSR,zhang2019ranksrgan,xxy-iccv17,sajjadi2017enhancenet}.
Johnson \etal \cite{loss_perceptual} propose a perceptual loss based on high-level features extracted from the VGG network~\cite{VGG}.
Ledig \etal \cite{ganSR} adopt the generative adversarial nets (GAN)~\cite{gan} to learn the distributions of real images.
Sajjadi \etal \cite{sajjadi2017enhancenet} further improve the adversarial loss by enforcing a Gram-matrix matching constraint.
Xu \etal \cite{xxy-iccv17} analyze the GAN loss from the maximum-a-posterior perspective and combine the pixel-wise loss, perceptual loss and GAN loss in an adversarial framework, which super-resolves text and face images well. 
However, the results hallucinated by GAN may contain unpleasant artifacts and details not present in the ground truth.

While the aforementioned methods are effective in interpolating pixels and hallucinating images, they are only based on processed color images and less effective in generating sharp image structures and high-fidelity details for real inputs. 
In contrast, we exploit both the raw and color images in a unified framework for better super-resolution results.
In addition, we propose a pipeline for synthesizing realistic training data to improve the performance.

\vspace{1ex}
\noindent \rev{{\bf{Super-resolution with variable degradations.}} 
While most of the aforementioned methods~\cite{SR-SCN,SR-VDSR,SR-dense,SRCNN,SR-residual-dense,SR-laplacian,FSRCNN,loss_perceptual,ganSR,sajjadi2017enhancenet} formulate the super-resolution problem with a fixed degradation model, some recent appraches~\cite{xxy-iccv17,zhang2018learning,zhang2019deep,zhang2020deep} also consider variable blur kernels and noise levels, which are closer to real scenes.
In particular, Zhang \etal~\cite{zhang2018learning} use the blur kernel and noise level as additional channels of the input and attain favorable results under a multiple-degradation setting.
A deep unfolding network is introduced in \cite{zhang2020deep} to super-resolve blurry and noisy images.
In this work, we also consider variable degradations in the proposed data generation pipeline, which facilitates training more robust models for real applications. 
}

\vspace{1ex}
{\noindent \bf{Joint super-resolution and demosaicing.}} %
Most existing methods for this problem estimate a high-resolution color
image with multiple low-resolution frames~\cite{farsiu2006multiframe,vandewalle2007joint}. 
Closely related to our work is the method based on
a deep residual network~\cite{SR_demosaic_deep} for single image super-resolution with mosaiced input. 
However, this model is trained on gamma-corrected image pairs which may not perform well on real-scene images.
More importantly, these methods do not consider the complex color correction steps in camera ISPs, and thus cannot recover high-fidelity color appearances from the raw input.
In contrast, the proposed algorithm simultaneously addresses the problems of image restoration and color correction to handle real-scene images.

\vspace{-2mm}
\subsection{Learning-based raw image processing}
In recent years, numerous learning-based methods have been proposed for raw image processing~\cite{jiang2017learning,learning_to_see_in_the_dark,deepisp_tip}.
Jiang \etal \cite{jiang2017learning} propose to learn a large collection of local linear filters to approximate the complex nonlinear ISP pipelines.
In \cite{deepisp_tip}, Schwartz \etal use deep CNNs for learning the color correction operations of specific digital cameras.
In addition, Chen~\etal~\cite{learning_to_see_in_the_dark} train a neural network with raw data as input for fast low-light imaging and coloring.
In this work, we learn color correction in the context of raw image super-resolution. Instead of learning a color correction pipeline for one specific camera, we use a low-resolution color image as reference for handling images from more diverse ISPs.

\begin{figure}[t]
	\footnotesize
	\begin{center}
		\begin{tabular}{c}
			\includegraphics[width=0.5\linewidth]{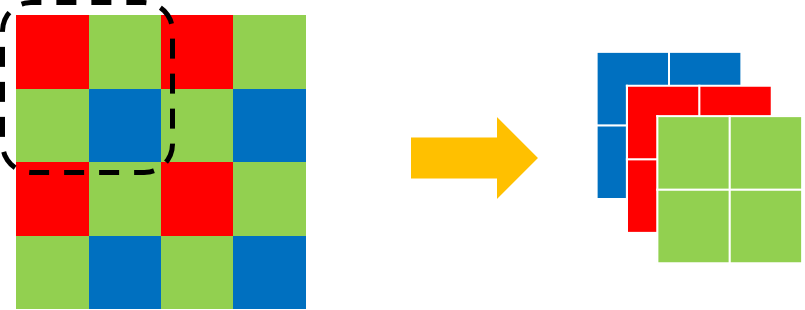}\\
		\end{tabular}
	\end{center}
	\vspace{-2mm}
	\caption{Synthesizing linear color measurements from raw data. The sensels surrounded by the black-dot curve is a Bayer pattern block and defined as a new virtual sensel in Section~\ref{sec:training_data}.}
	\label{fig:bayer pattern}
	\vspace{-4mm}
\end{figure}

\vspace{-2mm}
\section{Proposed Algorithm}
For high-quality single image super-resolution, we propose a pipeline to synthesize more realistic training images and a two-branch CNN model to exploit the radiance information recorded in raw data.

\vspace{-2mm}
\subsection{Data generation pipeline}
\label{sec:training_data}
Most super-resolution methods~\cite{SR-VDSR,SRCNN} generate training data with a fixed downsampling kernel in the nonlinear color space.
In addition, homoscedastic Gaussian noise is often added to the images for modeling the degradation process~\cite{SR-residual-dense,xxy-iccv17}.
However, as discussed in Section~\ref{sec:introduction}, the low-resolution images generated in this manner do not resemble real captured images and are less effective for training real-scene super-resolution models.
Furthermore, we need to generate low-resolution raw data for training the proposed neural network, which is often realized by directly mosaicing the low-resolution color images~\cite{demosaicking_denoising_deep,SR_demosaic_deep}.
This approach ignores the fact that the color images have already been processed by the nonlinear operations of the ISP, and the subsequent synthesized images suffer from information loss and do not represent the linear response of the camera sensors.
Instead, the raw data should be linear measurements of the original pixels.

To address the above issues, we use high-quality raw images~\cite{adobe-mit-5k} and propose a 
data generation pipeline by simulating the imaging process in real scenarios.
We first synthesize ground-truth linear color measurements by downsampling high-quality raw images such that each pixel can have its ground-truth red, green and blue values.
For the Bayer pattern raw data (Figure~\ref{fig:bayer pattern}), we define a block of Bayer pattern sensels (\ie sensor elements) as one new virtual sensel,  where all color measurements are available.
As such, we can obtain the desired images $X_{lin} \in \mathbb{R}^{H\times W \times 3}$ with linear measurements of all three colors for each pixel, where
$H$ and $W$ denote the height and width of the ground-truth linear image.
Similar to~\cite{demosaic_denoise_randomfields}, we compensate the color shift artifact by aligning the center of each color in the new sensel.
\begin{figure}[t]
	\footnotesize
	\begin{center}
		\begin{tabular}{c}
			\includegraphics[width = \linewidth]{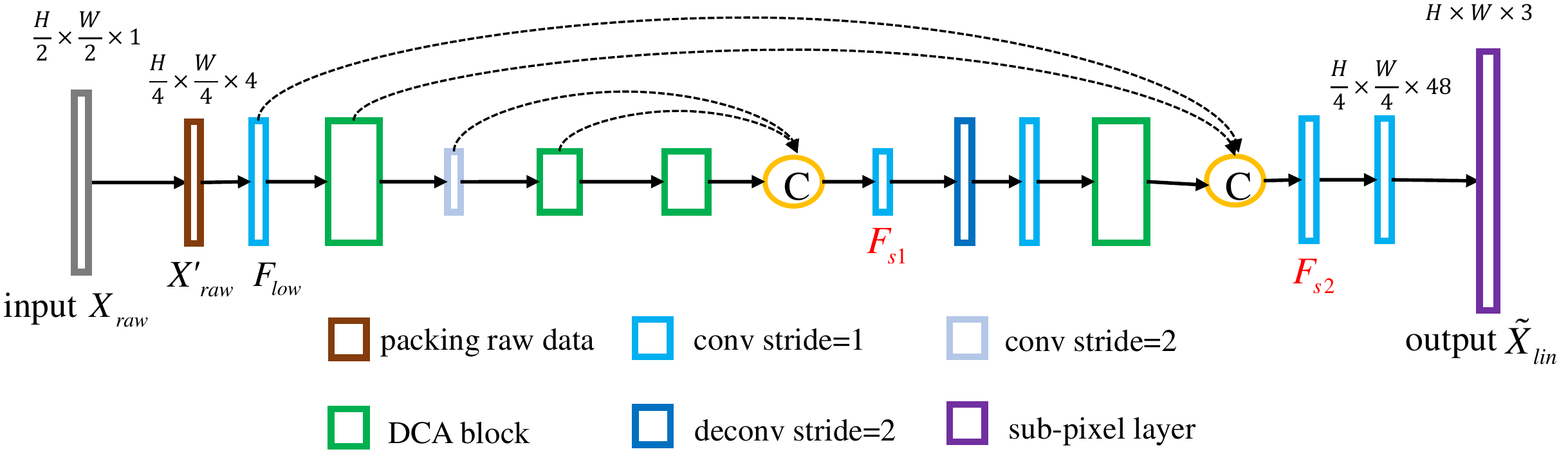} \\
		\end{tabular}
	\end{center}
	\vspace{-2mm}
	\caption{The image restoration module uses the DCA blocks in an encoder-decoder framework
		and reconstructs high-resolution linear color measurements $\tilde{X}_{lin}$ from the degraded low-resolution raw input $X_{raw}$. 
	}
	\label{fig:branch1}
	\vspace{-4mm}
\end{figure}

With the ground-truth linear measurements $X_{lin}$, we can easily generate the ground-truth color images $X_{gt} \in \mathbb{R}^{H\times W \times 3}$ by simulating the color correction steps of the ISP, such as color space conversion and tone adjustment. 
In this work, we apply the widely-used Dcraw~\cite{dcraw} toolkit to process the raw image data. 

To generate degraded low-resolution raw images $X_{raw} \in \mathbb{R}^{\frac{H}{2}\times \frac{W}{2}}$, we apply the blurring, downsampling, Bayer sampling operations to the linear color measurements:
\begin{equation}\label{eq:data_generation}
X_{raw}=f_{Bayer}(f_{down}(X_{lin} * k_{def}*k_{mot}))+n, 
\end{equation}
where $f_{Bayer}$ is the Bayer sampling function which mosaics images in accordance with the Bayer pattern (Figure~\ref{fig:bayer pattern}), and $n$ is the noise term.
In addition, $f_{down}$ represents the downsampling function with a sampling factor of two.
Since the imaging process is likely to be affected by out-of-focus effect and camera shake,
we consider both defocus blur $k_{def}$ modeled as disk kernels with variant sizes~\cite{xxy-iccv17,hradivs2015convolutional}, and modest motion blur $k_{mot}$ generated by random walk~\cite{schuler2013machine,xxy-tip18-edgedeblur}.
In~\eqref{eq:data_generation}, $*$ denotes the convolution operator. 
To synthesize more realistic training data,
we add heteroscedastic Gaussian noise~\cite{KPN,benchmarking_denoising,healey1994radiometric} to the generated raw data:
\begin{equation} \label{eq:noise}
n(x) \sim \mathcal{N}(0, \sigma_1^2 x+\sigma_2^2),
\end{equation}
where the variance of noise $n$ depends on the pixel intensity $x$,
and $\sigma_1$ and $\sigma_2$ are parameters of the noise.
Finally,  the raw image $X_{raw}$ is demosaiced with the AHD~\cite{AHD2005} method, and processed by the Dcraw toolkit to produce the low-resolution color image $X_{ref} \in \mathbb{R}^{\frac{H}{2}\times \frac{W}{2} \times 3}$.  
We compress $X_{ref}$ as 8-bit JPEG as normally done in digital cameras.
Note that the settings for the color correction steps in Dcraw are the same as those used in generating $X_{gt}$, such that the reference colors correspond well to the high-resolution ground-truth.
The proposed pipeline is able to synthesize realistic data for super-resolution, and thereby the models trained on $X_{raw}$ and $X_{ref}$ can effectively generalize to real-degraded raw and color images (Figure~\ref{fig:teaser}(e)).
\begin{figure}[t]
	\footnotesize
	\begin{center}
		\begin{tabular}{c}
			\includegraphics[width = .99\linewidth]{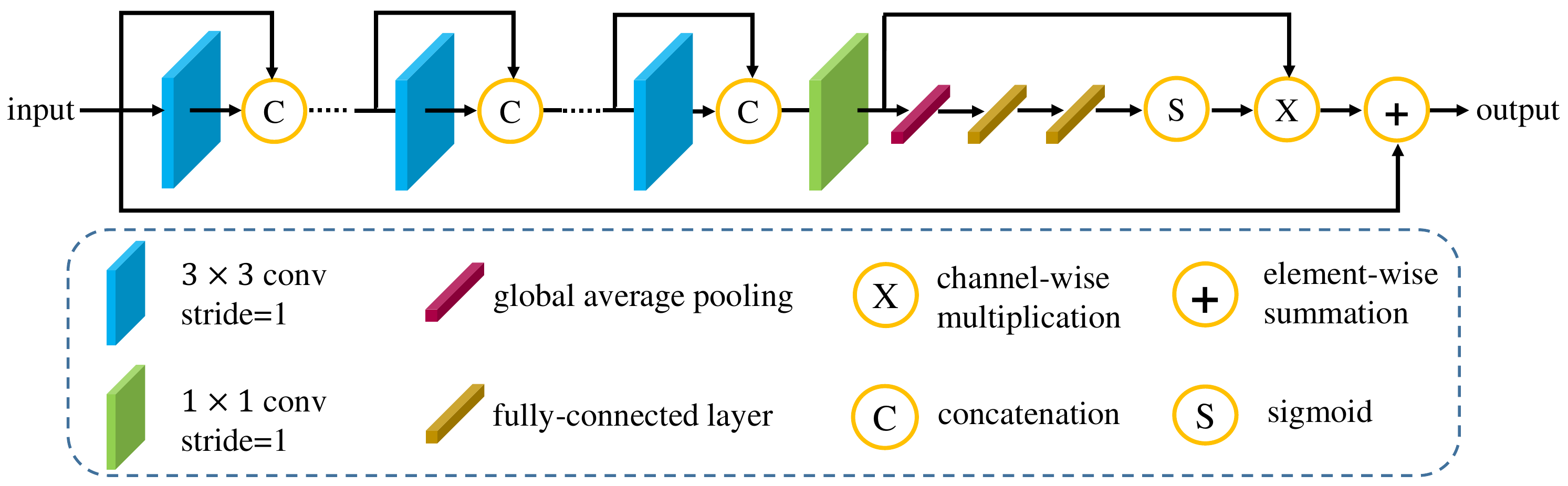} \\
		\end{tabular}
	\end{center}
	\vspace{-2mm}
\caption{Illustration of the proposed DCA block.
	}
	\label{fig:block}
	\vspace{-4mm}
\end{figure}

\vspace{-2mm}
\subsection{Network architecture}
\label{sec:network}
A straightforward approach to exploit raw data for super-resolution is to directly learn a mapping function from raw inputs to high-resolution color images with neural networks.
Whereas the raw data is important for image restoration,
such a simple approach does not perform well in practice.
The reason is that the raw data does not have the relevant information for color correction and tone enhancement within the ISP.
A raw image could potentially correspond to multiple ground-truth color images generated by different image processing algorithms of various ISPs, which will complicate the network training process and make it extremely difficult to train a model to reconstruct high-resolution images with desired colors.
In this work, we propose a two-branch CNN (see Figure~\ref{fig:network_overview}) where the first module exploits the linear radiance information recorded in the raw data $X_{raw}$ to restore the high-resolution linear measurements for all color channels with clear structures and fine details.
The second module estimates transformation matrices to recover high-fidelity color appearances for the restored linear measurements using the low-resolution color image $X_{ref}$ as a reference.
The reference image complements the limitations of raw data, and  jointly training two modules with both raw and color data helps reconstruct higher-quality results.

\vspace{-2mm}
{\bf\subsubsection{Image restoration}}\label{sec:image restoration}
The main components of the image restoration network are shown in Figure~\ref{fig:branch1}.
This network consists of four parts: packing raw data, low-level feature extraction, deep nonlinear mapping, and high-resolution image reconstruction. 
Similar to~\cite{learning_to_see_in_the_dark}, we first convert an input image $X_{raw}$ into a simpler form $X'_{raw}\in \mathbb{R}^{\frac{H}{4}\times \frac{W}{4} \times 4}$
by packing the Bayer raw data into four color channels which correspond to the R, G, B, G patterns in Figure~\ref{fig:bayer pattern}, and reducing the spatial resolution by a factor of two in each dimension.
Similar to \cite{SR-dense}, we extract low-level features $F_{low}$ from the packed four-channel input with a convolution layer with kernel of $3 \times 3$ pixels. %
To learn more complex nonlinear functions with the features $F_{low}$, we propose a dense channel-attention (DCA) block as shown in Figure~\ref{fig:block} for better image restoration performance.
Details about the DCA block are elaborated later in this section.
We consecutively apply four DCA blocks after $F_{low}$.
Different from prior work~\cite{densenet,SR-residual-dense,FSRCNN} which repeat the blocks at the same scale, we use the proposed blocks in an encoder-decoder framework (similar to U-net~\cite{ronneberger2015u}) to exploit multi-scale information. 
To downscale and upscale image features, we use convolution and deconvolution layers with stride of 2, respectively.
As shown in Section~\ref{sec:effect_restoration}, exploiting multi-scale features effectively reduces blurry artifacts and improves the super-resolution results.

Furthermore, we use skip connections to combine low-level and high-level feature maps with the same spatial size to generate the decoded features (\ie $F_{s1}$ and $F_{s2}$ in Figure~\ref{fig:branch1}).
We feed $F_{s2}$ into the final image reconstruction module consisting of a convolution layer to generate $48$ feature maps, and a sub-pixel layer~\cite{shi2016real} to rearrange the produced features into $\tilde{X}_{lin} \in \mathbb{R}^{H\times W \times 3}$. 
Here $\tilde{X}_{lin}$ represents the high-resolution linear output of the image restoration module which has clear structures and fine details, and is further processed in the second module to recover desirable color appearances.

\vspace{1ex}
{{\noindent \bf{DCA block.}}}  
The state-of-the-art neural networks~\cite{resnet,densenet,hu2018squeeze} usually rely on well-designed building blocks to achieve high-quality results.
Since it has recently been shown that the performance of convolutional networks can be substantially improved by dense connections~\cite{densenet}, we first apply $k$ densely-connected layers in the proposed DCA block.
As shown in Figure~\ref{fig:block}, for each layer of the dense-connection module we can obtain the output by first processing the input features with a 3$\times$3 convolution layer 
and then concatenating the convolved feature maps with the input.
The feature maps of all preceding layers are thus used as input of the current layer, and its own output features are used as inputs of all subsequent layers.

Note that the output of the dense module is a simple concatenation of 
features from all the convolution layers in the dense module.
Since these features are usually of different significance for final image restoration,
treating them channel-wise equally is not flexible during the training process and will restrict the representation strength of the network. 
To address this issue, we utilize the channel-attention mechanism~\cite{hu2018squeeze} which allows the network to recalibrate inter-channel features, such that informative feature maps are 
emphasized, and less useful ones are suppressed.

The channel attention module shown in Figure~\ref{fig:block} consists of a global average pooling which aggregates the input spatially to estimate channel-wise feature statistic, and two fully-connected layers which exploit the statistic of all channels to learn a gating function.
We use the sigmoid activation to normalize the gating function,
and the output can adaptively recalibrate the dense features with channel-wise multiplication. 
Similar to \cite{resnet}, we use a residual connection between the block input and output for better information flow. 
With the proposed DCA block, we can effectively improve the network representation capacity and achieve better performance for image super-resolution. 

\vspace{-2mm}
{\bf \subsubsection{Color correction}}\label{sec:color_correction}
After obtaining the linear measurements $\tilde{X}_{lin}$, we need to apply color correction and reconstruct the final result $\tilde{X}  \in \mathbb{R}^{H\times W \times 3}$ using the reference color image $X_{ref}$.
This can be formulated as a joint image filtering problem and one solution is to apply 
the guided filter~\cite{guidedFilter}:
\begin{equation}
\label{eq:guided}
\tilde{X}^c[\boldsymbol{x}]=\mathcal{A}_g(X^c_{ref},\tilde{X}^c_{lin})[\boldsymbol{x}] \tilde{X}^c_{lin}[\boldsymbol{x}] + \mathcal{B}_g(X^c_{ref},\tilde{X}^c_{lin})[\boldsymbol{x}],
\end{equation}
where $\mathcal{A}_g \in \mathbb{R}^{H\times W}$ and $\mathcal{B}_g \in \mathbb{R}^{H\times W}$ are the transformation matrices of the guided filter, which transfer the color appearances of $X_{ref}$ to $\tilde{X}$ while preserving the structures of $\tilde{X}_{lin}$. 
In \eqref{eq:guided}, $X^c$ indicates the $c$-th channel of the image $X$, and 
$[\boldsymbol{x}]$ is the spatial indexing operation which fetches an element from the spatial position $\boldsymbol{x} \in \mathbb{R}^{2}$.

However, both $\mathcal{A}_g$ and $\mathcal{B}_g$ are estimated based on local constraints~\cite{guidedFilter} that the transformation coefficients are constant in local windows.
Specifically, to solve $\mathcal{A}_g$ and $\mathcal{B}_g$ for a pixel $\mathbf{x}$, the guided filter takes a local window $\mathcal{N}_{\mathbf{x}}$ where $\mathbf{x} \in \mathcal{N}_{\mathbf{x}}$, and minimizes the following cost function:
\begin{equation} \label{eq:local_constant}
\sum_{\mathbf{y} \in \mathcal{N}_{\mathbf{x}}} ( \mathcal{A}_g [\mathbf{y}]  \tilde{X}_{lin}^c[\mathbf{y}] + \mathcal{B}_g [\mathbf{y}] - X_{ref}^c[\mathbf{y}] )^2,
\end{equation} 
where the transformed $\tilde{X}_{lin}$ is used to approximate the reference image $X_{ref}$.
Since the local constraints assume $\mathcal{A}_g$ and $\mathcal{B}_g$ are constant in $\mathcal{N}_{\mathbf{x}}$, $\mathcal{A}_g [\mathbf{x}]$ and $\mathcal{B}_g [\mathbf{x}]$ can be obtained by solving \eqref{eq:local_constant} with linear regression~\cite{guidedFilter}.
However, as the pixel $\boldsymbol{x}$ is involved in all windows that contain $\boldsymbol{x}$, the values of $\mathcal{A}_g[\mathbf{x}]$ and $\mathcal{B}_g[\mathbf{x}]$ 
are not the same when they are computed in different windows.
The final coefficients $\mathcal{A}_g[\mathbf{x}]$ and $\mathcal{B}_g[\mathbf{x}]$ are estimated by averaging all possible values.
Thus, the local constant assumption leads to oversmoothed transformation matrix $\mathcal{A}_g$ and $\mathcal{B}_g$, which cannot well handle the color correction around high-frequency structures (Figure~\ref{fig:guided}(d)).
To address this issue, we propose a learning-based guided filter for better color correction.

\begin{figure}[t]
	\footnotesize
	\begin{center}
		\begin{tabular}{c}
			\includegraphics[width = 0.95\linewidth]{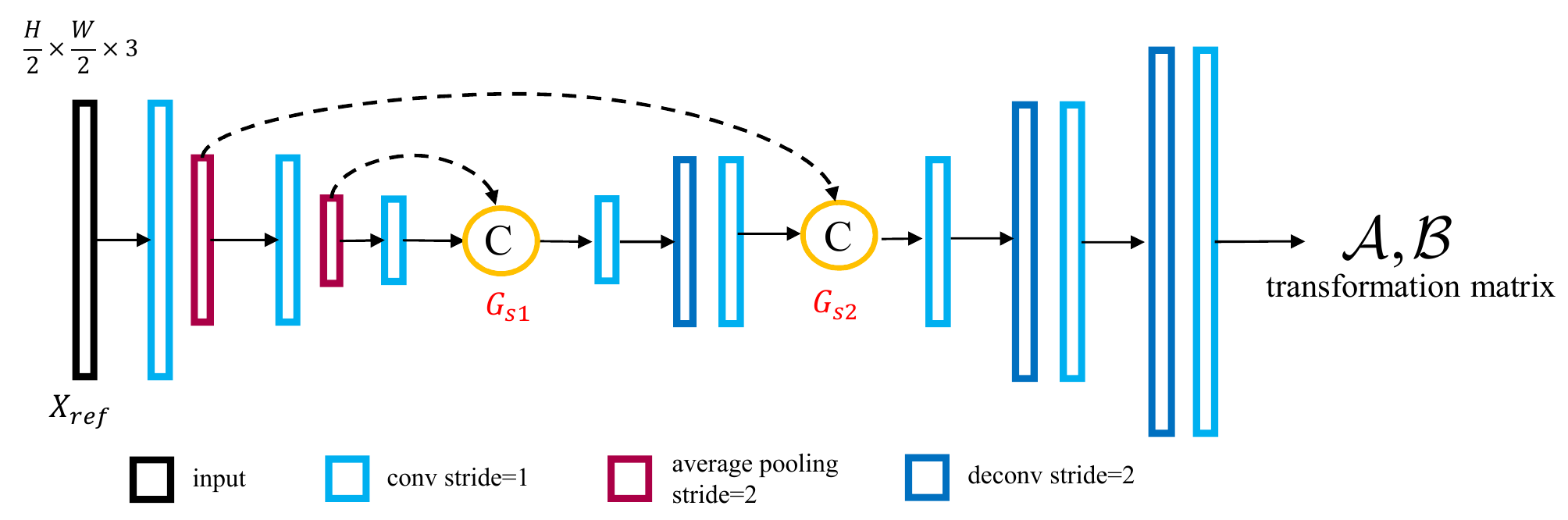} \\
		\end{tabular}
	\end{center}
	\vspace{-2mm}
\caption{Network architecture of the color correction module.
		Our model predicts the pixel-wise transformations $\mathcal{A}$ and $\mathcal{B}$ with a reference color image.
	}
	\label{fig:branch2}
	\vspace{-4mm}
\end{figure}

\begin{figure*}[t]
	\newcommand{\thisheight}{0.13\linewidth}
	\footnotesize
	\begin{center}
		\begin{tabular}{ccccccc}
			\hspace{-3mm} 
			\includegraphics[height = \thisheight]{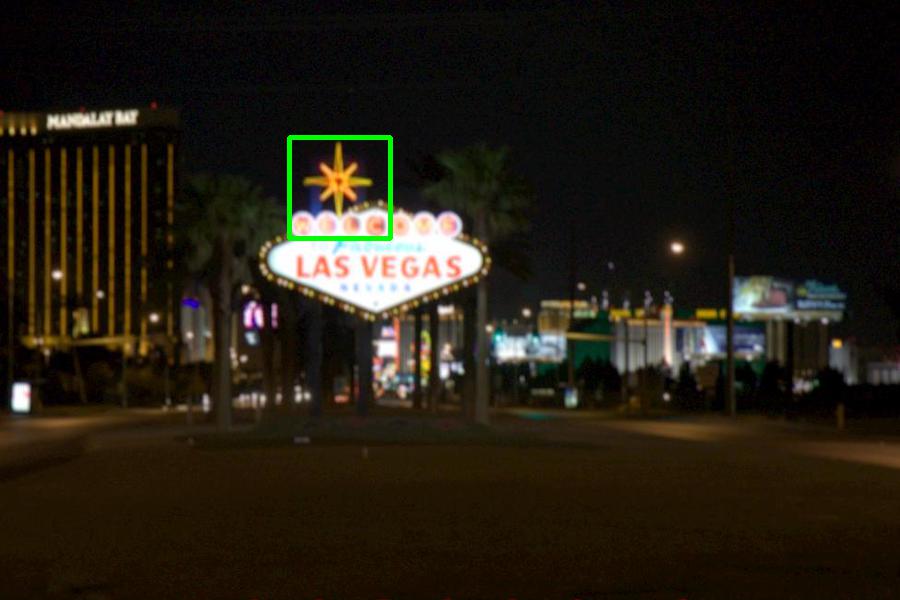} &
			\hspace{-4mm} 
			\includegraphics[height = \thisheight]{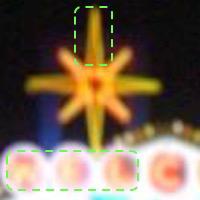} & 
			\hspace{-4mm} 
			\includegraphics[height = \thisheight]{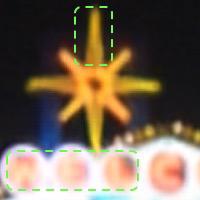} & 
			\hspace{-4mm}
			\includegraphics[height = \thisheight]{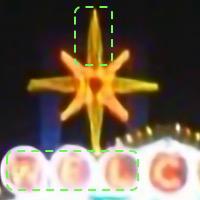} & 
			\hspace{-4mm}
			\includegraphics[height = \thisheight]{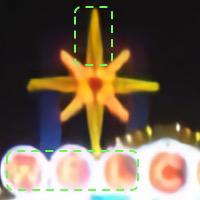} & 
			\hspace{-4mm}
			\includegraphics[height = \thisheight]{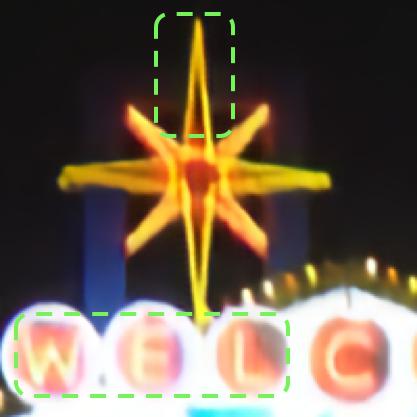} & 
			\hspace{-4mm}
			\includegraphics[height = \thisheight]{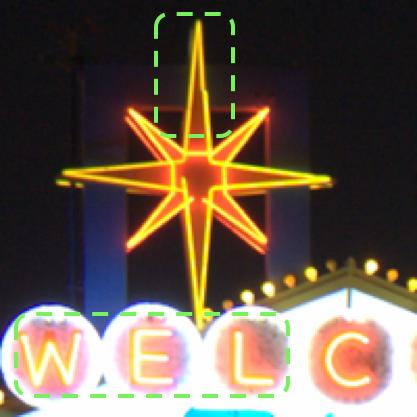} \\
			\hspace{-3mm} 
			\includegraphics[height = \thisheight]{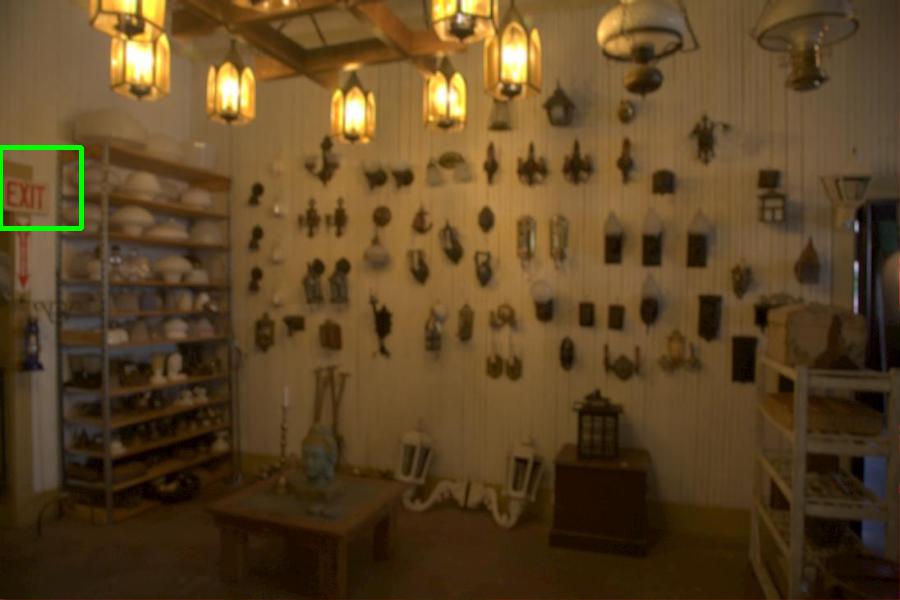} &
			\hspace{-4mm} 
			\includegraphics[height = \thisheight]{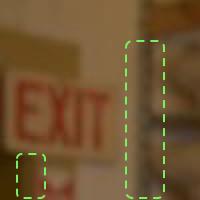} & 
			\hspace{-4mm} 
			\includegraphics[height = \thisheight]{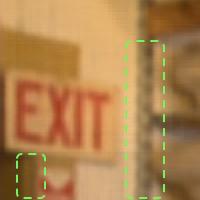} & 
			\hspace{-4mm}
			\includegraphics[height = \thisheight]{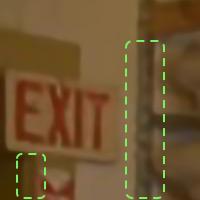} & 
			\hspace{-4mm}
			\includegraphics[height = \thisheight]{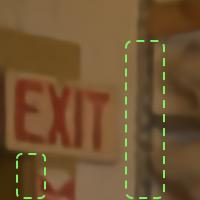} & 
			\hspace{-4mm}
			\includegraphics[height = \thisheight]{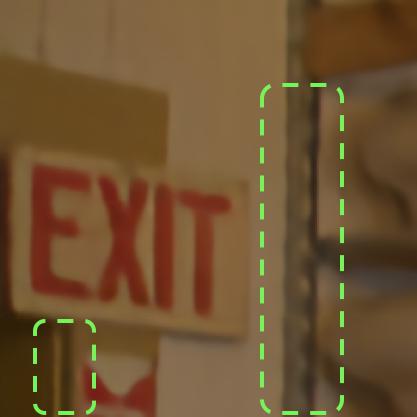} & 
			\hspace{-4mm}
			\includegraphics[height = \thisheight]{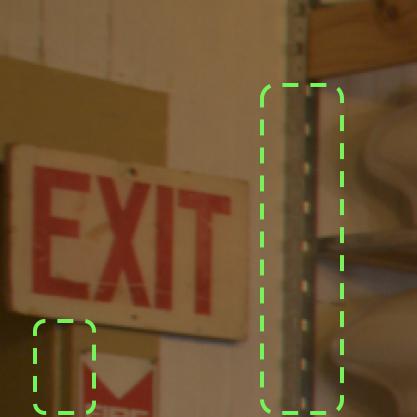} \\
			\hspace{-3mm} 
			\includegraphics[height = \thisheight]{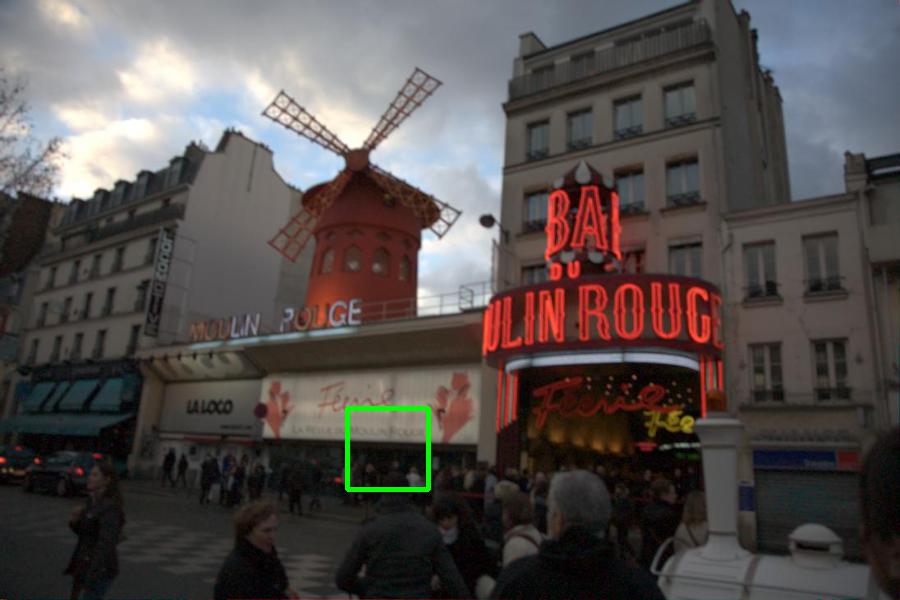} &
			\hspace{-4mm} 
			\includegraphics[height = \thisheight]{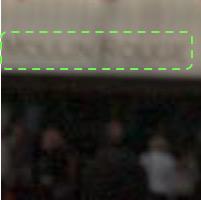} & 
			\hspace{-4mm} 
			\includegraphics[height = \thisheight]{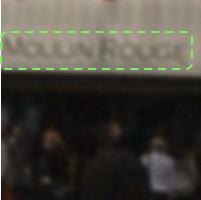} & 
			\hspace{-4mm}
			\includegraphics[height = \thisheight]{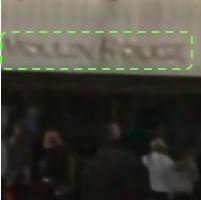} & 
			\hspace{-4mm}
			\includegraphics[height = \thisheight]{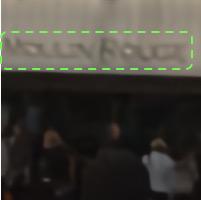} & 
			\hspace{-4mm}
			\includegraphics[height = \thisheight]{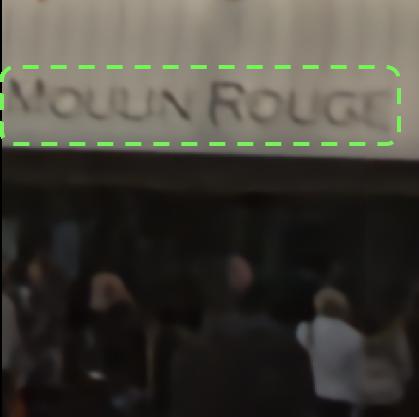} & 
			\hspace{-4mm}
			\includegraphics[height = \thisheight]{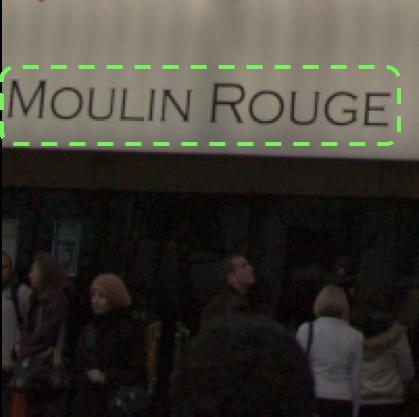} \\
			\hspace{-3mm} 
			\includegraphics[height = \thisheight]{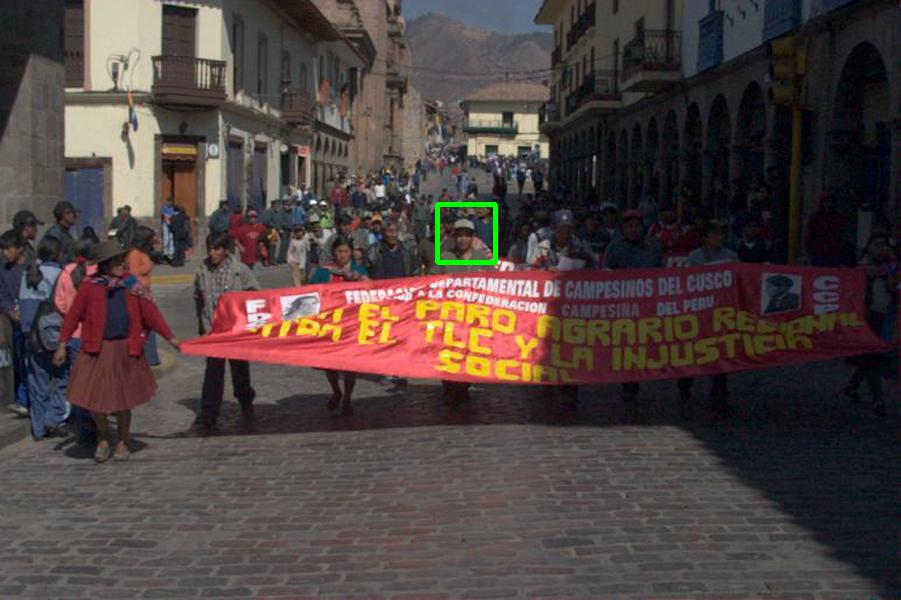} &
			\hspace{-4mm} 
			\includegraphics[height = \thisheight]{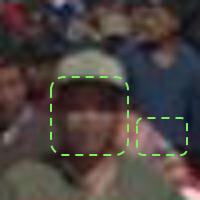} & 
			\hspace{-4mm} 
			\includegraphics[height = \thisheight]{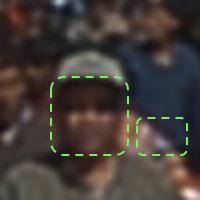} & 
			\hspace{-4mm}
			\includegraphics[height = \thisheight]{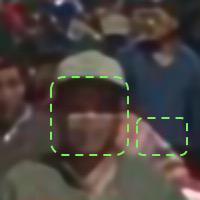} & 
			\hspace{-4mm}
			\includegraphics[height = \thisheight]{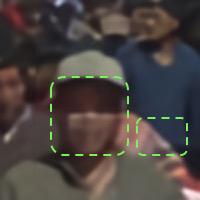} & 
			\hspace{-4mm}
			\includegraphics[height = \thisheight]{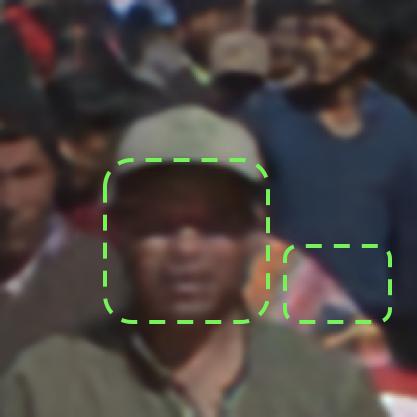} & 
			\hspace{-4mm}
			\includegraphics[height = \thisheight]{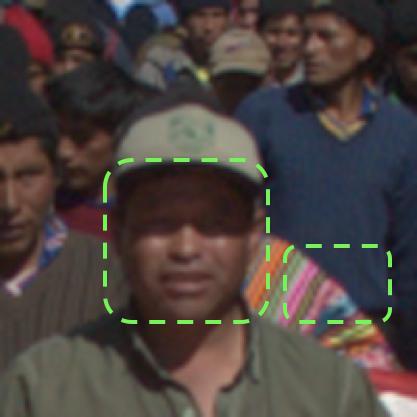} \\
			\hspace{-3mm}
			(a) Whole input image &
			\hspace{-4mm}
			(b) Input &
			\hspace{-4mm}
			(c) SID &
			\hspace{-4mm}
			(d) SRDenseNet &
			\hspace{-4mm}
			(e) RDN &
			\hspace{-4mm}
			(f) Ours &
			\hspace{-4mm}
			(g) GT \\
		\end{tabular}
	\end{center}
	\vspace{-2mm}
\caption{Results from the proposed synthetic dataset. References for the baseline methods can be found in Table~\ref{tab:syn1}. ``GT" represents ground-truth.
	}
	\label{fig:blind}
	\vspace{-4mm}
\end{figure*}

\begin{figure*}[t]
	\newcommand{\thisheight}{0.13\linewidth}
	\footnotesize
	\begin{center}
		\begin{tabular}{ccccccc}
			\hspace{-3mm} 
			\includegraphics[height = \thisheight]{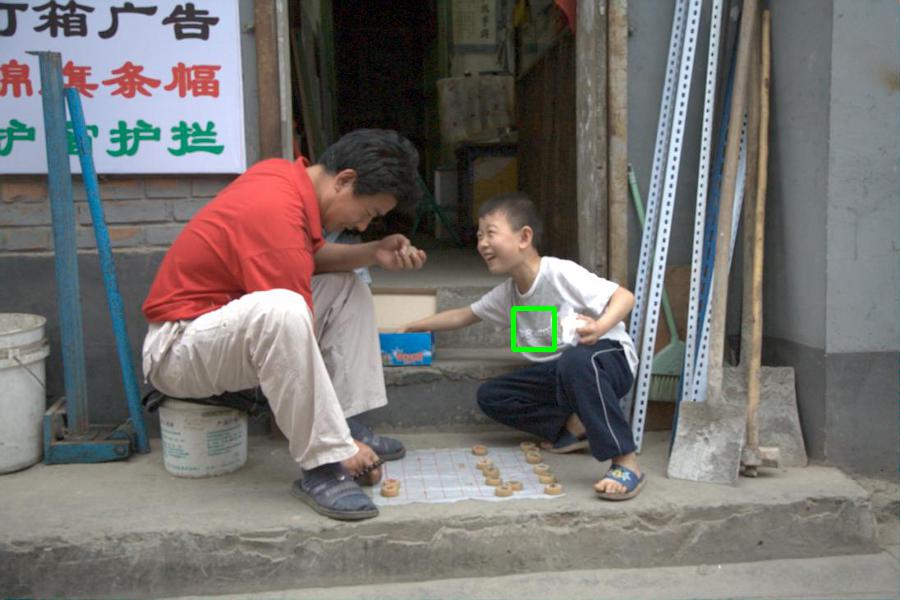} &
			\hspace{-4mm} 
			\includegraphics[height = \thisheight]{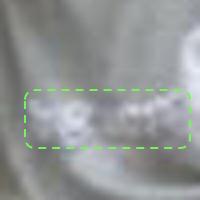} & 
			\hspace{-4mm} 
			\includegraphics[height = \thisheight]{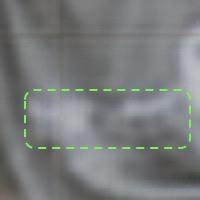} & 
			\hspace{-4mm}
			\includegraphics[height = \thisheight]{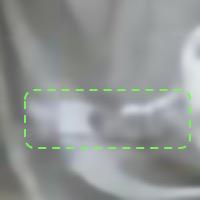} & 
			\hspace{-4mm}
			\includegraphics[height = \thisheight]{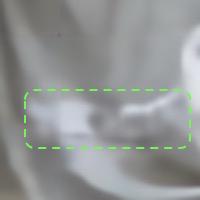} & 
			\hspace{-4mm}
			\includegraphics[height = \thisheight]{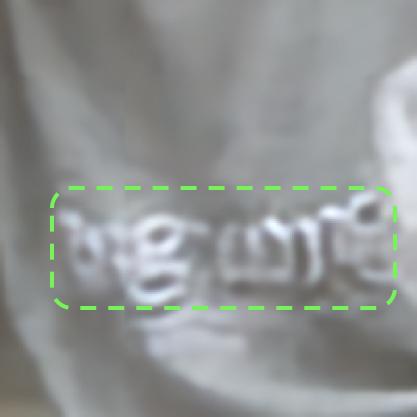} & 
			\hspace{-4mm}
			\includegraphics[height = \thisheight]{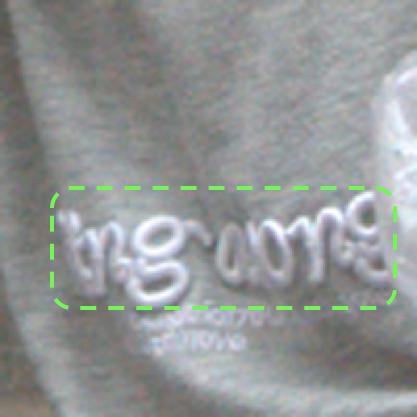} \\
			\hspace{-3mm} 
			\includegraphics[height = \thisheight]{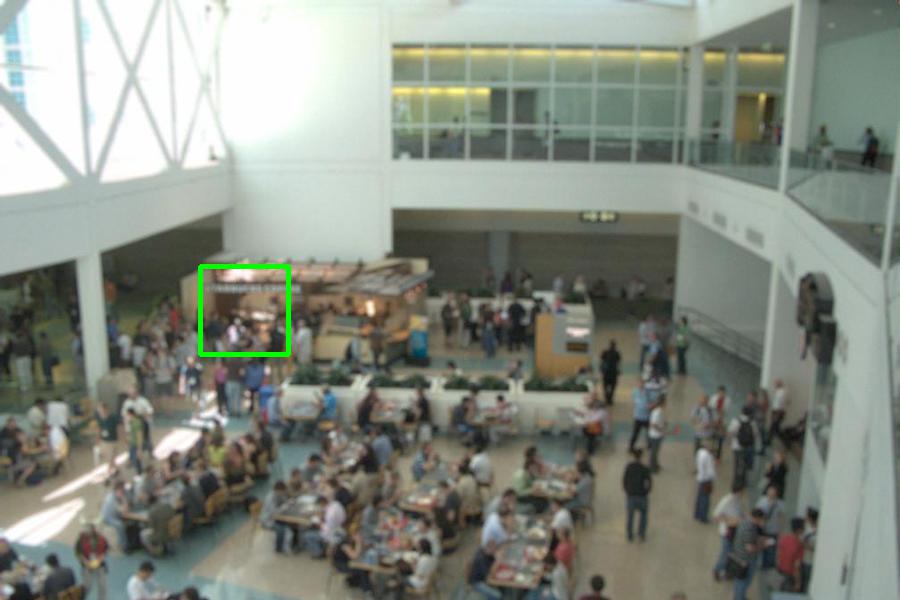} &
			\hspace{-4mm} 
			\includegraphics[height = \thisheight]{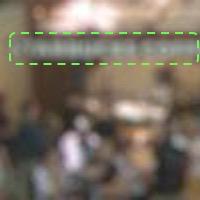} & 
			\hspace{-4mm} 
			\includegraphics[height = \thisheight]{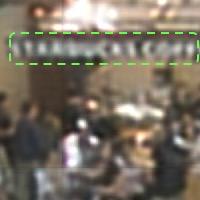} & 
			\hspace{-4mm}
			\includegraphics[height = \thisheight]{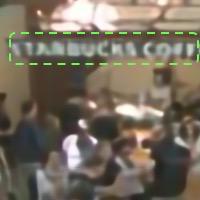} & 
			\hspace{-4mm}
			\includegraphics[height = \thisheight]{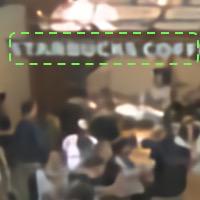} & 
			\hspace{-4mm}
			\includegraphics[height = \thisheight]{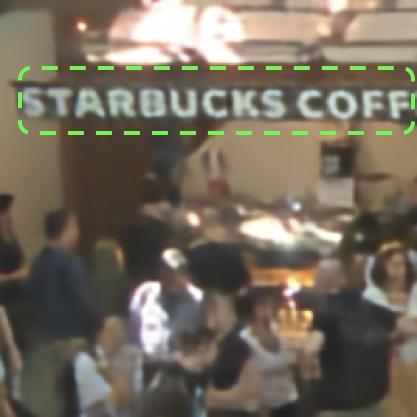} & 
			\hspace{-4mm}
			\includegraphics[height = \thisheight]{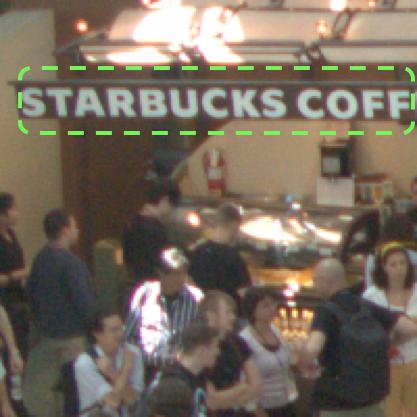} \\
			\hspace{-3mm} 
			\includegraphics[height = \thisheight]{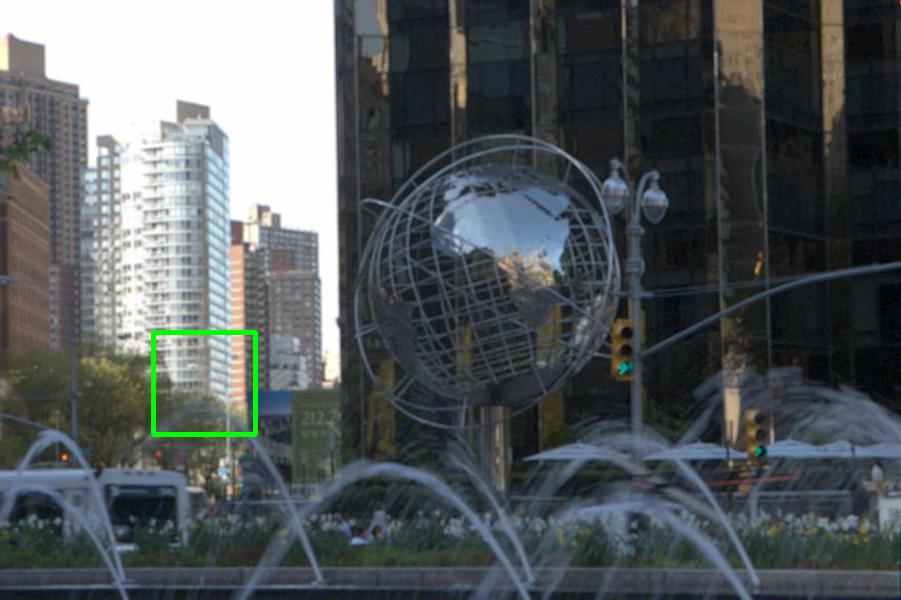} &
			\hspace{-4mm} 
			\includegraphics[height = \thisheight]{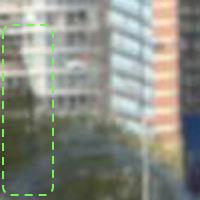} & 
			\hspace{-4mm} 
			\includegraphics[height = \thisheight]{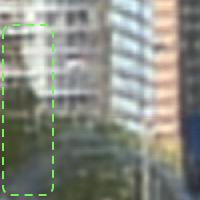} & 
			\hspace{-4mm}
			\includegraphics[height = \thisheight]{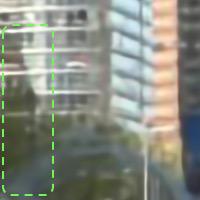} & 
			\hspace{-4mm}
			\includegraphics[height = \thisheight]{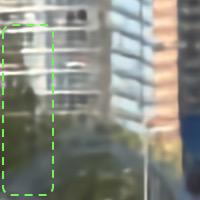} & 
			\hspace{-4mm}
			\includegraphics[height = \thisheight]{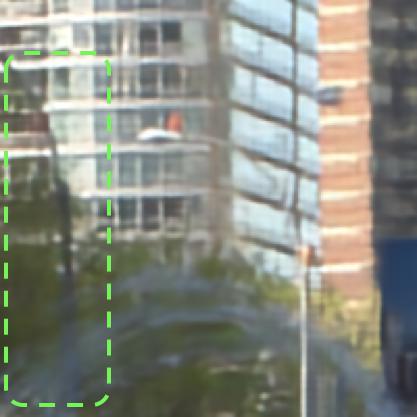} & 
			\hspace{-4mm}
			\includegraphics[height = \thisheight]{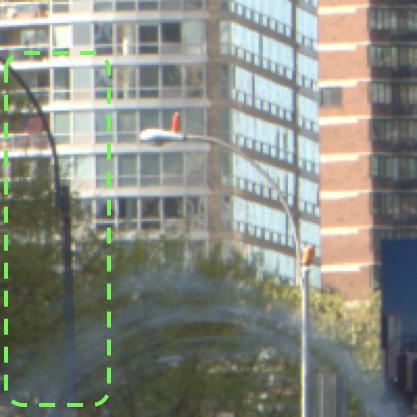} \\
			\hspace{-3mm} 
			\includegraphics[height = \thisheight]{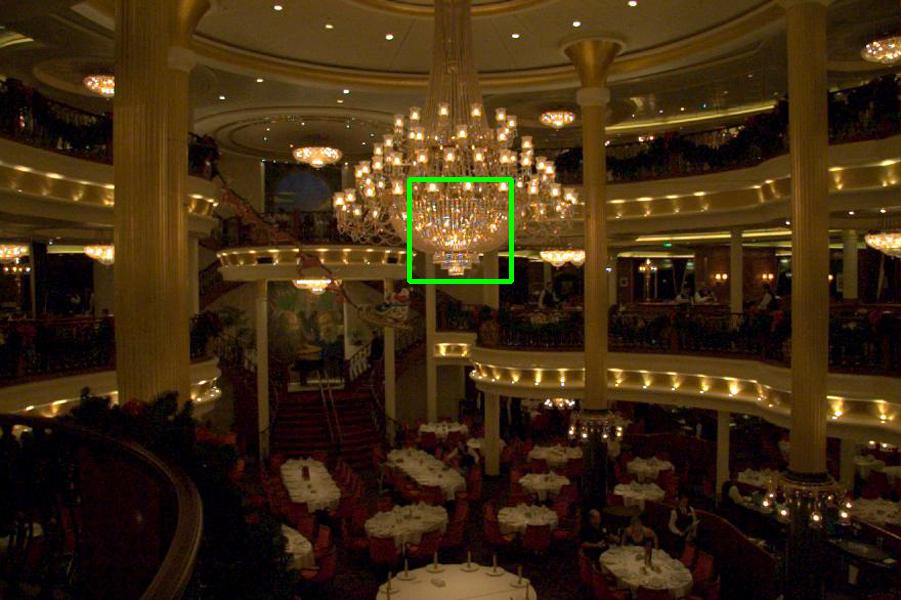} &
			\hspace{-4mm} 
			\includegraphics[height = \thisheight]{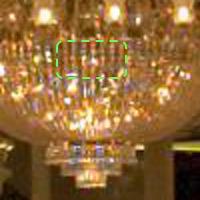} & 
			\hspace{-4mm} 
			\includegraphics[height = \thisheight]{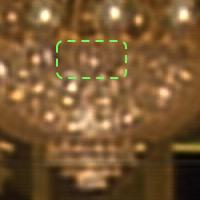} & 
			\hspace{-4mm}
			\includegraphics[height = \thisheight]{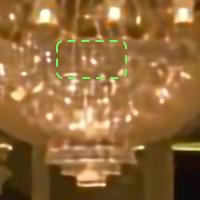} & 
			\hspace{-4mm}
			\includegraphics[height = \thisheight]{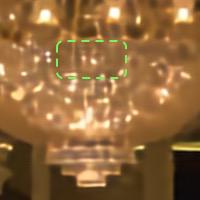} & 
			\hspace{-4mm}
			\includegraphics[height = \thisheight]{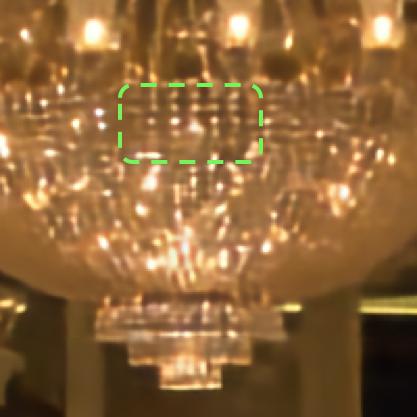} & 
			\hspace{-4mm}
			\includegraphics[height = \thisheight]{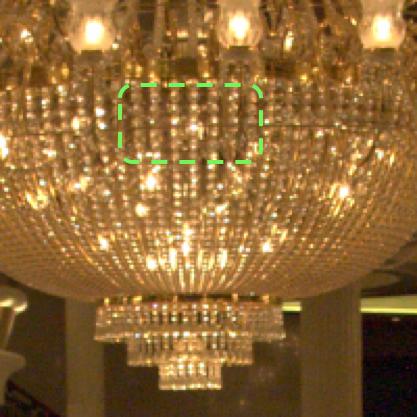} \\
			\hspace{-3mm}
			(a) Whole input image &
			\hspace{-4mm}
			(b) Input &
			\hspace{-4mm}
			(c) SID &
			\hspace{-4mm}
			(d) SRDenseNet &
			\hspace{-4mm}
			(e) RDN &
			\hspace{-4mm}
			(f) Ours &
			\hspace{-4mm}
			(g) GT \\
		\end{tabular}
	\end{center}
	\vspace{-2mm}
\caption{Results of non-blind image super-resolution. References for the baseline methods can be found in Table~\ref{tab:syn1}. ``GT" represents ground-truth.
	}
	\label{fig:nonblind}
	\vspace{-4mm}
\end{figure*}

\vspace{1ex}
{\noindent \bf{Learning-based guided filtering.}}
In contrast to the original guided filter which estimates $\mathcal{A}_g$ and $\mathcal{B}_g$ with  local constraints, we propose to learn the color correction matrices with neural networks.
Similar to \eqref{eq:guided}, the proposed filtering process can be formulated as:
\begin{equation}
\label{eq:our_guided}
\tilde{X}[\boldsymbol{x}]=\mathcal{A}(X_{ref},\tilde{X}_{lin})[\boldsymbol{x}] \tilde{X}_{lin}[\boldsymbol{x}] + \mathcal{B}(X_{ref},\tilde{X}_{lin})[\boldsymbol{x}],
\end{equation}
where $\mathcal{A} \in \mathbb{R}^{H\times W \times 3 \times 3}$ and $\mathcal{B} \in \mathbb{R}^{H\times W \times 3}$ are the transformation matrices predicted by a deep CNN (Figure~\ref{fig:branch2}).
In \eqref{eq:our_guided}, $\mathcal{A}[\boldsymbol{x}] \in \mathbb{R}^{3\times 3}$ transforms the RGB vector $\tilde{X}_{lin}[\boldsymbol{x}] \in \mathbb{R}^{3}$ to a new 3-channel color space, 
and $\mathcal{B}[\boldsymbol{x}] \in \mathbb{R}^{3}$ serves as biases of the transformation.
Since the learning-based guided filter is not restricted by the local constant assumption, the learned transformations can better adapt to high-frequency color changes and achieve higher-quality color correction results.
In addition, the original guided filter processes each image channel independently, whereas we learn pixel-wise 3$\times$3 matrices which can exploit the inter-channel information for better color transformation.

We use a light-weight CNN to predict the transformations $\mathcal{A}$ and $\mathcal{B}$ as shown in Figure~\ref{fig:branch2}.
The CNN begins with an encoder including three convolution layers with 2$\times$2 average pooling to extract the color features from $X_{ref}$.
To estimate the spatially-variant transformations, 
we use a decoder consisting of consecutive deconvolution and convolution layers, which expands the encoded features to the desired resolution $H\times W$.
We concatenate the same-resolution feature maps in the encoder and decoder to exploit hierarchical representations.
Finally, the output layer of the decoder  generates 3$\times$3 weights and 3$\times$1 biases for each pixel, which form the pixel-wise transformation matrices $\mathcal{A}$ and $\mathcal{B}$.

Note that in addition to the network input $X_{ref}$, $\mathcal{A}(X_{ref},\tilde{X}_{lin})$ and $\mathcal{B}(X_{ref},\tilde{X}_{lin})$ are estimated based on 
the output of the first module $\tilde{X}_{lin}$ as well.
Instead of directly using $\tilde{X}_{lin}$ in the estimation network, we exploit 
richer features extracted from the restoration module which have been used to generate $\tilde{X}_{lin}$.
Specifically, we fuse the decoded features $F_{s1}$ and $F_{s2}$ into the color correction module via pixel-wise summation:
\vspace{-1mm}
\begin{equation}
	G'_{si} = G_{si} + \phi_i(F_{si}), ~~ i=1, 2,
\vspace{-1mm}
\end{equation}
where $G_{si}$ is the feature after the concatenation layer of the color correction network as shown in Figure~\ref{fig:branch2}.
As $G_{si}$ and $F_{si}$ usually have different numbers of channels, we fuse them by introducing a 1$\times$1 convolution layer $\phi_i$ whose weights are learned to adaptively transform $F_{si}$.
After feature fusion, we put the updated features $G'_{si}$ back to the color correction network, and the following layers can further exploit the fused features for transformation matrix prediction.
Note that we adopt a multi-scale feature fusion strategy  to improve our early model~\cite{rawSR} where only single scale features (\ie $F_{s1}$ and $G_{s1}$) are used.  
As the predicted transformations are applied to $\tilde{X}_{lin}$, it is beneficial to make the matrix estimation network aware of the multi-scale features in the image restoration module, such that $\mathcal{A}$ and $\mathcal{B}$ can adapt more accurately to the structures of the restored image.

\vspace{1ex}
{\noindent \bf{Relationship to global color correction.}}
Schwartz~\etal \cite{deepisp_tip} estimate a global transformation for color correction of the image.
Similarly, we can also apply this method to replace the proposed learning-based guided filter:
\begin{equation} \label{eq:global_color}
\tilde{X}[\boldsymbol{x}]=\mathcal{A}_s(X_{ref},\tilde{X}_{lin}) \tilde{X}_{lin}[\boldsymbol{x}] + \mathcal{B}_s(X_{ref},\tilde{X}_{lin}),
\end{equation}
where $\mathcal{A}_s \in \mathbb{R}^{3 \times 3}$ and $\mathcal{B}_s \in \mathbb{R}^{3}$ represent the global transformation predicted by a CNN with global average pooling and fully-connected layers~\cite{deepisp_tip}.
However, this global approach does not perform well when the color transformation of the ISP involves spatially-variant operations. 
In contrast, we learn pixel-wise transformations $\mathcal{A}[\boldsymbol{x}]$ and $\mathcal{B}[\boldsymbol{x}]$ which allow different color corrections for each spatial location $\boldsymbol{x}$. 
Note that \cite{deepisp_tip} uses a quadratic form of the RGB vector $\tilde{X}_{lin}[\boldsymbol{x}]$ for \eqref{eq:global_color}, from which, we find no obvious improvements empirically.%

\vspace{1ex}
{\noindent \bf{Relationship to end-to-end trainable guided filter.}}
Another related method of the color correction network is the end-to-end trainable guided filter (ETGF)~\cite{wu2018fast}:
\vspace{-1mm}
\begin{align}
{	\label{eq:et_guided} \footnotesize
	\tilde{X}^c[\boldsymbol{x}]=\mathcal{A}_g(X^c_{ref},\varphi(\tilde{X}_{lin})^c)[\boldsymbol{x}] \varphi(\tilde{X}_{lin})^c[\boldsymbol{x}] + \mathcal{B}_g(X^c_{ref},\varphi(\tilde{X}_{lin})^c)[\boldsymbol{x}], }
\vspace{-1mm}
\end{align}
where $\varphi$ is a two-layer CNN to transform $\tilde{X}_{lin}$ into a new 3-channel representation for filtering, and 
$\varphi(\tilde{X}_{lin})^c$ is the $c$-th channel of $\varphi(\tilde{X}_{lin})$.
Since $\mathcal{A}_g$ and $\mathcal{B}_g$ are computed with the same method as the original guided filter \eqref{eq:guided},
the ETGF is also restricted by the local constant assumption \cite{guidedFilter}, and thereby cannot adapt to high-frequency color changes as effectively as the proposed learning-based guided filter~\eqref{eq:our_guided}.

\vspace{4mm}
\section{Experimental Results}
\label{sec:experiment}
We first describe the implementation details as well as datasets, and then evaluate the proposed algorithm on both synthetic and real images. 
We present the main findings in this section and more results on the project website \url{https://sites.google.com/view/xiangyuxu/rawsr_pami}.

\begin{table}[t]
	\caption{\label{tab:syn1}Quantitative evaluations on the synthetic dataset. ``Blind" represents the images with variable blur kernels, and ``Non-blind" denotes fixed kernel.
	}
	\vspace{-2mm}
	\footnotesize
	\centering
	\begin{tabular}{ccccc}
		\toprule
		\multicolumn{1}{c}{\multirow{2}*{Methods}}&\multicolumn{2}{c}{Blind}&\multicolumn{2}{c}{Non-blind}\\
		\cmidrule{2-5}
		& PSNR & SSIM & PSNR & SSIM \\    %
		\midrule
		SID~\cite{learning_to_see_in_the_dark} & 21.87 & 0.7325 & 21.89 & 0.7346  \\
		DeepISP~\cite{deepisp_tip}& 21.71 & 0.7323 & 21.84 & 0.7382  \\
		SRCNN~\cite{SRCNN}& 28.83 & 0.7556 & 29.19 & 0.7625  \\
		VDSR~\cite{SR-VDSR} & 29.32 & 0.7686 & 29.88 & 0.7803  \\
		SRDenseNet~\cite{SR-dense} & 29.46 & 0.7732 & 30.05 & 0.7844  \\
		RDN~\cite{SR-residual-dense}& 29.93 & 0.7804 & 30.46 & 0.7897  \\
		\midrule
		SID$^*$~\cite{learning_to_see_in_the_dark}& 29.55 & 0.7715 & 29.76 & 0.7749 \\
		RDN$^*$~\cite{SR-residual-dense} & 30.35 & 0.7954  & 31.19 & 0.8211 \\
		Ours & \bf 31.06 & \bf 0.8088 & \bf 32.07 & \bf 0.8314 \\
		\bottomrule
	\end{tabular}
	\vspace{-4mm}
\end{table}

\vspace{-2mm}
\subsection{Implementation details}\label{sec:implementation}
To generate the training data, we use the MIT-Adobe 5K dataset~\cite{adobe-mit-5k} which consists of  $5000$ raw photographs with image size around $2000\times 3000$ pixels. 
After manually removing images with noticeable noise and blur,
we obtain $1300$ high-quality images for training and $150$ for testing which are captured by $19$ types of Canon cameras.
We use the method described in Section~\ref{sec:training_data} to synthesize the training and test datasets.
The radius of the defocus blur is randomly sampled from $[1, 5]$ pixels and the maximum size of the motion kernel is sampled from $[3, 11]$ pixels. 
The noise parameters $\sigma_1, \sigma_2$ are respectively sampled from $[0,10^{-2}]$ and $[0,10^{-3}]$.
%
%
%
%
%
%
%The source code, datasets and models will be made publicly available on the project website.

We use the Xavier initializer~\cite{glorot2010understanding} for the network weights and LeakyReLU~\cite{leakyRelu} with slope  of 0.2 as the activation function.
For the DCA block, we set the growth rate as 16 and $k$ as $8$.
We crop $256\times 256$ patches as input and use a batch size of $6$ for training. 
During the testing phase, 
we divide an input image into overlapping patches and process each patch separately.
The reconstructed high-resolution patches are placed back to the corresponding locations and averaged in overlapping regions.
This testing strategy is also used in other image processing algorithms~\cite{schuler2013machine,SR-VDSR,SRCNN} to alleviate  memory issues.

\begin{figure*}[t]
	\newcommand{\thisheight}{0.14\linewidth}
	\footnotesize
	\begin{center}
		\begin{tabular}{cccccc}
			\hspace{-3mm} 
			\includegraphics[height = \thisheight]{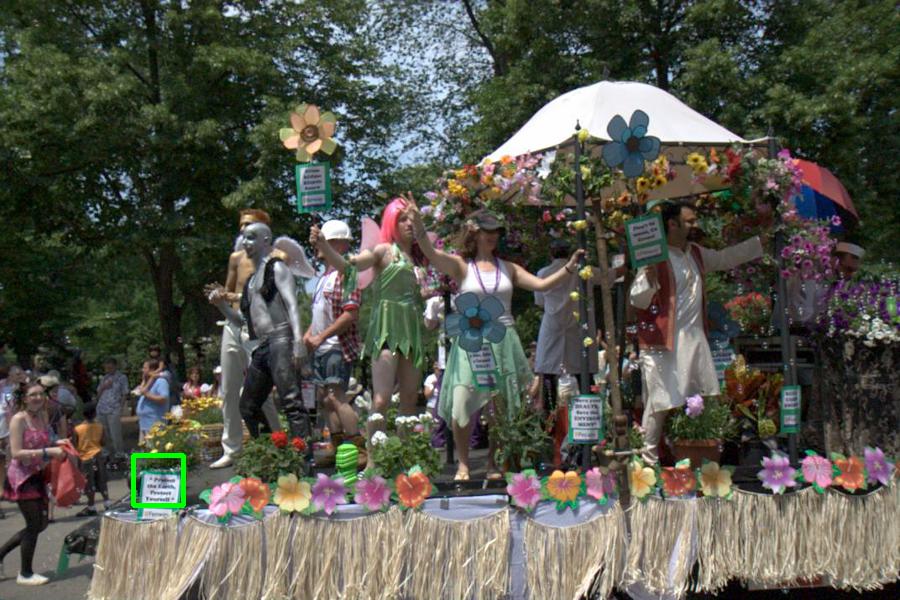} &
			\hspace{-4mm} 
			\includegraphics[height = \thisheight]{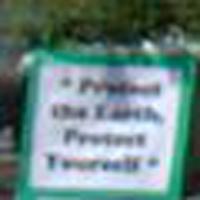} & 
			\hspace{-4mm} 
			\includegraphics[height = \thisheight]{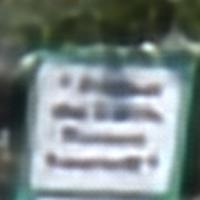} & 
			\hspace{-4mm}
			\includegraphics[height = \thisheight]{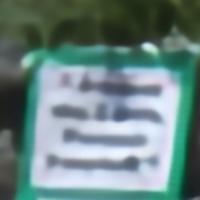} & 
						\hspace{-4mm}
						\includegraphics[height = \thisheight]{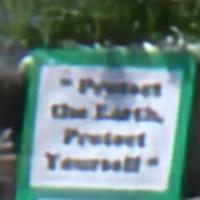} & 
			\hspace{-4mm}
			\includegraphics[height = \thisheight]{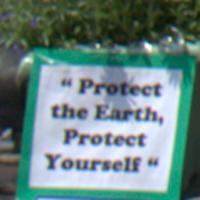} \\
			\hspace{-3mm} 
			\includegraphics[height = \thisheight]{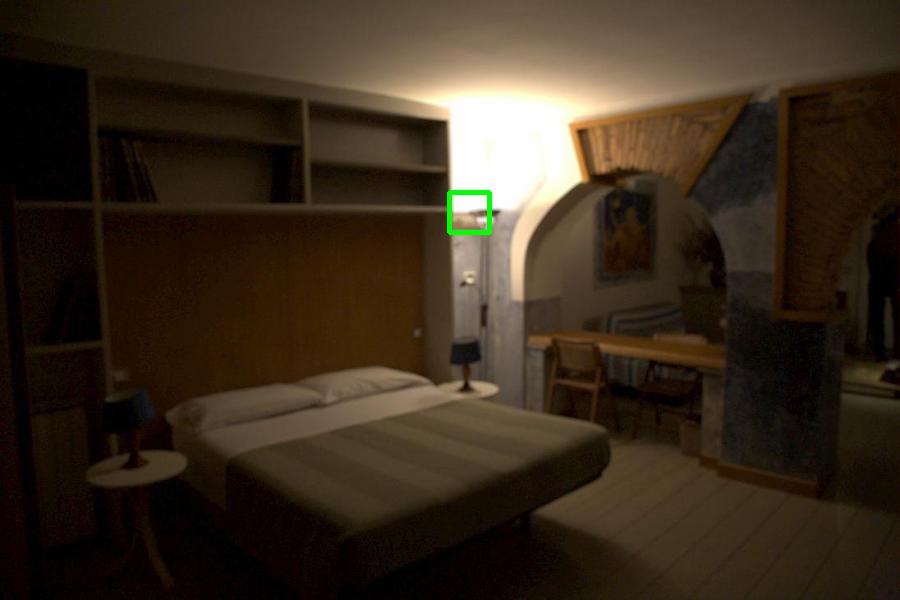} &
			\hspace{-4mm} 
			\includegraphics[height = \thisheight]{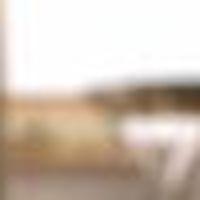} & 
			\hspace{-4mm} 
			\includegraphics[height = \thisheight]{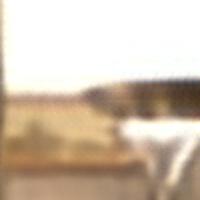} & 
			\hspace{-4mm}
			\includegraphics[height = \thisheight]{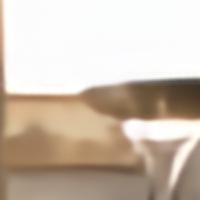} & 
			\hspace{-4mm}
			\includegraphics[height = \thisheight]{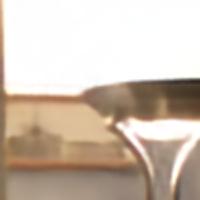} & 
			\hspace{-4mm}
			\includegraphics[height = \thisheight]{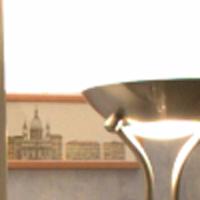} \\
			\hspace{-3mm}
			(a) Whole input image &
			\hspace{-4mm}
			(b) Input &
			\hspace{-4mm}
			(c) w/o color &
			\hspace{-4mm}
			(d) w/o raw &
			\hspace{-4mm}
			(e) Our full model &
			\hspace{-4mm}
			(f) GT \\
		\end{tabular}
	\end{center}
	\vspace{-2mm}
	\caption{Visual comparisons of different variants of the proposed network. ``w/o color'' indicates the model without the color correction module. ``w/o raw'' represents super-resolution without raw input by using the low-resolution color image as the input for both modules of our network. %
	}
	\label{fig:ablation}
	\vspace{-3mm}
\end{figure*}

\begin{table}[t]
	\caption{\label{tab:ablation} Ablation study of the proposed model on the synthetic dataset. 
	}
	\vspace{-2mm}
	\footnotesize
	\centering
	\begin{tabular}{ccccc}
		\toprule
		\multicolumn{1}{c}{\multirow{2}*{Methods}}&\multicolumn{2}{c}{Blind}&\multicolumn{2}{c}{Non-blind}\\
		\cmidrule{2-5}
		& PSNR & SSIM & PSNR & SSIM \\    %
		\midrule
		w/o color module& 22.04 & 0.7426  & 22.13 & 0.7599 \\
		w/o raw input & 30.21 & 0.7864 & 30.65 & 0.7948  \\
		Our full model & \bf 31.06 & \bf 0.8088 & \bf 32.07 & \bf 0.8314 \\
		\bottomrule
	\end{tabular}
	\vspace{-4mm}
\end{table}

\vspace{1ex}
{\noindent \bf{Model training.}} %
We use the $L_1$ loss function to train the network and adopt the Adam optimizer~\cite{kingma2014adam} with initial learning rate of $2\times 10^{-4}$.
Let $\{X^{(i)}_{raw}, X^{(i)}_{ref}, X^{(i)}_{lin}, X^{(i)}_{gt}\}_{i=1}^N$ denotes our dataset which has $N$ training samples.
We first pre-train the image restoration module using  $\frac{1}{N} \sum_{i=1}^{N} \|\tilde{X}^{(i)}_{lin}-X^{(i)}_{lin}\|_1$ for $4\times10^{4}$ iterations with a fixed learning rate.
Then we train the whole network using $\frac{1}{N} \sum_{i=1}^{N} \|\tilde{X}^{(i)}-X^{(i)}_{gt}\|_1$ for $8\times10^{4}$ iterations,
where we decrease the learning rate by a factor of $0.96$ every $2\times 10^3$ updates during the first $4\times10^{4}$ iterations, and then reduce it to a lower rate  of $10^{-5}$ for the latter $4\times10^{4}$ iterations.

\vspace{1ex}
{\noindent \bf{Baseline methods.}} %
\rev{We evaluate our model with the state-of-the-art super-resolution methods~\cite{SRCNN,SR-VDSR,SR-dense,SR-residual-dense} which take the degraded color image as input, and the learning-based raw image processing algorithms~\cite{learning_to_see_in_the_dark,deepisp_tip} that use low-resolution raw data.}
As the raw image processing networks~\cite{learning_to_see_in_the_dark,deepisp_tip} are not designed for super-resolution, we add deconvolution layers 
to increase the resolution of the output.
All the baseline models are trained on our dataset with the same settings.

\vspace{-2mm}
\subsection{Evaluation with the state-of-the-arts}
We quantitatively evaluate our method using the test dataset described above.
As shown in Table~\ref{tab:syn1}, the proposed algorithm performs favorably against the baseline methods in terms of PSNR and structural similarity (SSIM)~\cite{wang2004image}. 
Figures~\ref{fig:blind} shows some restored results from our synthetic dataset. 
\rev{Since the low-resolution raw input does not have the relevant information for color conversion and tone adjustment carried out within the ISP, the raw image processing methods~\cite{learning_to_see_in_the_dark,deepisp_tip} can only approximate 
color corrections for one specific camera, and thereby do not recover desirable color appearances 
for diverse camera types in our test set (Figure~\ref{fig:blind}(c)). 
In addition, the training process of the raw processing models is negatively affected by the significant color differences between the
predictions and ground truth, and the network cannot focus on restoring sharp structures due to the distraction of the color errors.
Thus, the restored images in Figures~\ref{fig:blind}(c) are still blurry.}
Figure~\ref{fig:blind}(d)-(e) show that the super-resolution methods with low-resolution color image as input can generate clear results with correct colors. 
However, some fine details are missing. 
In contrast, our method generates better results with appealing color appearances as well as sharp structures and fine details, as shown in Figure~\ref{fig:blind}(f), 
by exploiting both the color input and raw radiance information.  
\rev{
For a more comprehensive study, we improve the two representative baselines SID~\cite{learning_to_see_in_the_dark} and RDN~\cite{SR-residual-dense} by feeding both the raw and color images into these models.   
As shown in Table~\ref{tab:syn1}, this simple modification leads to a significant performance boost for both the SID and RDN, which demonstrates the effectiveness of exploiting the complementary information from the two inputs.
On the other hand, our network still outperforms the improved baselines by a clear margin, which 
shows the effectiveness of the proposed two-branch model.
}

\vspace{1ex}
{\noindent \bf{Non-blind super-resolution.}} %
Since most super-resolution methods~\cite{SRCNN,SR-VDSR,SR-residual-dense} are non-blind with the assumption that the downsampling blur kernel is known and fixed,
we also evaluate our algorithm for this case by using fixed defocus kernel with radius of 5 pixels to synthesize training and test datasets.
As shown in Table~\ref{tab:syn1} and Figure~\ref{fig:nonblind}, the proposed algorithm can also achieve better results than the baseline methods under the non-blind settings.

\begin{figure*}[t]
	\newcommand{\thisheight}{0.14\linewidth}
	\footnotesize
	\begin{center}
		\begin{tabular}{cccccc}
			\hspace{-3mm} 
			\includegraphics[height = \thisheight]{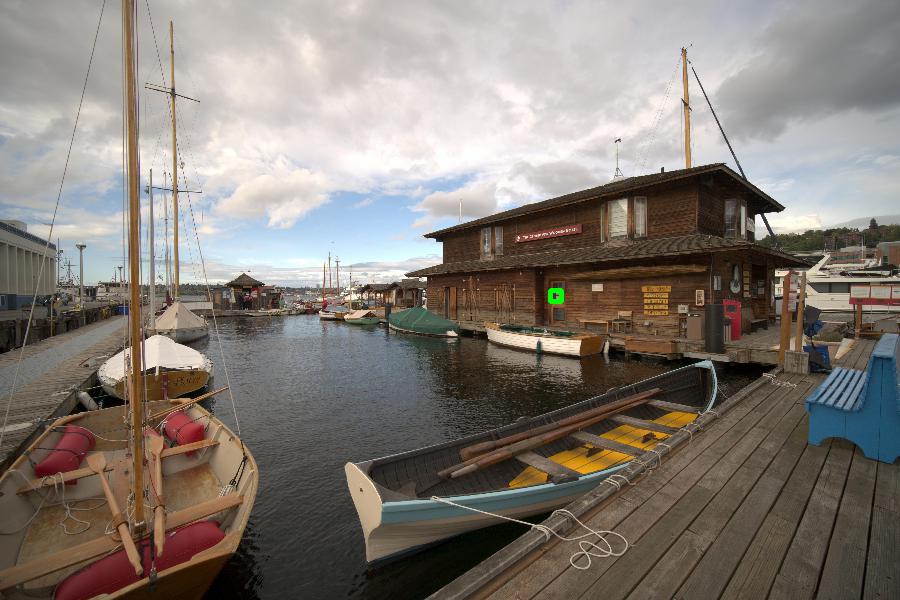} &
			\hspace{-4mm} 
			\includegraphics[height = \thisheight]{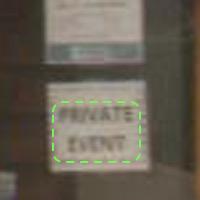} & 
			\hspace{-4mm} 
			\includegraphics[height = \thisheight]{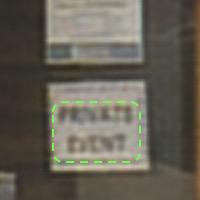} & 
			\hspace{-4mm}
			\includegraphics[height = \thisheight]{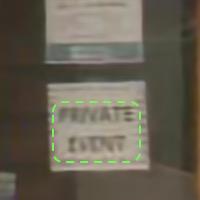} & 
			\hspace{-4mm}
			\includegraphics[height = \thisheight]{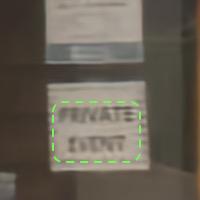} & 
			\hspace{-4mm}
			\includegraphics[height = \thisheight]{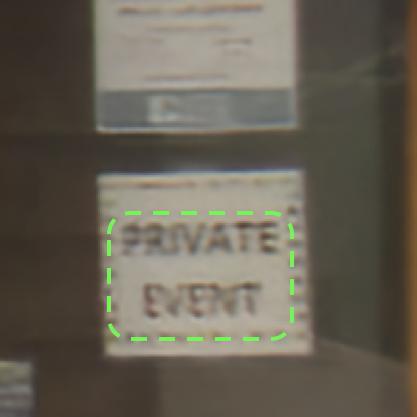} \\ 
			\hspace{-3mm} 
			\includegraphics[height = \thisheight]{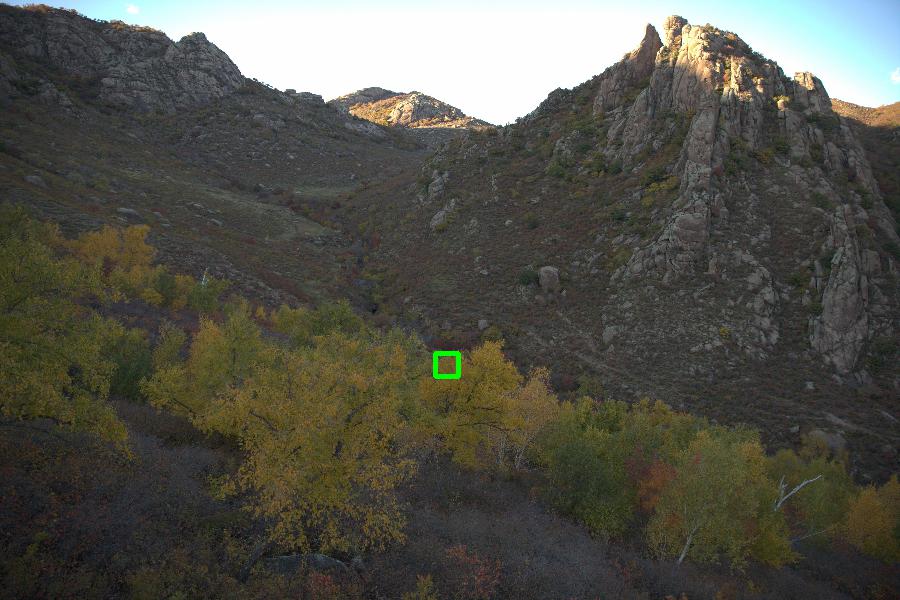} &
			\hspace{-4mm} 
			\includegraphics[height = \thisheight]{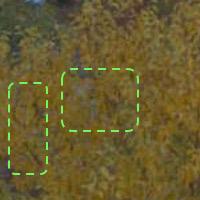} & 
			\hspace{-4mm} 
			\includegraphics[height = \thisheight]{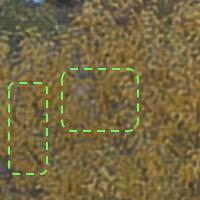} & 
			\hspace{-4mm}
			\includegraphics[height = \thisheight]{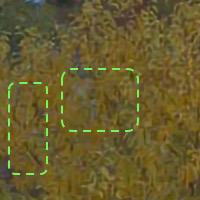} & 
			\hspace{-4mm}
			\includegraphics[height = \thisheight]{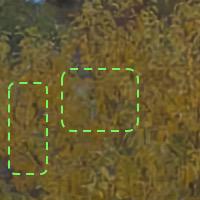} & 
			\hspace{-4mm}
			\includegraphics[height = \thisheight]{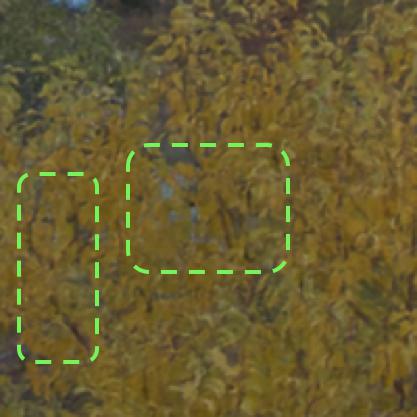} \\ 
			\hspace{-3mm} 
			\includegraphics[height = \thisheight]{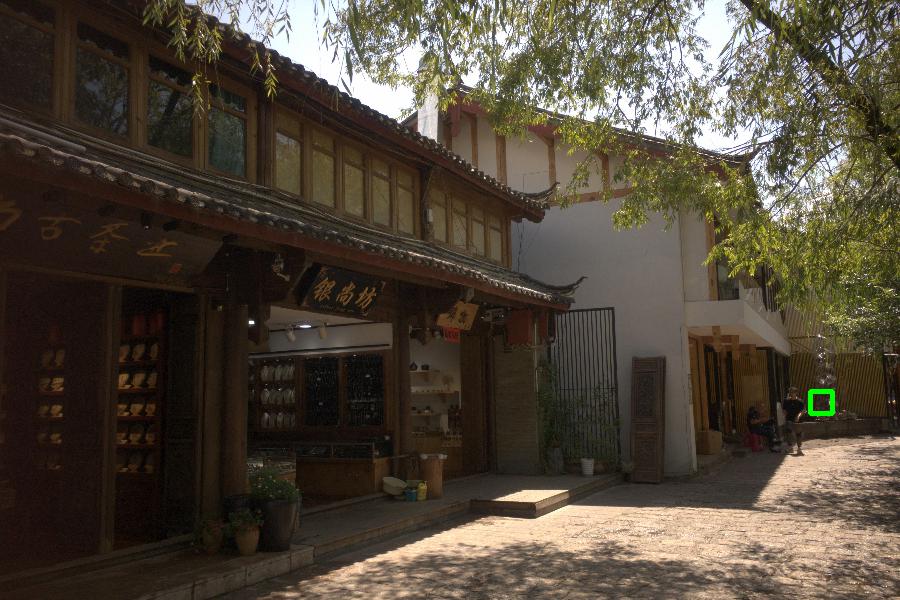} &
			\hspace{-4mm} 
			\includegraphics[height = \thisheight]{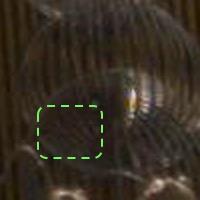} & 
			\hspace{-4mm} 
			\includegraphics[height = \thisheight]{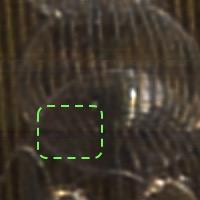} & 
			\hspace{-4mm}
			\includegraphics[height = \thisheight]{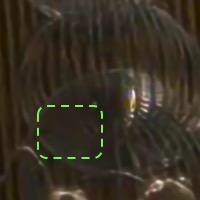} & 
			\hspace{-4mm}
			\includegraphics[height = \thisheight]{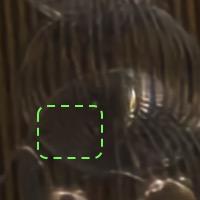} & 
			\hspace{-4mm}
			\includegraphics[height = \thisheight]{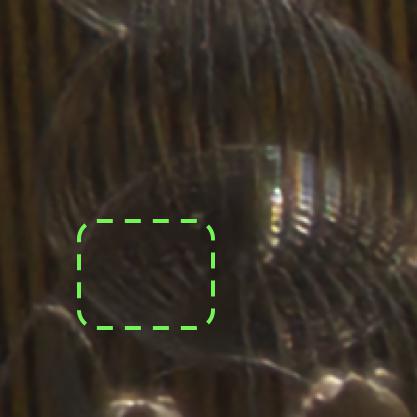} \\
			\hspace{-3mm} 
			\includegraphics[height = \thisheight]{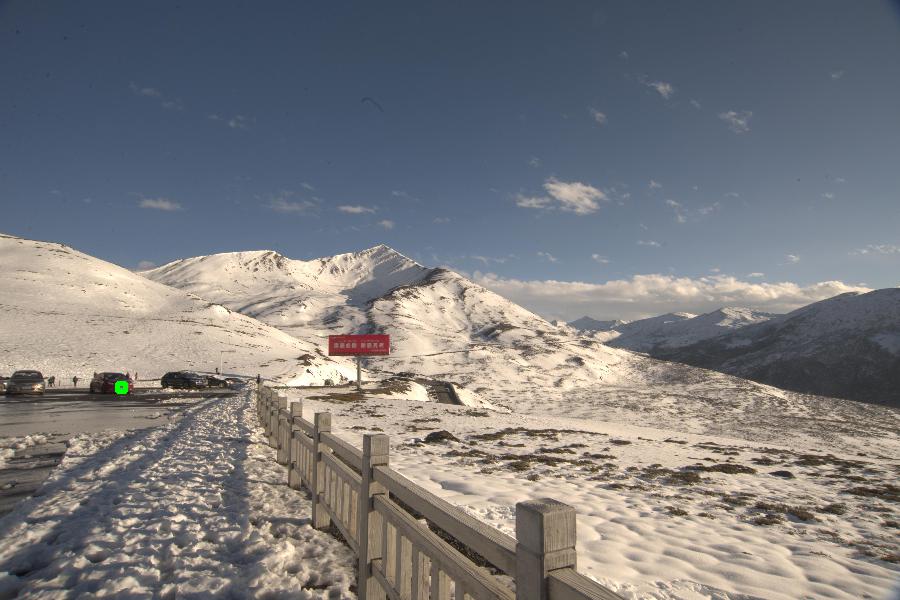} &
			\hspace{-4mm} 
			\includegraphics[height = \thisheight]{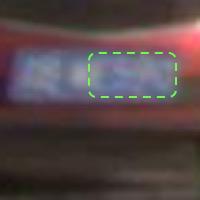} & 
			\hspace{-4mm} 
			\includegraphics[height = \thisheight]{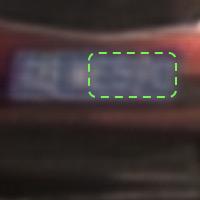} & 
			\hspace{-4mm}
			\includegraphics[height = \thisheight]{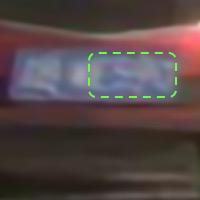} & 
			\hspace{-4mm}
			\includegraphics[height = \thisheight]{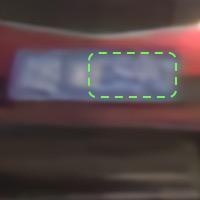} & 
			\hspace{-4mm}
			\includegraphics[height = \thisheight]{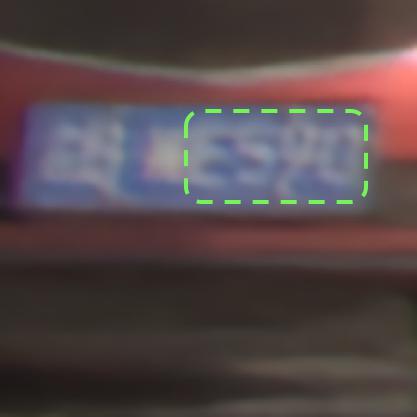} \\
			\hspace{-3mm}
			(a) Whole input image &
			\hspace{-4mm}
			(b) Input &
			\hspace{-4mm}
			(c) SID &
			\hspace{-4mm}
			(d) SRDenseNet &
			\hspace{-4mm}
			(e) RDN &
			\hspace{-4mm}
			(f) Ours \\
		\end{tabular}
	\end{center}
	\vspace{-2mm}
	\caption{Comparison with the state-of-the-arts on real-captured images.
		Since the outputs are of ultra-high resolution, spanning from 6048$\times$8064 to 12416$\times$17472, we only show image patches cropped from the tiny green boxes in (a).
		The input images from top to bottom are captured by Sony, Canon, iPhone 6s Plus, and Nikon cameras, respectively.
	}
	\label{fig:real_1}
	\vspace{-3mm}
\end{figure*}

\vspace{-2mm}
\subsection{Ablation study}
 \label{sec:ablation}
To thoroughly evaluate the proposed algorithm,
we experiment with two variants of our model by removing different components.
We show the quantitative results on the synthetic datasets in Table~\ref{tab:ablation} and provide qualitative comparisons in Figure~\ref{fig:ablation} with detailed explanations.

\vspace{1ex}
\noindent \textbf{Color correction.} As shown in Figure~\ref{fig:ablation}(c),  without the color correction module, the model directly uses the low-resolution raw input for super-resolution, which cannot effectively recover high-fidelity colors for different cameras.
Furthermore, it cannot learn sharp structures during training due to the distractions of the significant color differences similar to the raw-processing methods (\eg SID~\cite{learning_to_see_in_the_dark} and DeepISP~\cite{deepisp_tip}).
\vspace{1ex}
\noindent \textbf{Raw input.} To evaluate the importance of the raw information, we use the low-resolution color image as the input for both modules of the proposed network, where we accordingly change the packing-raw layer of the image restoration module.
Figure~\ref{fig:ablation}(d) shows that the model without raw input cannot generate clear structures or fine details due to the information loss in the ISP.
In contrast, our full model effectively integrates different components, and generates sharper results with more details and fewer artifacts in Figure~\ref{fig:ablation}(e) by exploiting the complementary information in the raw data and processed color images.

\begin{table}[t]
	\caption{\label{tab:real}Quantitative evaluations on the proposed real image dataset with blind image quality evaluation metrics~\cite{Diivine,brisque,nima}. 
	Smaller values indicate better image qualities for DIIVINE~\cite{Diivine} and BRISQUE~\cite{brisque}, and worse image qualities for NIMA~\cite{nima}.
}
	\vspace{-2mm}
	\footnotesize
	\centering
	\begin{tabular}{cccc}
\toprule
		Method & DIIVINE $\downarrow$ & BRISQUE $\downarrow$ & NIMA $\uparrow$ \\
\midrule
		SID~\cite{learning_to_see_in_the_dark} & 55.03 & 48.75 & 4.452 \\
		SRDenseNet~\cite{SR-dense} & 52.94 & 48.13 & 4.72 \\
		RDN~\cite{SR-residual-dense} & 54.80 & 48.39 & 4.91 \\
		Ours & \bf 46.83 & \bf 46.08 & \bf 4.94 \\
\bottomrule
	\end{tabular}
	\vspace{-4mm}
\end{table}

\begin{figure*}[t]
	\newcommand{\thisheight}{0.14\linewidth}
	\footnotesize
	\begin{center}
		\begin{tabular}{cccccc}
			\hspace{-3mm} 
			\includegraphics[height = \thisheight]{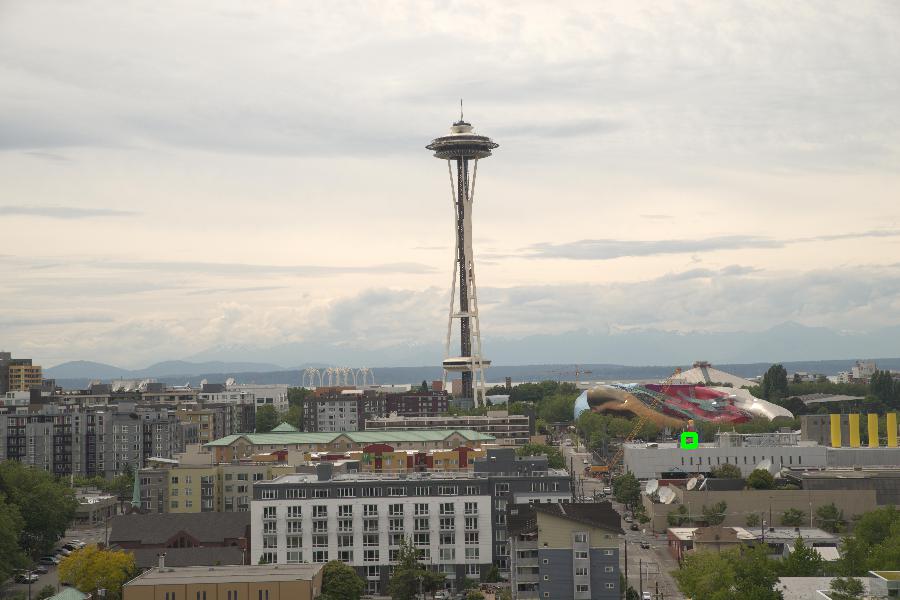} &
			\hspace{-4mm} 
			\includegraphics[height = \thisheight]{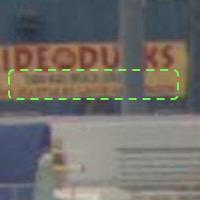} & 
			\hspace{-4mm} 
			\includegraphics[height = \thisheight]{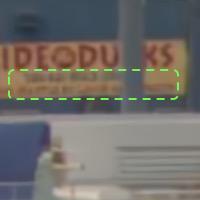} & 
			\hspace{-4mm}
			\includegraphics[height = \thisheight]{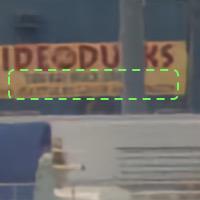} & 
			\hspace{-4mm}
			\includegraphics[height = \thisheight]{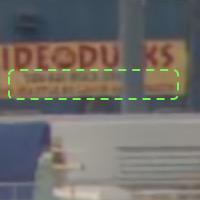} & 
			\hspace{-4mm}
			\includegraphics[height = \thisheight]{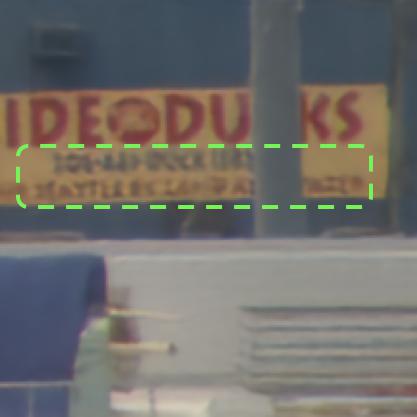} \\
			\hspace{-3mm} 
			\includegraphics[height = \thisheight]{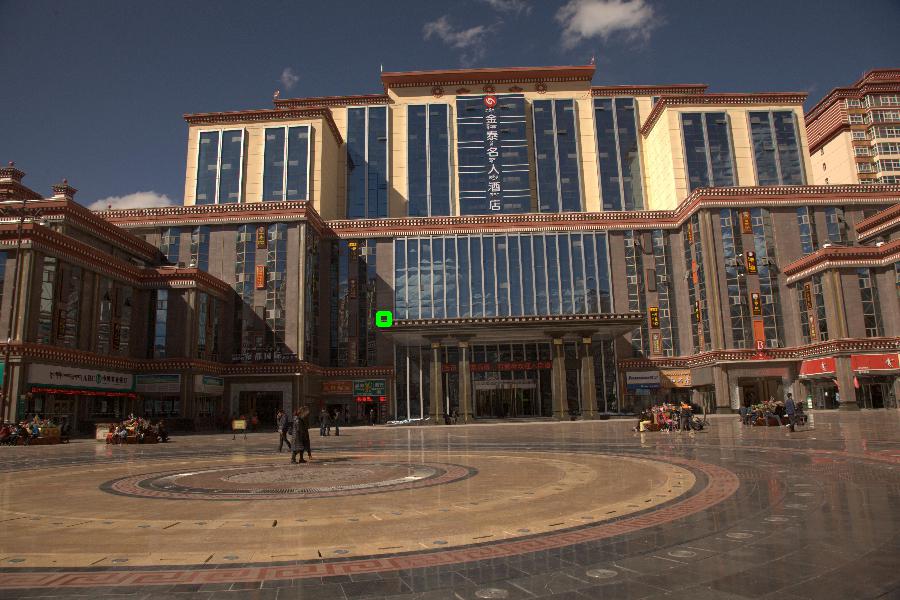} &
			\hspace{-4mm} 
			\includegraphics[height = \thisheight]{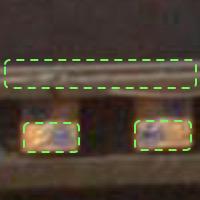} & 
			\hspace{-4mm} 
			\includegraphics[height = \thisheight]{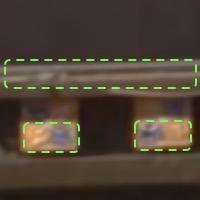} & 
			\hspace{-4mm}
			\includegraphics[height = \thisheight]{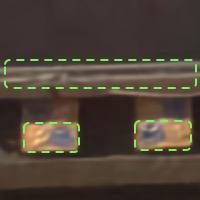} & 
			\hspace{-4mm}
			\includegraphics[height = \thisheight]{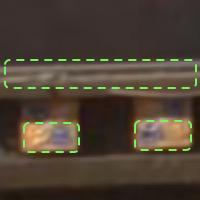} & 
			\hspace{-4mm}
			\includegraphics[height = \thisheight]{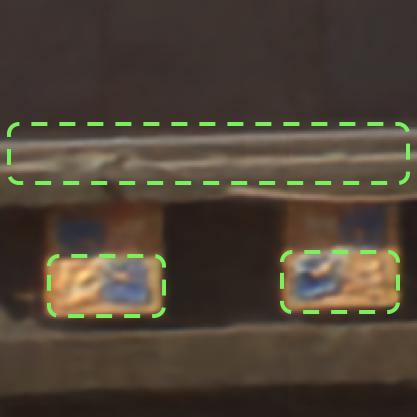} \\ 
			\hspace{-3mm} 
			\includegraphics[height = \thisheight]{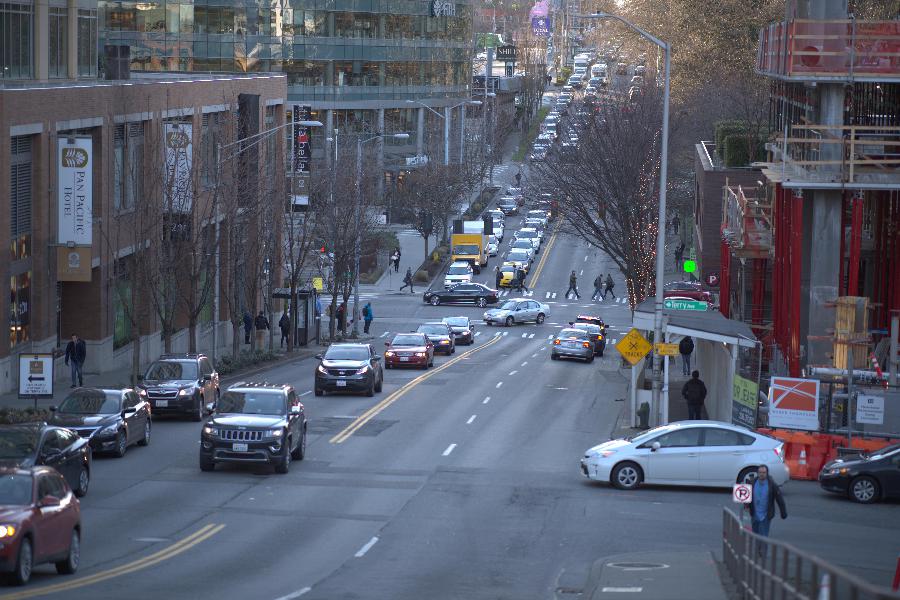} &
			\hspace{-4mm} 
			\includegraphics[height = \thisheight]{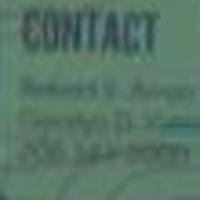} & 
			\hspace{-4mm} 
			\includegraphics[height = \thisheight]{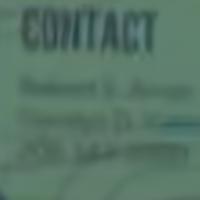} & 
			\hspace{-4mm}
			\includegraphics[height = \thisheight]{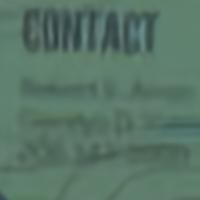} & 
			\hspace{-4mm}
			\includegraphics[height = \thisheight]{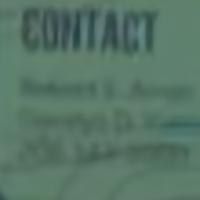} & 
			\hspace{-4mm}
			\includegraphics[height = \thisheight]{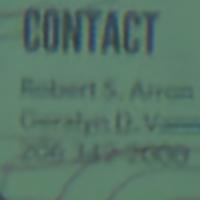} \\
			\hspace{-3mm} 
			\includegraphics[height = \thisheight]{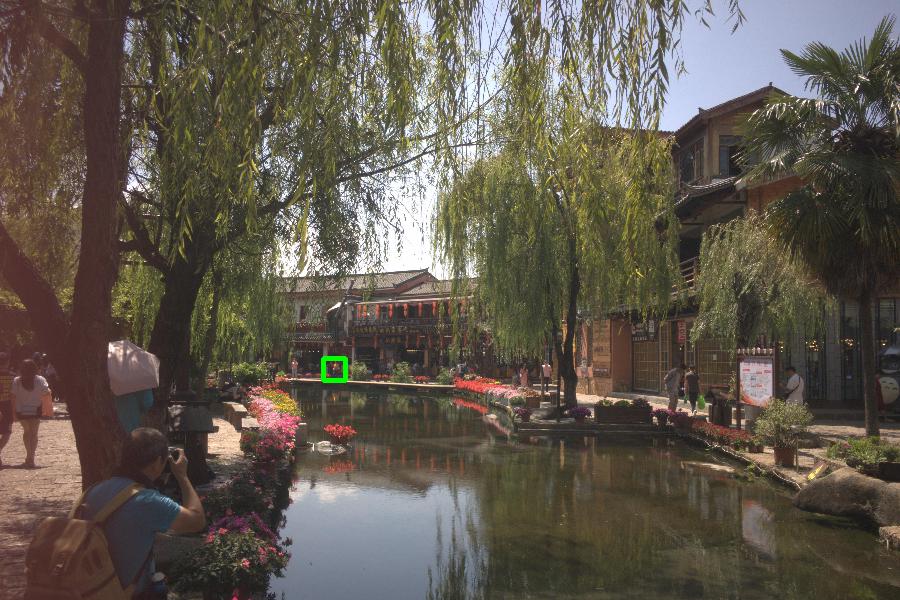} &
			\hspace{-4mm} 
			\includegraphics[height = \thisheight]{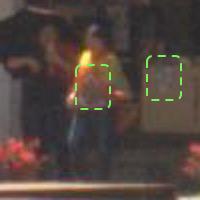} & 
			\hspace{-4mm} 
			\includegraphics[height = \thisheight]{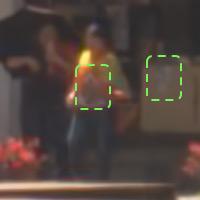} & 
			\hspace{-4mm}
			\includegraphics[height = \thisheight]{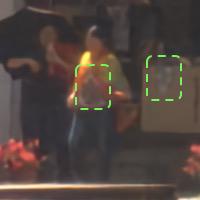} & 
			\hspace{-4mm}
			\includegraphics[height = \thisheight]{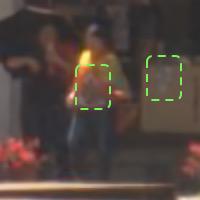} & 
			\hspace{-4mm}
			\includegraphics[height = \thisheight]{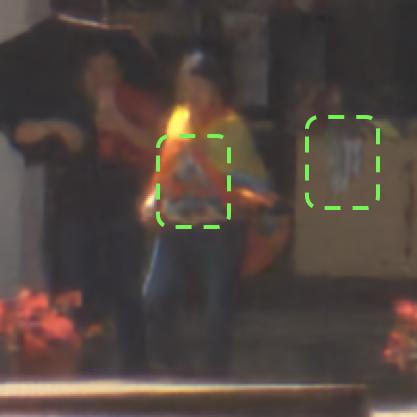} \\
			\hspace{-3mm} \scriptsize
			(a) Whole input image &
			\hspace{-4mm} \scriptsize
			(b) Input &
			\hspace{-4mm} \scriptsize
			(c) RDN with pre. data &
			\hspace{-4mm} \scriptsize
			(d) RDN with our data &
			\hspace{-4mm} \scriptsize
			(e) Ours with pre. data &
			\hspace{-4mm} \scriptsize
			(f) Ours with our data \\
		\end{tabular}
	\end{center}
	\vspace{-2mm}
	\caption{Qualitative evaluations of the proposed data generation pipeline on real images. The images from top to bottom are captured by Pantax, Canon, Nikon, and iPad cameras, respectively.
	}
	\label{fig:real_2}
	\vspace{-4mm}
\end{figure*}

\vspace{-1mm}
\subsection{Evaluation on real images}
\label{sec:real_images}
To evaluate our algorithm on real images, 
we propose a new test dataset (no overlap with the training set)  by collecting diverse real-captured raw data paired with the corresponding color images from the Internet.
There are 100 image pairs captured by different cameras including Nikon, Sony, Canon, Leica, Pentax, Fuji, and  Apple (iPad Pro and iPhone 6s Plus).
The resolution of these images spans from  3024$\times$4032 to 6208$\times$8736.
Several example images from this dataset are shown in Figure~\ref{fig:teaser}(a), \ref{fig:real_1}(a), and \ref{fig:real_2}(a).

We first present qualitative comparisons between the baselines~\cite{learning_to_see_in_the_dark,SR-dense,SR-residual-dense} and our method.
As shown in Figure \ref{fig:real_1}(c)-(e), both the state-of-the-art raw-processing and super-resolution approaches do not perform well on real inputs and generate incorrect colors and oversmoothing artifacts.
In contrast, the proposed algorithm (Figure~\ref{fig:real_1}(f)) generates better results with sharper edges and finer details.
Although the operations within the ISPs for the real images are unknown to our network, the proposed algorithm can effectively recover high-fidelity color appearances with the color correction module.

We also quantitatively evaluate these algorithms on the real image dataset using the blind image quality evaluation metrics: Distortion Identification-based Image Verity and Integrity Evaluation (DIIVINE)~\cite{Diivine}, Blind/Referenceless Image Spatial Quality Evaluator (BRISQUE)~\cite{brisque}, and Neural Image Assessment (NIMA)~\cite{nima}.
As the images are too large (especially after super-resolution), it is computationally infeasible to compute these blind metrics on the whole outputs. 
Thus, we crop 5 patches of  512$\times$512 pixels from the center and four corners of each input image, and use this subset of 500 patches for quantitative evaluation.
Table~\ref{tab:real} shows that our method performs 
favorably against the baseline approaches on all the metrics. 
These results are consistent with the visual comparisons in Figure~\ref{fig:real_1} and further demonstrate the effectiveness of our method.

\begin{table}[t]
	\caption{\label{tab:real_2}Effectiveness of the proposed data generation pipeline. 
The models are trained with different datasets and assessed
using the blind image quality evaluation metrics~\cite{Diivine,brisque,nima}. 
``Our data'' is generated by the proposed pipeline in Section~\ref{sec:training_data}, whereas ``previous data'' is generated by bicubic downsampling, homoscedastic Gaussian noise and nonlinear space mosaicing~\cite{SRCNN,xxy-iccv17,demosaicking_denoising_deep}.
``Our data$^*$'' represents a variant of our data generation pipeline by performing the realistic image degradations in the nonlinear color image space.
	Smaller values indicate better image qualities  for  DIIVINE~\cite{Diivine} and BRISQUE~\cite{brisque}, and worse image qualities for NIMA~\cite{nima}.
	}
	\vspace{-2mm}
	\footnotesize
	\centering
	\begin{tabular}{@{}c@{}ccc}
\toprule
		Method & DIIVINE $\downarrow$  & BRISQUE $\downarrow$ & NIMA $\uparrow$ \\
\midrule
		RDN w/ previous data & 59.11 & 50.26 & 4.74\\
		RDN w/ our data & 54.80 & 48.39 & 4.91 \\
		Our model w/ previous data & 59.01 & 48.23 & 4.73 \\
		Our model w/ our data$^*$ & 47.51 & 47.09 & 4.91 \\
		Our model w/ our data & \bf 46.83 & \bf 46.08 & \bf 4.94 \\
\bottomrule
	\end{tabular}
	\vspace{-4mm}
\end{table}

\begin{figure}[t]
	\footnotesize
	\begin{center}
		\begin{tabular}{c}
			\includegraphics[width=0.7\linewidth]{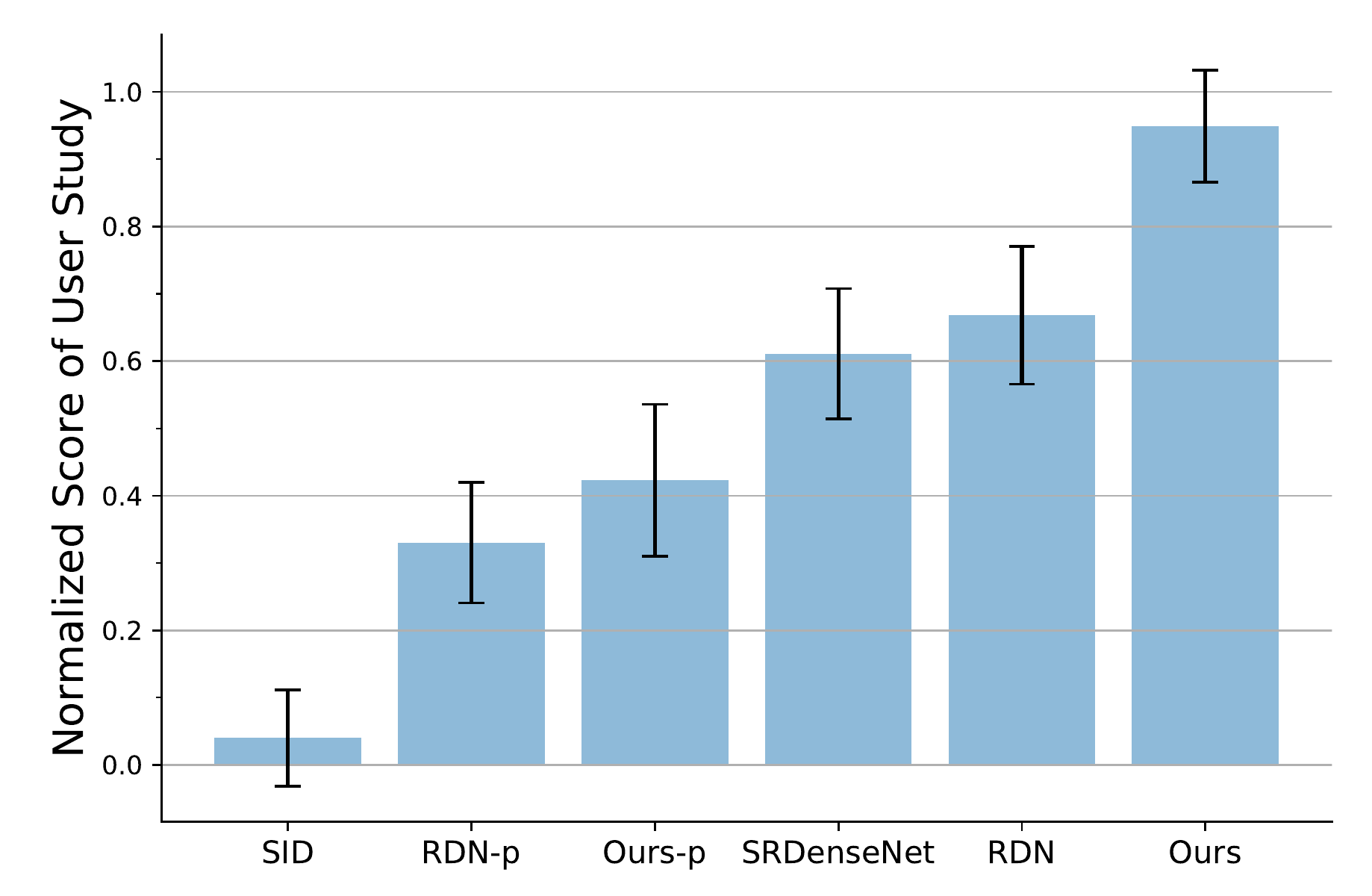}\\
		\end{tabular}
	\end{center}
	\vspace{-2mm}
	\caption{User study of different super-resolution algorithms on real images. Each bar shows the mean and standard deviation of the normalized scores of each method. ``RDN-p'' and ``Ours-p'' represents the RDN~\cite{SR-residual-dense} and our network trained on previous data, respectively.}
	\label{fig:user_study}
	\vspace{-4mm}
\end{figure}

\begin{figure*}[t]
	\newcommand{\thisheight}{0.14\linewidth}
	\footnotesize
	\begin{center}
		\begin{tabular}{cccccc}
			\hspace{-3mm} 
			\includegraphics[height = \thisheight]{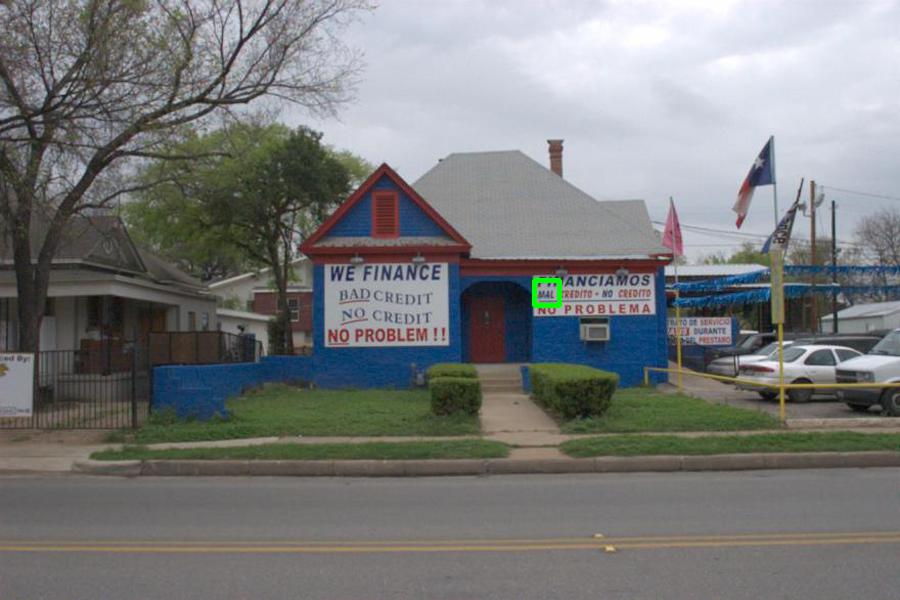} &
			\hspace{-4mm} 
			\includegraphics[height = \thisheight]{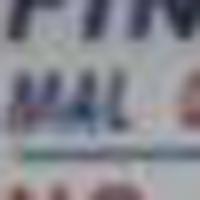} & 
			\hspace{-4mm} 
			\includegraphics[height = \thisheight]{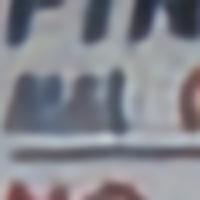} & 
			\hspace{-4mm}
			\includegraphics[height = \thisheight]{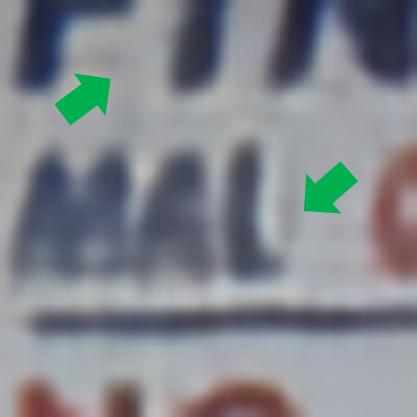} & 
			\hspace{-4mm}
			\includegraphics[height = \thisheight]{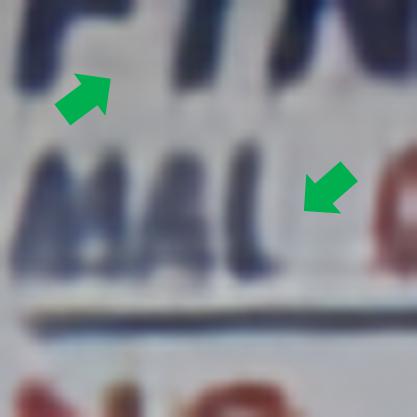} & 
			\hspace{-4mm}
			\includegraphics[height = \thisheight]{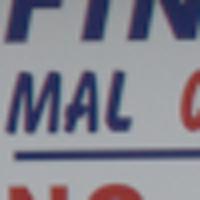} \\
			\hspace{-3mm} 
			(a) Whole input image &
			\hspace{-4mm} 
			(b) Input &
			\hspace{-4mm} 
			(c) w/o ED &
			\hspace{-4mm} 
			(d) w/o CA &
			\hspace{-4mm} 
			(e) Ours &
			\hspace{-4mm} 
			(f) GT \\
		\end{tabular}
	\end{center}
	\vspace{-2mm}
	\caption{Qualitative evaluations of the image restoration module. ``ED'' and ``CA'' represent the encoder-decoder structure and channel attention, respectively. Note the blurry shadows in the blank regions of (d).
	}
	\label{fig:restoration}
	\vspace{-4mm}
\end{figure*}

\begin{figure*}[t]
	\newcommand{\thisheight}{0.11\linewidth}
	\footnotesize
	\begin{center}
		\begin{tabular}{cccccccc}
			\hspace{-3mm} 
			\includegraphics[height = \thisheight]{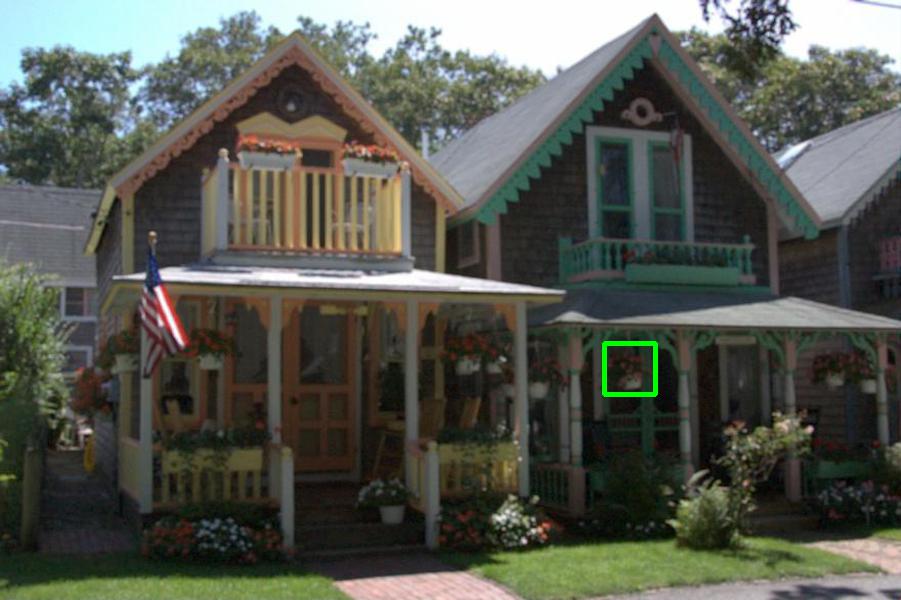} &
			\hspace{-4mm} 
			\includegraphics[height = \thisheight]{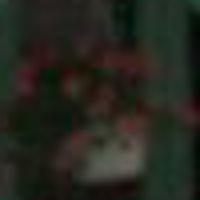} & 
			\hspace{-4mm} 
			\includegraphics[height = \thisheight]{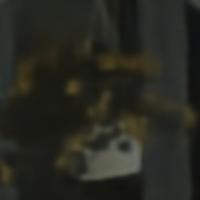} & 
			\hspace{-4mm}
			\includegraphics[height = \thisheight]{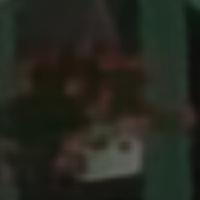} & 
			\hspace{-4mm}
			\includegraphics[height = \thisheight]{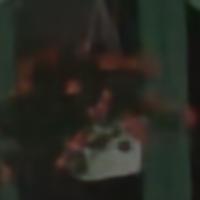} & 
			\hspace{-4mm}
			\includegraphics[height = \thisheight]{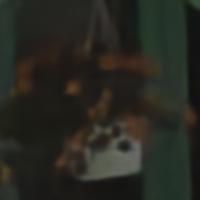} &
			\hspace{-4mm}
			\includegraphics[height = \thisheight]{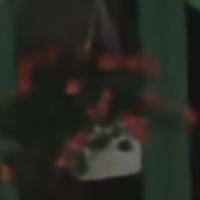} &  
			\hspace{-4mm}
			\includegraphics[height = \thisheight]{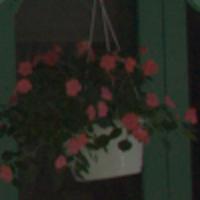} \\			
			\hspace{-3mm} \scriptsize
			(a) Whole input image &
			\hspace{-4mm} \scriptsize
			(b) Input &
			\hspace{-4mm} \scriptsize
			(c) Global &
			\hspace{-4mm} \scriptsize
			(d) GF &
			\hspace{-4mm} \scriptsize
			(e) ETGF &
			\hspace{-4mm} \scriptsize
			(f) w/o FF &
			\hspace{-4mm} \scriptsize
			(g) Ours &
			\hspace{-4mm} \scriptsize
			(h) GT \\
		\end{tabular}
	\end{center}
	\vspace{-2mm}
\caption{Visual comparison of different methods for the color correction module. (c) ``Global'' represents the global color correction~\cite{deepisp_tip}. (d) ``GF'' indicates the guided filtering~\cite{guidedFilter}. (f) ``w/o FF'' is our learning-based guided filter without using feature fusion. 
	}
	\label{fig:guided}
	\vspace{-4mm}
\end{figure*} 

\vspace{-1mm}
\section{Analysis and Extension}
In this section, we demonstrate the effectiveness of our data generation pipeline, and the image restoration as well as color correction modules. 
We then present the results of a user study for more comprehensive evaluation.
Further, we show that our model can be well extended to higher upsampling factors.
Finally, we show that the proposed network is able to generalize to more complex color corrections.

\vspace{-1mm}
\subsection{Effectiveness of the data generation pipeline}
We use the real image dataset to evaluate our data generation method.
As shown in Figure~\ref{fig:teaser}(d) and \ref{fig:real_2}(d),
the results of RDN become sharper after re-training with the data generated by our pipeline.
On the other hand, 
the proposed two-branch CNN model cannot generate clear images (Figure~\ref{fig:real_2}(e)) when trained on previous data synthesized by bicubic downsampling, homoscedastic Gaussian noise and nonlinear space mosaicing~\cite{SRCNN,xxy-iccv17,demosaicking_denoising_deep}.
In contrast, our method generates images with sharper edges and finer details by using the training data from our data generation pipeline as shown in Figure~\ref{fig:teaser}(e) and \ref{fig:real_2}(f). 

We present quantitative results in Table~\ref{tab:real_2} with the blind image quality metrics~\cite{Diivine,brisque,nima}. 
Both RDN~\cite{SR-residual-dense} and the proposed model obtain significant improvements in terms of DIIVINE~\cite{Diivine}, BRISQUE~\cite{brisque}, and NIMA~\cite{nima} by training the networks on our synthetic dataset, which shows the effectiveness of the proposed data generation approach.
\rev{We also present a variant of our data generation pipeline (``our data$^*$'') by performing the realistic image degradation operations in the nonlinear color image space, which leads to inferior results than our full data generation pipeline and further demonstrates the importance of the raw-space data synthesis scheme.}

\begin{table}[t]
	\caption{\label{tab:restoration} Quantitative evaluation of the image restoration module network.
	 ``ED'' and ``CA'' represent the encoder-decoder structure and channel attention, respectively.
	}
	\vspace{-2mm}
	\footnotesize
	\centering
	\begin{tabular}{ccccc}
\toprule
		\multicolumn{1}{c}{\multirow{2}*{Methods}}&\multicolumn{2}{c}{Blind}&\multicolumn{2}{c}{Non-blind}\\
\cmidrule{2-5}
		& PSNR & SSIM & PSNR & SSIM \\    %
\midrule
		w/o ED & 30.68 & 0.8016 & 31.31 & 0.8202 \\
		w/o CA & 30.90 & 0.8065 & 31.94 & 0.8294 \\
		Ours & \bf 31.06 & \bf 0.8088 & \bf 32.07 & \bf 0.8314  \\
\bottomrule
	\end{tabular}
		\vspace{-4mm}
\end{table}

\begin{table}[t]
	\caption{\label{tab:guided filter} Quantitative evaluation of the proposed learning-based guided filter.
	}
	\vspace{-2mm}
	\footnotesize
	\centering
	\begin{tabular}{ccccc}
\toprule
		\multicolumn{1}{c}{\multirow{2}*{Methods}}&\multicolumn{2}{c}{Blind}&\multicolumn{2}{c}{Non-blind}\\
\cmidrule{2-5}
		& PSNR & SSIM & PSNR & SSIM \\    %
\midrule
		Guided filtering~\cite{guidedFilter} & 29.86 & 0.7816 & 29.99 & 0.7864 \\
		Global color correction~\cite{deepisp_tip} & 30.18 & 0.8013 & 30.72 & 0.8161 \\
		ETGF~\cite{wu2018fast} & 30.56 & 0.8026 & 31.65 & 0.8235 \\
		w/o feature fusion & 30.82 & 0.8049 & 31.85 & 0.8278  \\
		Ours & \bf 31.06 & \bf 0.8088 & \bf 32.07 & \bf 0.8314  \\
\bottomrule
	\end{tabular}
		\vspace{-4mm}
\end{table}

\vspace{-2mm}
\subsection{User study}\label{sec:user_study}
We conduct a user study for more comprehensive evaluation of the super-resolution algorithms.
This study uses 10 images randomly selected from the real image dataset, and each low-resolution input is upsampled by 6 different methods: SID~\cite{learning_to_see_in_the_dark}, SRDenseNet~\cite{SR-dense}, RDN with previous data~\cite{SR-residual-dense}, RDN with our data, our model with previous data, and our model with our data. 
Twenty subjects are asked to rank the reconstructed images by each method with the input image as reference (1 for poor and 6 for excellent).
We normalize the rank values to $[0, 1]$, and use the normalized scores for measuring image quality in Figure~\ref{fig:user_study}, where we visualize the mean scores and the standard deviation.
The user study shows that the proposed method can generate results with high perceptual image quality for real-captured inputs.
In addition, our data generation pipeline is effective in synthesizing realistic training data and significantly improves the performance of both RDN and the proposed network.

\begin{figure*}[t]
	\newcommand{\thisheight}{0.13\linewidth}
	\footnotesize
	\begin{center}
		\begin{tabular}{ccccccc}
			\hspace{-3mm} 
			\includegraphics[height = \thisheight]{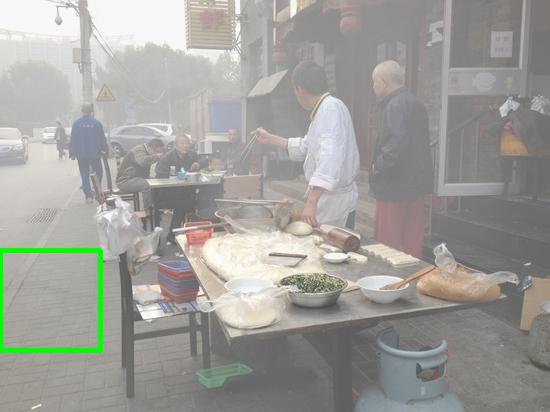} &
			\hspace{-4mm} 
			\includegraphics[height = \thisheight]{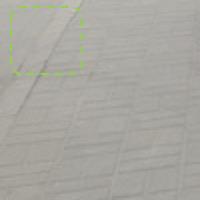} & 
			\hspace{-4mm} 
			\includegraphics[height = \thisheight]{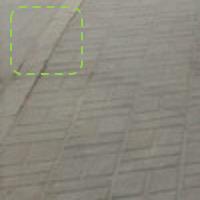} & 
			\hspace{-4mm}
			\includegraphics[height = \thisheight]{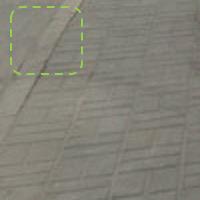} & 
			\hspace{-4mm}
			\includegraphics[height = \thisheight]{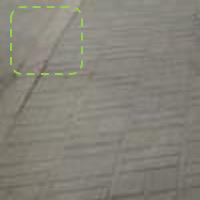} & 
			\hspace{-4mm}
			\includegraphics[height = \thisheight]{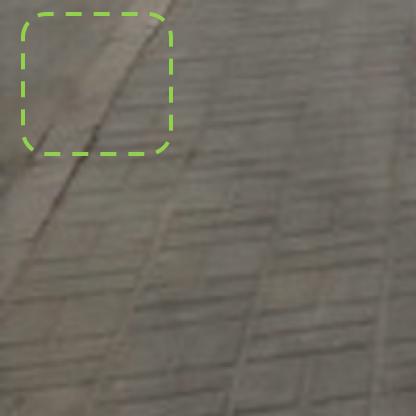} & 
			\hspace{-4mm}
			\includegraphics[height = \thisheight]{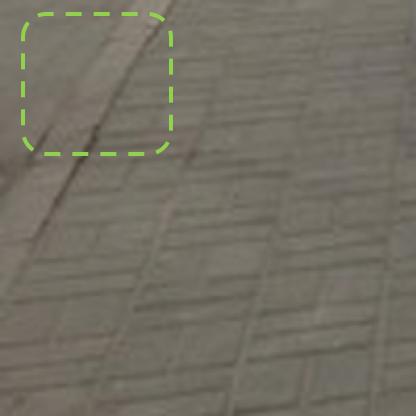} \\
			\hspace{-3mm} 
			\includegraphics[height = \thisheight]{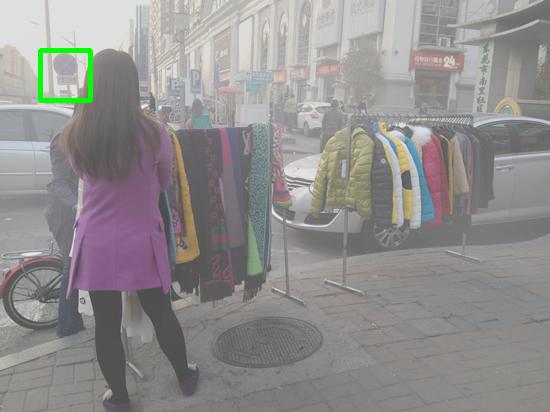} &
			\hspace{-4mm} 
			\includegraphics[height = \thisheight]{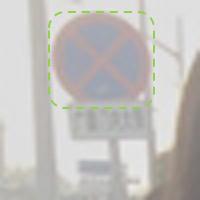} & 
			\hspace{-4mm} 
			\includegraphics[height = \thisheight]{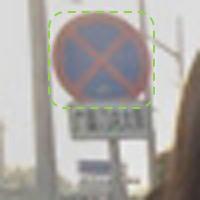} & 
			\hspace{-4mm}
			\includegraphics[height = \thisheight]{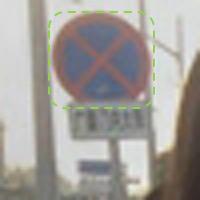} & 
			\hspace{-4mm}
			\includegraphics[height = \thisheight]{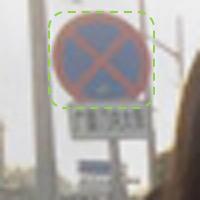} & 
			\hspace{-4mm}
			\includegraphics[height = \thisheight]{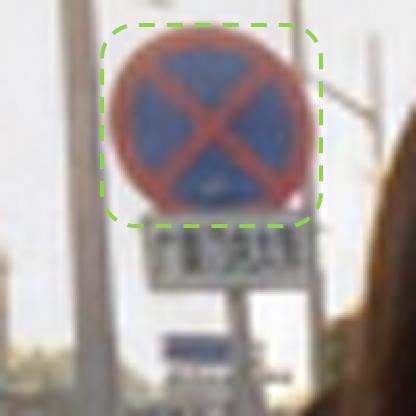} & 
			\hspace{-4mm}
			\includegraphics[height = \thisheight]{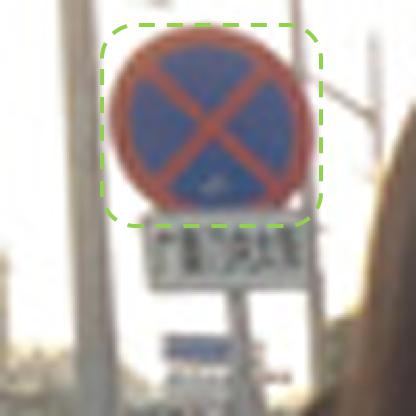} \\
			\hspace{-3mm}
			(a) Whole input image &
			\hspace{-4mm}
			(b) Input &
			\hspace{-4mm}
			(c) DehazeNet &
			\hspace{-4mm}
			(d) GFN &
			\hspace{-4mm}
			(e) GridDehazeNet &
			\hspace{-4mm}
			(f) Ours &
			\hspace{-4mm}
			(g) GT \\
		\end{tabular}
	\end{center}
	\vspace{-3mm}
\caption{Qualitative evaluation for image dehazing. Compared with the existing dehazing algorithms~\cite{cai2016dehazenet,ren2018gated,liu2019griddehazenet}, our method can better remove haze from the input and generate fewer artifacts.
	}
	\label{fig:dehazing}
	\vspace{-3mm}
\end{figure*}

\begin{figure*}[t]
	\newcommand{\thisheight}{0.102\linewidth}
	\footnotesize
	\begin{center}
		\begin{tabular}{ccccccccc}
			\hspace{-3mm} 
			\includegraphics[height = \thisheight]{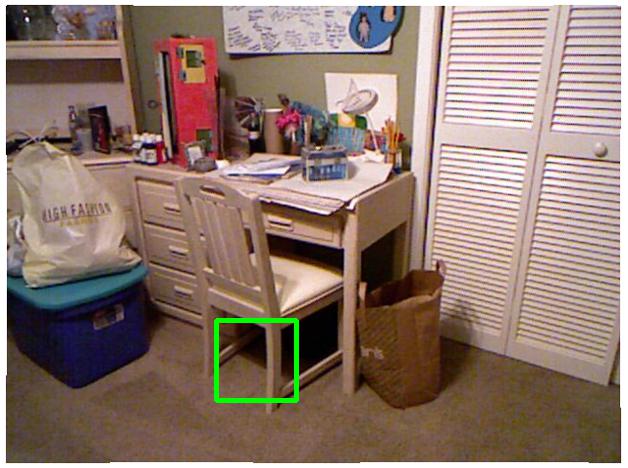} &
			\hspace{-4mm} 
			\includegraphics[height = \thisheight]{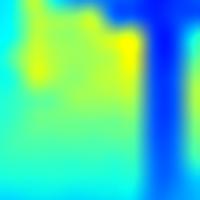} & 
			\hspace{-4mm} 
			\includegraphics[height = \thisheight]{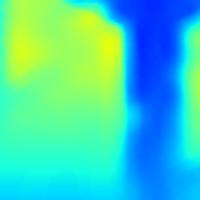} & 
			\hspace{-4mm}
			\includegraphics[height = \thisheight]{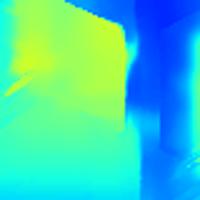} & 
			\hspace{-4mm}
			\includegraphics[height = \thisheight]{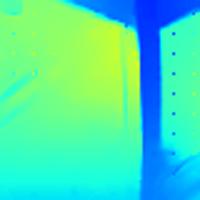} & 
			\hspace{-4mm}
			\includegraphics[height = \thisheight]{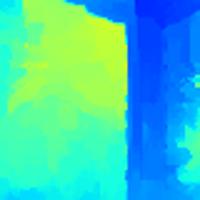} & 
			\hspace{-4mm}
			\includegraphics[height = \thisheight]{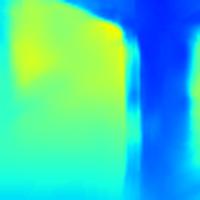} & 
			\hspace{-4mm}
			\includegraphics[height = \thisheight]{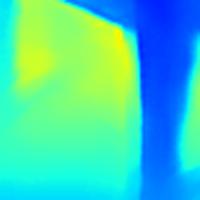} & 
			\hspace{-4mm}
			\includegraphics[height = \thisheight]{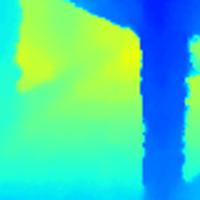}\\
			\hspace{-3mm} 
			\includegraphics[height = \thisheight]{fig/exp/depth_upsample/x8/1074/full_input_rgb} &
			\hspace{-4mm} 
			\includegraphics[height = \thisheight]{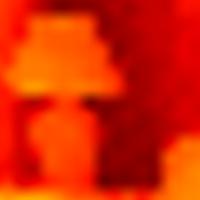} & 
			\hspace{-4mm} 
			\includegraphics[height = \thisheight]{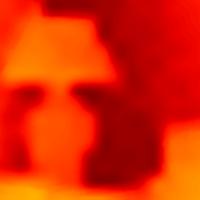} & 
			\hspace{-4mm}
			\includegraphics[height = \thisheight]{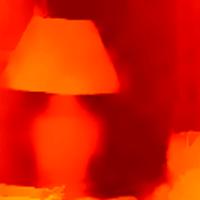} & 
			\hspace{-4mm}
			\includegraphics[height = \thisheight]{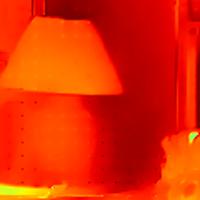} & 
			\hspace{-4mm}
			\includegraphics[height = \thisheight]{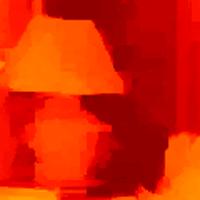} & 
			\hspace{-4mm}
			\includegraphics[height = \thisheight]{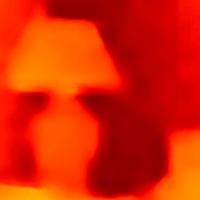} & 
			\hspace{-4mm}
			\includegraphics[height = \thisheight]{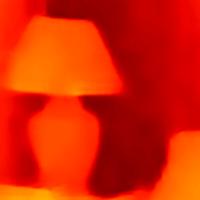} & 
			\hspace{-4mm}
			\includegraphics[height = \thisheight]{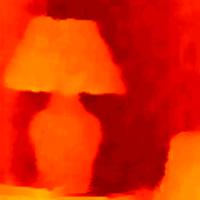}\\
			\hspace{-3mm}
			(a) Guidance &
			\hspace{-4mm}
			(b) Input &
			(c) GF &
			\hspace{-4mm}
			(d) JBU &
			\hspace{-4mm}
			(e) TGV &
			\hspace{-4mm}
			(f) Park &
			\hspace{-4mm}
			(g) DJF &
			\hspace{-4mm}
			(h) Ours &
			\hspace{-4mm}
			(i) GT
			\\
		\end{tabular}
	\end{center}
	\vspace{-3mm}
\caption{Qualitative evaluation for guided depth upsampling. We compare our method against existing algorithms~\cite{guidedFilter,kopf2007joint,ferstl2013image,park2011high,li2019joint} for a scaling factor of 8.
	}
	\label{fig:depth_sr}
	\vspace{-3mm}
\end{figure*}

\vspace{-2mm}
\subsection{Effectiveness of the image restoration module}
\label{sec:effect_restoration}

The proposed image restoration module differs from the existing  networks~\cite{SRCNN,SR-VDSR,SR-dense,SR-residual-dense} in two aspects.
First, while existing super-resolution algorithms often learn the nonlinear mapping functions at the same scale~\cite{SRCNN,SR-VDSR,SR-dense,SR-residual-dense}, we use the deep convolutional blocks in an encoder-decoder manner similar to the U-net~\cite{ronneberger2015u}. 
As shown in Figure~\ref{fig:restoration}(c), simply applying the blocks at one scale cannot effectively exploit multi-scale features and may lead to lower-quality results.
Second, we introduce the channel attention mechanism to improve the dense blocks \cite{densenet,SR-dense} by recalibrating the naively concatenated features. 
The proposed DCA blocks can increase the representational capacity of the image restoration model and facilitate  learning better nonlinear mapping functions.
As shown in Figure~\ref{fig:restoration}, using the DCA block effectively reduces artifacts of the result produced by the model without channel attention (Figure~\ref{fig:restoration}(d)) and generates a clearer image in Figure~\ref{fig:restoration}(e).
In addition, we also present quantitative evaluation results in Table~\ref{tab:restoration}. Both the encoder-decoder and channel attention strategies effectively improve the super-resolution performance.
\vspace{1ex}
{\noindent \bf{Image dehazing.}} %
The proposed image restoration module can also be applied to other image processing tasks, such as image dehazing.
As the input and output of the dehazing task have the same image resolution, we remove the packing raw and sub-pixel layers in Figure~\ref{fig:branch1}.
Similar to the GridDehazeNet~\cite{liu2019griddehazenet}, we train and test the proposed network using the RESIDE benchmark dataset~\cite{reside}.
Table~\ref{tab:dehazing} shows that our method performs favorably against the state-of-the-art dehazing algorithms~\cite{he2010single,cai2016dehazenet,ren2016single,li2017aod,ren2018gated,liu2019griddehazenet}.
We present some visual examples from the RESIDE test set in Figure~\ref{fig:dehazing} for qualitative comparisons.
Our image restoration module can better remove thick haze from the input and produce fewer artifacts as shown in Figure~\ref{fig:dehazing}(f).

\begin{table}[t]
	\caption{\label{tab:dehazing}Image dehazing results on the test sets of RESIDE~\cite{reside}. 
	Numbers in bold indicate the best performance, and underscored numbers indicate the second best. 
	}
	\vspace{-2mm}
	\footnotesize
	\centering
	\begin{tabular}{ccccc}
\toprule
		\multicolumn{1}{c}{\multirow{2}*{Methods}}&\multicolumn{2}{c}{Indoor}&\multicolumn{2}{c}{Outdoor}\\
\cmidrule{2-5}
		& PSNR & SSIM & PSNR & SSIM \\    %
\midrule
		DCP~\cite{he2010single} & 16.61 & 0.8546 & 19.14 & 0.8605  \\
		DehazeNet~\cite{cai2016dehazenet}& 19.82 & 0.8209 & 24.75 & 0.9269  \\
		MSCNN~\cite{ren2016single}& 19.84 & 0.8327 & 22.06 & 0.9078  \\
		AOD-Net~\cite{li2017aod} & 20.51 & 0.8162 & 24.14 & 0.9198  \\
		GFN~\cite{ren2018gated} & 24.91 & 0.9186 & 28.29 & 0.9621  \\
		GridDehazeNet~\cite{liu2019griddehazenet}& \bf 32.16 & \bf 0.9836 & \underline{30.86} & \underline{0.9819}  \\
		Ours & \underline{31.91} & \underline{0.9778} & \bf 33.05 & \bf 0.9834 \\
\bottomrule
	\end{tabular}
		\vspace{-4mm}
\end{table}

\begin{figure*}[t]
	\newcommand{\thisheight}{0.13\linewidth}
	\footnotesize
	\begin{center}
		\begin{tabular}{ccccccc}
			\hspace{-3mm} 
			\includegraphics[height = \thisheight]{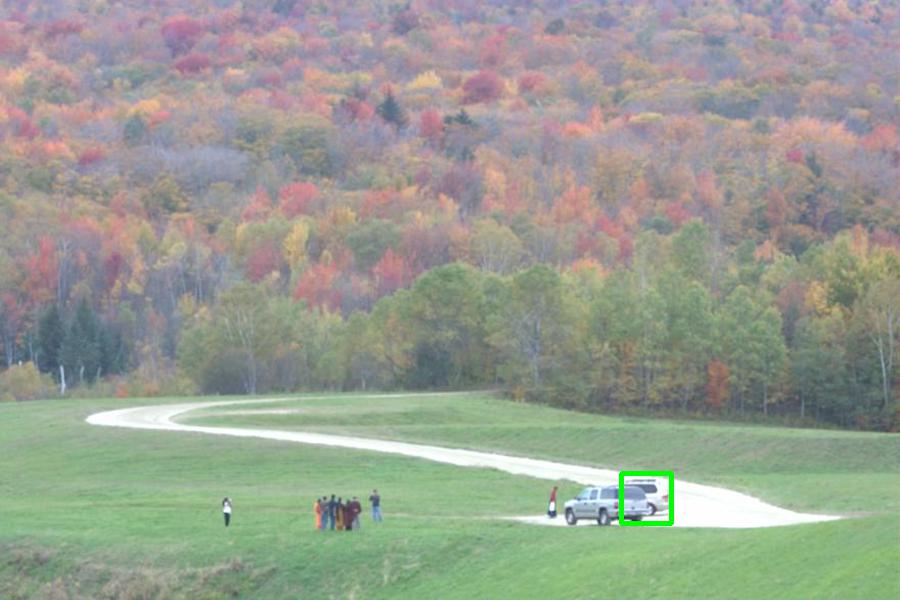} &
			\hspace{-4mm} 
			\includegraphics[height = \thisheight]{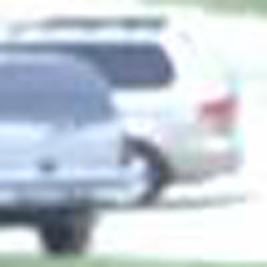} & 
			\hspace{-4mm} 
			\includegraphics[height = \thisheight]{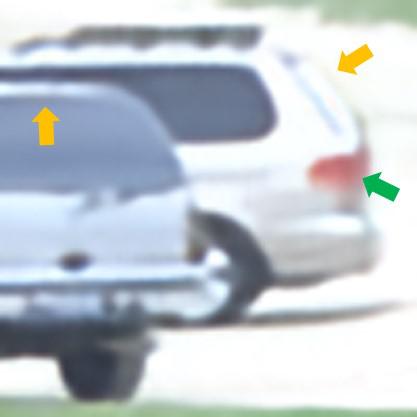} & 
			\hspace{-4mm}
			\includegraphics[height = \thisheight]{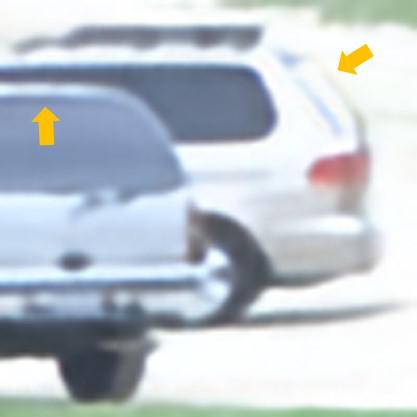} & 
			\hspace{-4mm}
			\includegraphics[height = \thisheight]{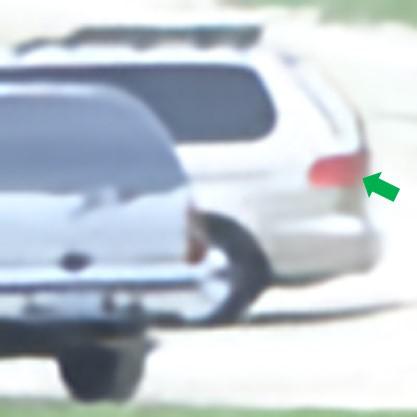} & 
			\hspace{-4mm}
			\includegraphics[height = \thisheight]{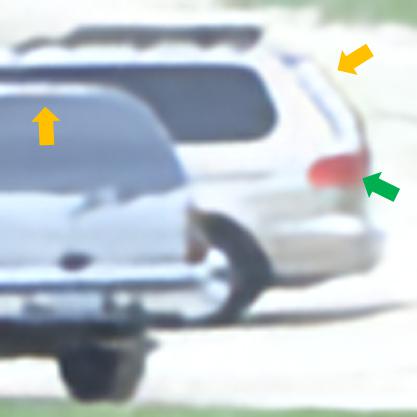} & 
			\hspace{-4mm}
			\includegraphics[height = \thisheight]{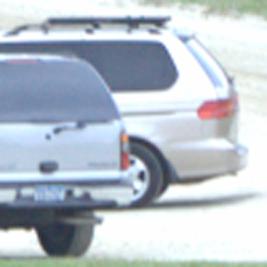} \\
			\hspace{-3mm}
			(a) Whole input image &
			\hspace{-4mm}
			(b) Input &
			\hspace{-4mm}
			(c) dense + sFF &
			\hspace{-4mm}
			(d) DCA + sFF &
			\hspace{-4mm}
			(e) dense + FF &
			\hspace{-4mm}
			(f) DCA + FF &
			\hspace{-4mm}
			(g) GT \\
		\end{tabular}
	\end{center}
	\vspace{-4mm}
\caption{Visual comparison against the model of \cite{rawSR}. ``dense'' denotes the dense block. ``sFF'' and ``FF'' respectively represent the single-scale and multi-scale feature fusion. 
The proposed DCA block can reduce artifacts better than the dense blocks and achieve clearer edges (yellow arrow). 
The multi-scale feature fusion can generate more vivid colors closer to the ground truth (green arrow). Note that (c) is the model of \cite{rawSR}, and (f) is the full model of this work. 
	}
	\label{fig:compare_cvpr}
	\vspace{-4mm}
\end{figure*}

\begin{figure*}[t]
	\newcommand{\thisheight}{0.13\linewidth}
	\footnotesize
	\begin{center}
		\begin{tabular}{ccccccc}
			\hspace{-3mm} 
			\includegraphics[height = \thisheight]{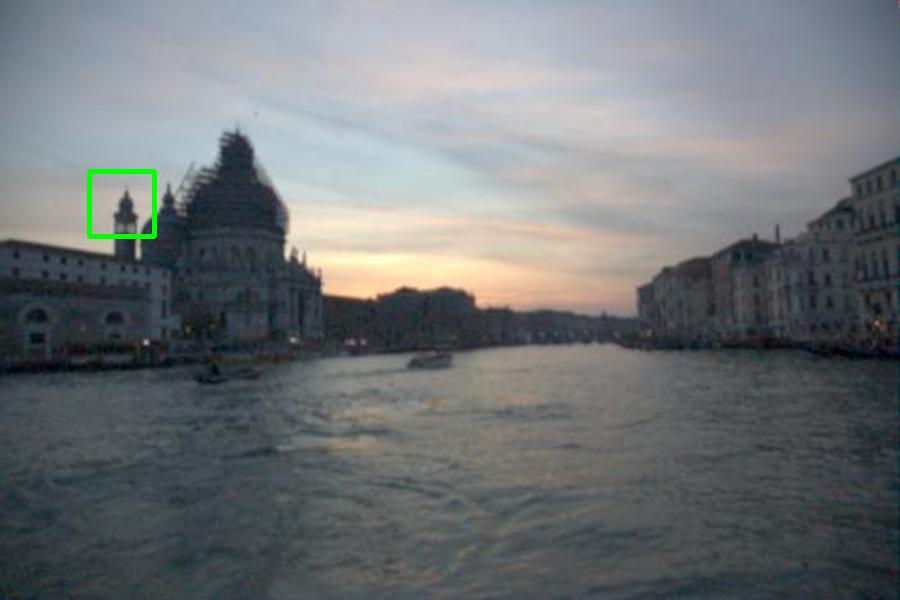} &
			\hspace{-4mm} 
			\includegraphics[height = \thisheight]{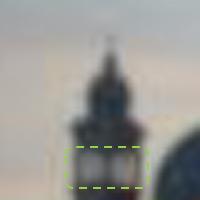} & 
			\hspace{-4mm} 
			\includegraphics[height = \thisheight]{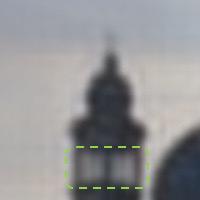} & 
			\hspace{-4mm}
			\includegraphics[height = \thisheight]{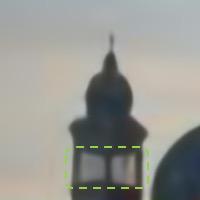} & 
			\hspace{-4mm}
			\includegraphics[height = \thisheight]{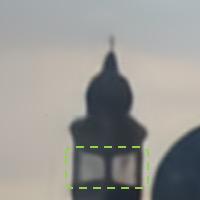} & 
			\hspace{-4mm}
			\includegraphics[height = \thisheight]{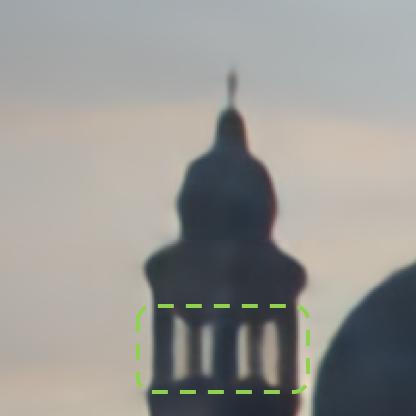} & 
			\hspace{-4mm}
			\includegraphics[height = \thisheight]{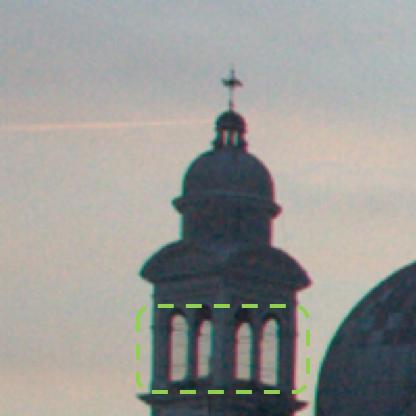} \\
			\hspace{-3mm} 
			\includegraphics[height = \thisheight]{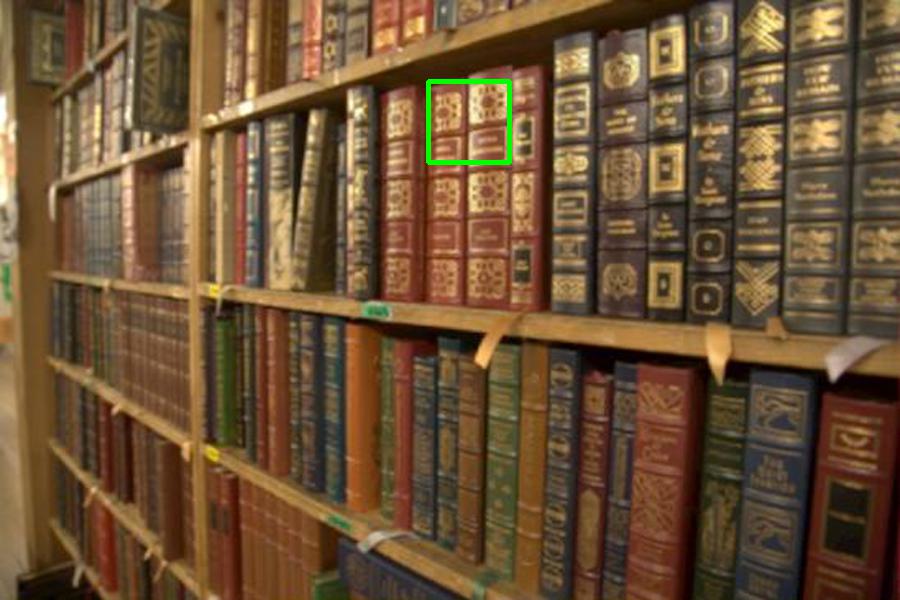} &
			\hspace{-4mm} 
			\includegraphics[height = \thisheight]{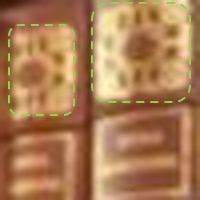} & 
			\hspace{-4mm} 
			\includegraphics[height = \thisheight]{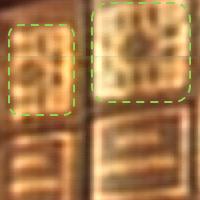} & 
			\hspace{-4mm}
			\includegraphics[height = \thisheight]{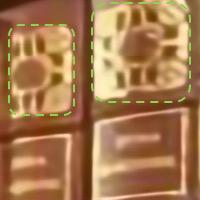} & 
			\hspace{-4mm}
			\includegraphics[height = \thisheight]{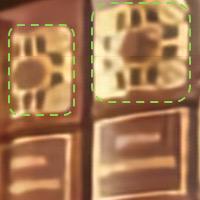} & 
			\hspace{-4mm}
			\includegraphics[height = \thisheight]{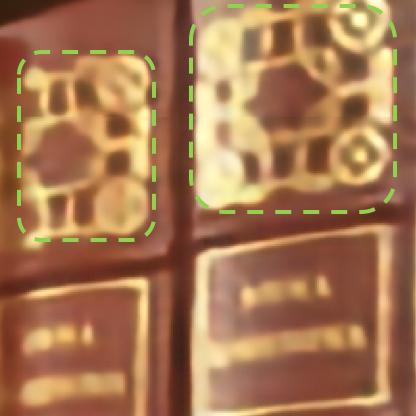} & 
			\hspace{-4mm}
			\includegraphics[height = \thisheight]{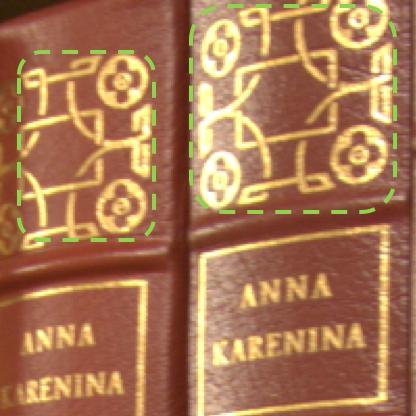} \\
			\hspace{-3mm}
			(a) Whole input image &
			\hspace{-4mm}
			(b) Input &
			\hspace{-4mm}
			(c) SID &
			\hspace{-4mm}
			(d) SRDenseNet &
			\hspace{-4mm}
			(e) RDN &
			\hspace{-4mm}
			(f) Ours &
			\hspace{-4mm}
			(g) GT \\
		\end{tabular}
	\end{center}
	\vspace{-2mm}
\caption{Qualitative evaluation of the proposed method for 4$\times$ image super-resolution. The first and second examples are from the blind and non-blind image datasets, respectively. 
	}
	\label{fig:x4}
	\vspace{-4mm}
\end{figure*}

\vspace{-2mm}
\subsection{Effectiveness of the color correction module}
\label{sec:effect_color}
As introduced in Section~\ref{sec:color_correction}, we propose a learning-based guided filter for the color correction module, which helps address the oversmoothing issue of the guided filter~\cite{guidedFilter}.
There are two existing approaches closely related to our work, \ie global color correction~\cite{deepisp_tip} and ETGF~\cite{wu2018fast}, which we evaluate in Figure~\ref{fig:guided} and Table~\ref{tab:guided filter} and analyze as follows.

First, simply adopting the global color correction strategy from~\cite{deepisp_tip} can only recover holistic color appearances, such as in the background region of Figure~\ref{fig:guided}(c), but there are still significant distortions in local regions without the pixel-wise transformations of the proposed method.
Second, the ETGF~\cite{wu2018fast} method learns better guidance maps with CNNs (formulated in \eqref{eq:et_guided}) than the guided filter, and generates high-fidelity colors as shown in Figure~\ref{fig:guided}(e).
However, the ETGF method is still limited by the local constant assumption, and thus cannot adapt to high-frequency color changes as well as our learning-based guided filter.
In contrast, the learning-based guided filter learns pixel-wise transformation matrix for spatially-variant color corrections, and recovers high-fidelity color appearances without the local constant constraint (Figure~\ref{fig:guided}(g)). 
In addition, our method achieves significant quantitative improvement over the baseline models as shown in Table~\ref{tab:guided filter}.
Note that the proposed learning-based guided filter relies on the information of the first module for color transformation estimation. 
The network without feature fusion cannot predict accurate transformation matrix and tends to bring color artifacts around subtle structures in Figure~\ref{fig:guided}(f). 

\begin{table}[t]
	\caption{\label{tab:depth_sr} Evaluations of guided depth upsampling algorithms in terms of RMSE on the NYUv2 test set~\cite{NYUdataset}. The depth values are measured in centimeter similar to \cite{li2019joint}. Our method can achieve high-quality results for all the 4$\times$, 8$\times$, and 16$\times$ depth upsampling.
	}
	\vspace{-2mm}
	\footnotesize
	\centering
	\begin{tabular}{cccc}
\toprule
		Methods & 4$\times$ & 8$\times$ & 16$\times$ \\    %
\midrule
		GF~\cite{guidedFilter}& 7.32 & 13.62 & 22.03   \\
		JBU~\cite{kopf2007joint}& 4.07 & 8.29 & 13.35  \\
		TGV~\cite{ferstl2013image} & 6.98 & 11.23 & 28.13 \\
		Park~\cite{park2011high}& 5.21 & 9.56 & 18.10  \\
		Ham~\cite{ham2015robust} & 5.27 & 12.31 & 19.24  \\
		DJF~\cite{li2019joint}& 3.38 & 5.86 & 10.11 \\
		Ours & \bf 2.85 & \bf 5.49 & \bf 9.23 \\
\bottomrule
	\end{tabular}
	\vspace{-4mm}
\end{table}

\vspace{1ex}
{\noindent \bf{Guided depth upsampling.}} %
The proposed color correction network can also be used for other joint image filtering tasks, such as guided depth upsampling.
We apply our network to the depth upsampling problem in which the input is a low-resolution depth map paired with a corresponding high-resolution RGB image, which respectively correspond to $X_{ref}$ and $\tilde{X}_{lin}$ in \eqref{eq:our_guided}.
Since there is no image restoration network here, we do not use the feature fusion strategy and instead directly feed the depth map and RGB image into the color correction module to estimate the transformation matrix.
The predicted transformations are applied to the high-resolution RGB image to generate the desired high-resolution depth map as in \eqref{eq:our_guided}.
\rev{Intuitively, our method is in spirit similar to the guided filter~\cite{guidedFilter} which uses the high-resolution image to approximate the coarse depth such that the output is in the same modality as the input depth while preserving the sharp structures of the high-resolution image.}
More details about the network structure are presented on the project web page:  \url{https://sites.google.com/view/xiangyuxu/rawsr_pami}.
Similar to DJF~\cite{li2019joint}, we use the NYUv2 dataset~\cite{NYUdataset} for training and testing. 
Table~\ref{tab:depth_sr} shows that the proposed method performs favorably against the state-of-the-art guided depth upsampling approaches~\cite{guidedFilter,kopf2007joint,ferstl2013image,park2011high,ham2015robust,li2019joint}.
In addition, we present visual examples from the NYUv2 test set in Figure~\ref{fig:depth_sr}.
Overall, the proposed method can better exploit the structures of the RGB image by learning spatially-variant transformations and generate higher-quality depth maps.

\begin{figure}[t]
	\newcommand{\thisheight}{0.3\linewidth}
	\footnotesize
	\begin{center}
		\begin{tabular}{cc}
			\hspace{-4mm} 
			\includegraphics[height = \thisheight]{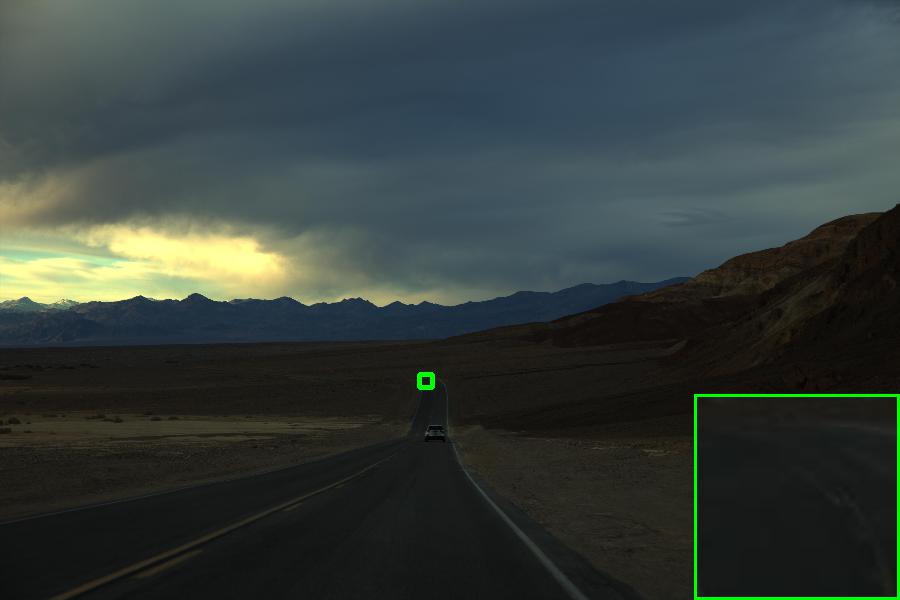} &
			\hspace{-2mm} 
			\includegraphics[height = \thisheight]{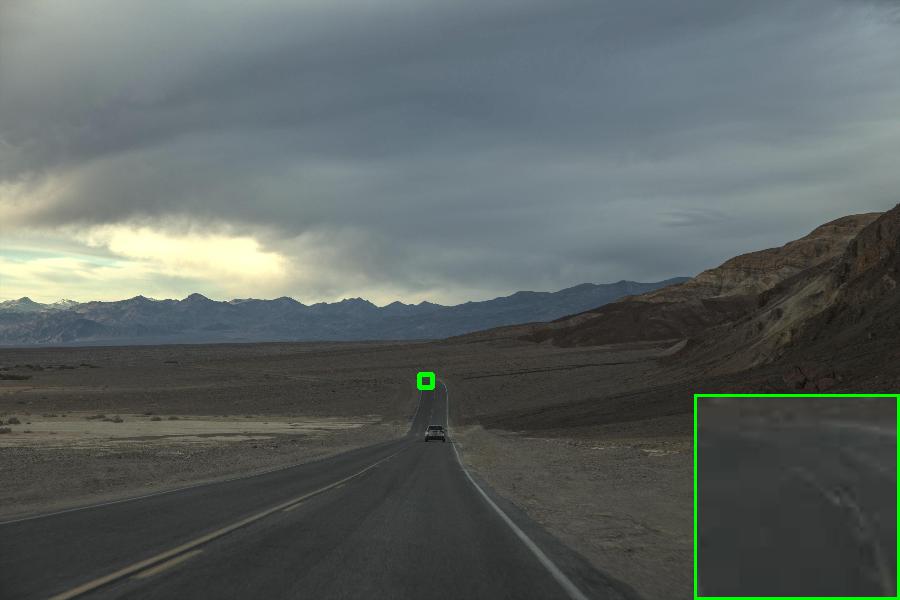} \\
			\hspace{-4mm} \scriptsize
			(a) Input processed by gamma correction &
			\hspace{-2mm} \scriptsize
			(b) Input processed by \cite{paris2011local} \\
			\hspace{-4mm} \scriptsize
			\includegraphics[height = \thisheight]{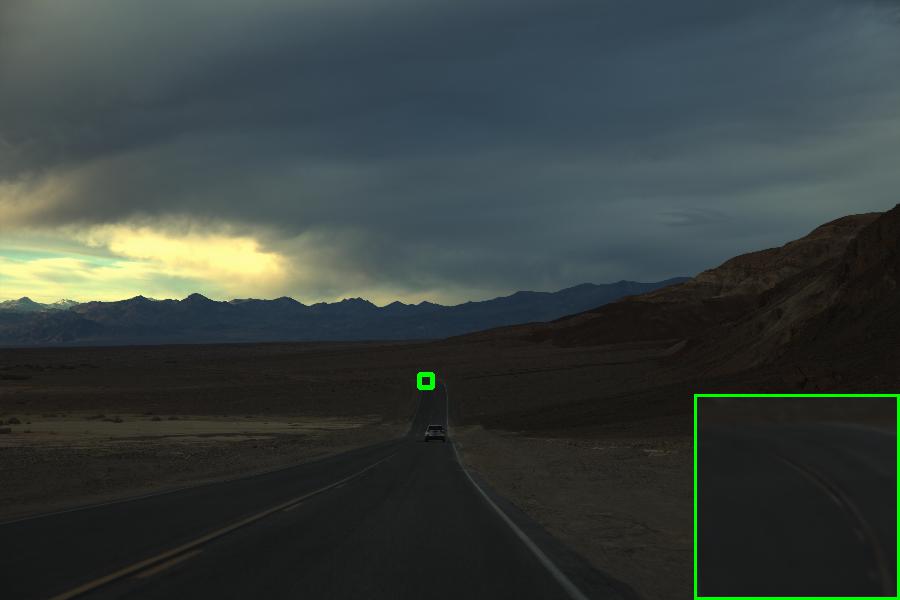} &
			\hspace{-2mm} \scriptsize
			\includegraphics[height = \thisheight]{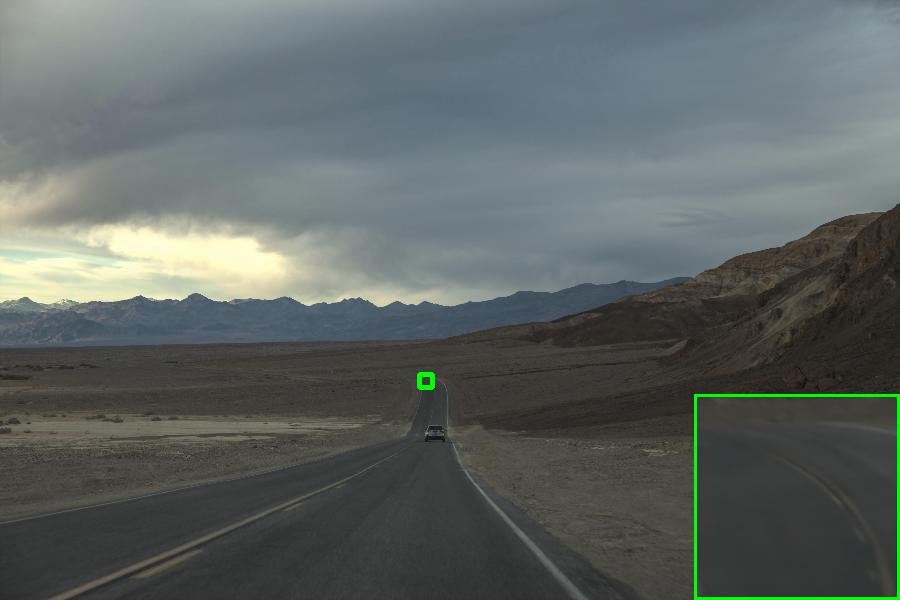} \\	
			\hspace{-4mm} \scriptsize
			(c) Super-resolution with (a) &
			\hspace{-2mm} \scriptsize
			(d) Super-resolution with (b) \\
		\end{tabular}
	\end{center}
	\vspace{-2mm}
	\caption{Generalization of the proposed method to different color correction operations.
		Our method can generate high-quality super-resolution results using the reference images produced by both gamma correction (a) and local Laplacian filter~\cite{paris2011local} (b). 
	}
	\label{fig:hdr}
	\vspace{-4mm}
\end{figure}

\begin{figure*}[t]
	\newcommand{\thisheight}{0.12\linewidth}
	\footnotesize
	\begin{center}
		\begin{tabular}{ccccc}
			\hspace{-4mm} 
			\includegraphics[height = \thisheight]{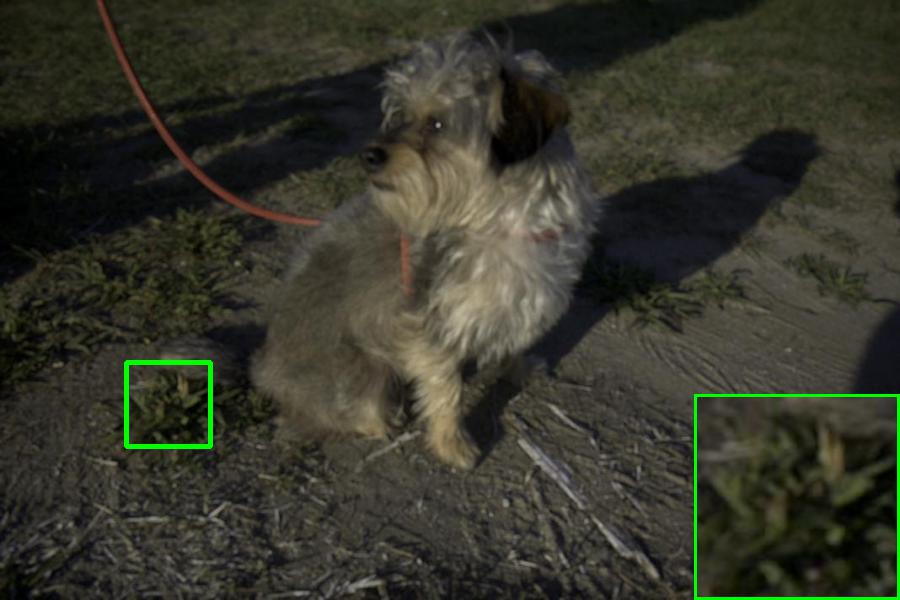} &
			\hspace{-4mm} 
			\includegraphics[height = \thisheight]{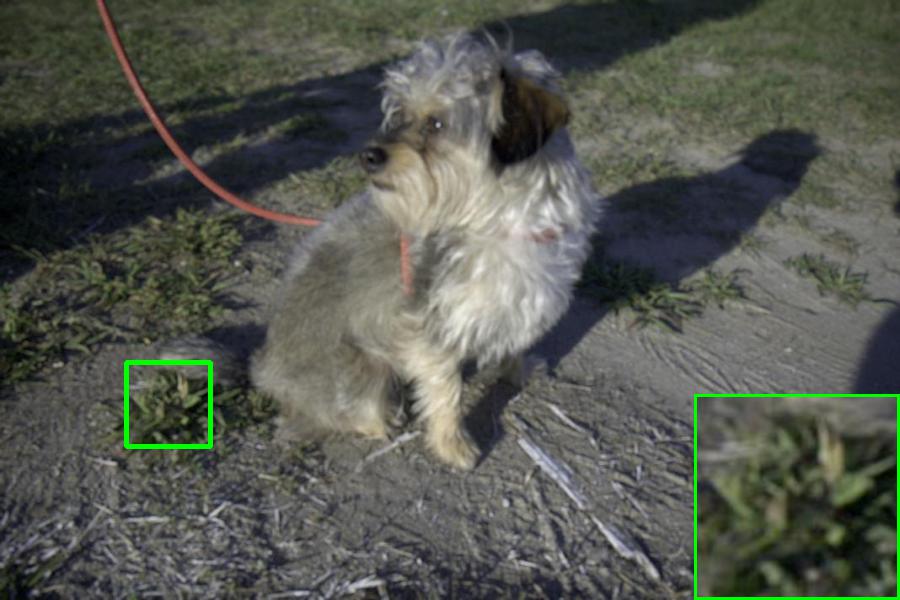} & 
			\hspace{-4mm} 
			\includegraphics[height = \thisheight]{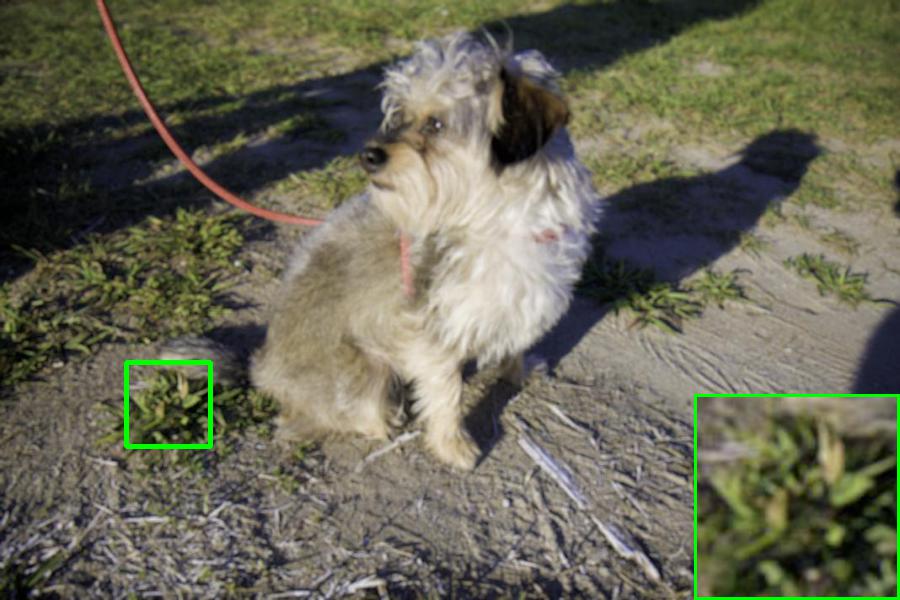} & 
			\hspace{-4mm}
			\includegraphics[height = \thisheight]{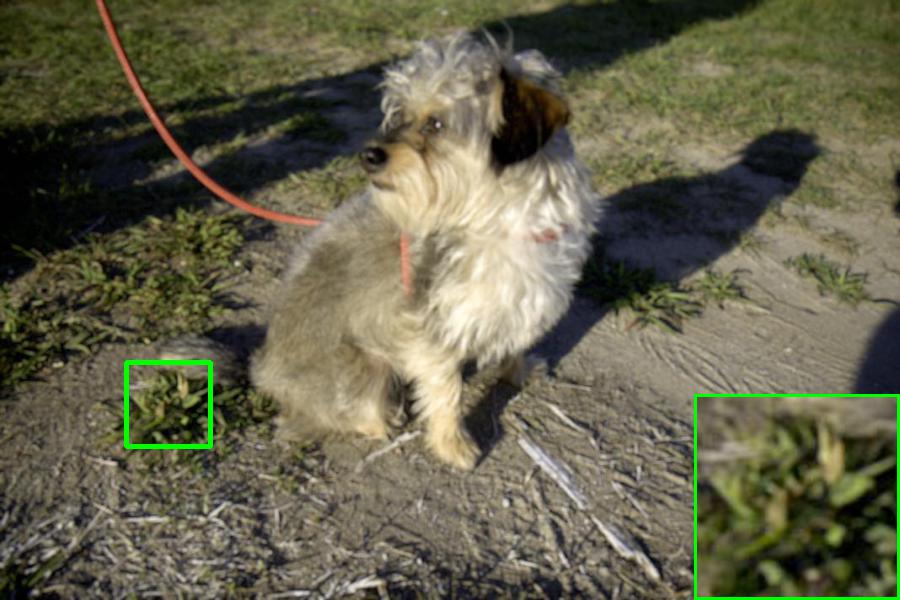} & 
			\hspace{-4mm}
			\includegraphics[height = \thisheight]{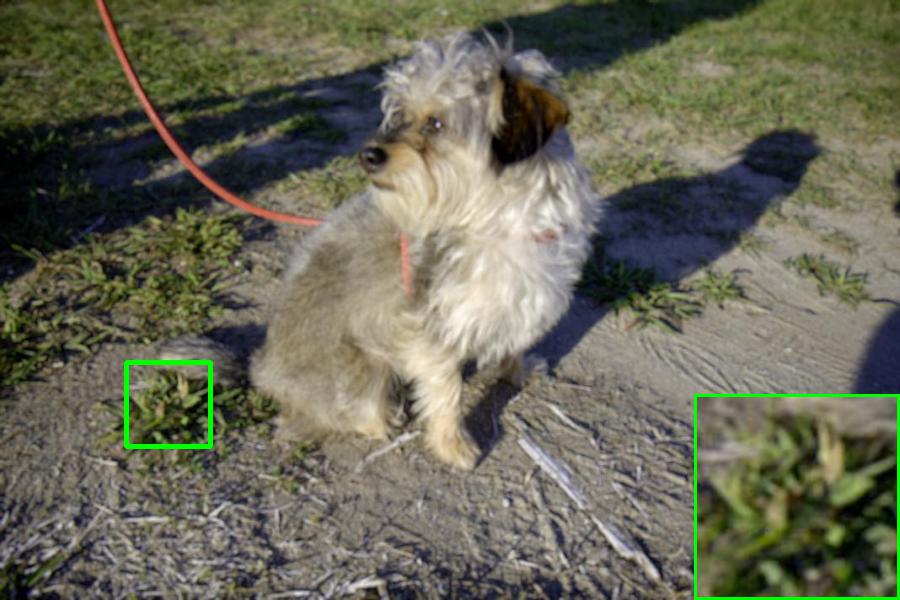} \\
			\hspace{-4mm} 
			Input 1 &
			\hspace{-4mm} 
			Input 2 &
			\hspace{-4mm} 
			Input 3 &
			\hspace{-4mm} 
			Input 4 &
			\hspace{-4mm} 
			Input 5 \\
			\hspace{-4mm}
			\includegraphics[height = \thisheight]{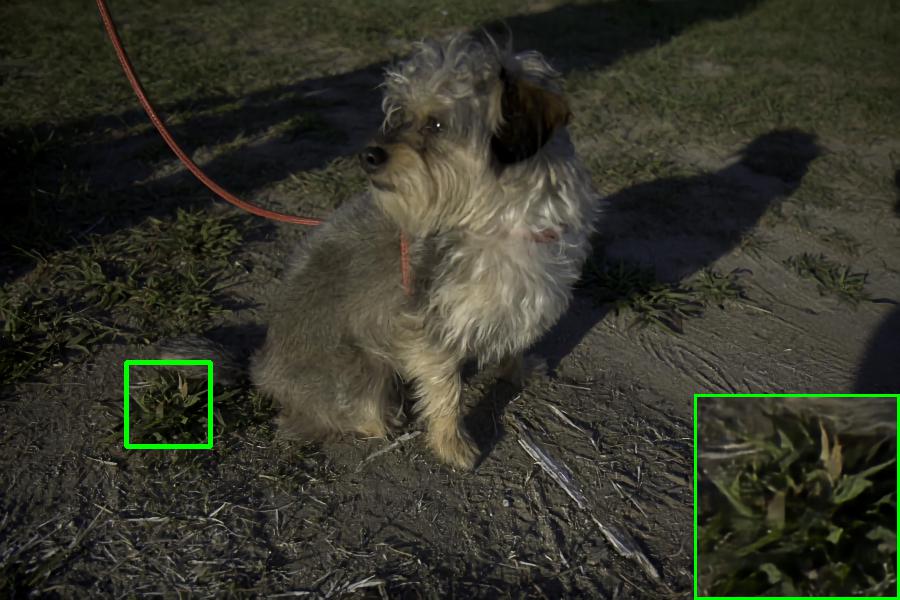} &
			\hspace{-4mm}
			\includegraphics[height = \thisheight]{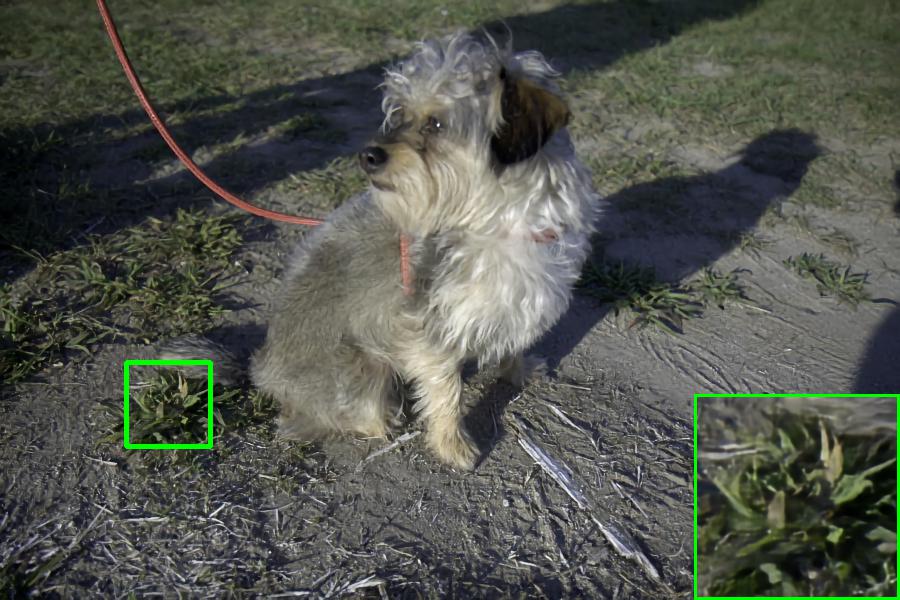} &	
			\hspace{-4mm}
			\includegraphics[height = \thisheight]{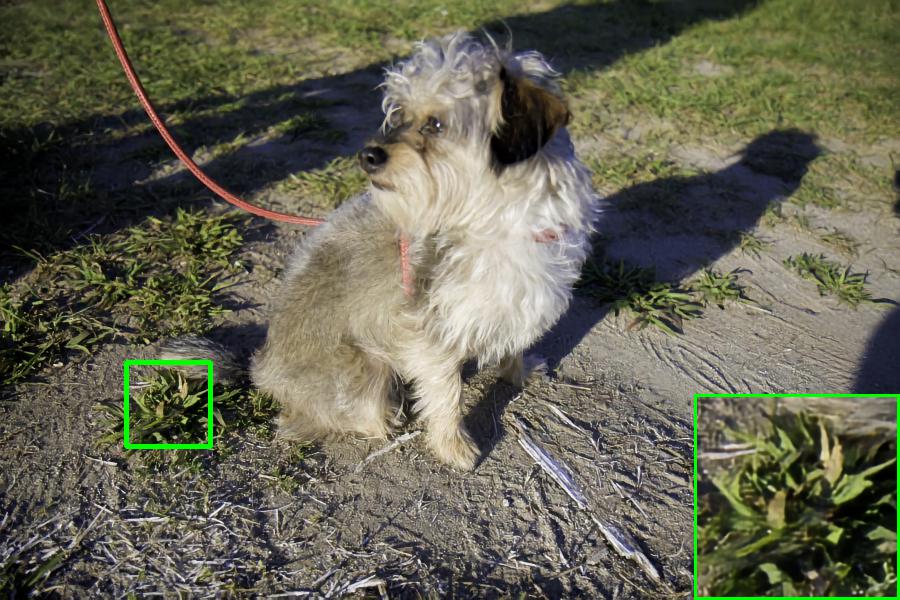} &
			\hspace{-4mm}
			\includegraphics[height = \thisheight]{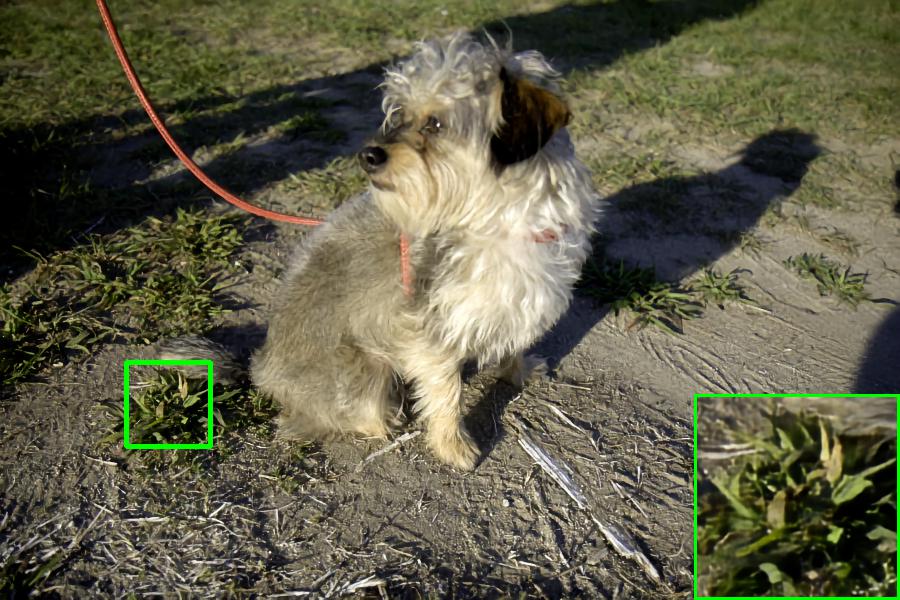} &
			\hspace{-4mm}
			\includegraphics[height = \thisheight]{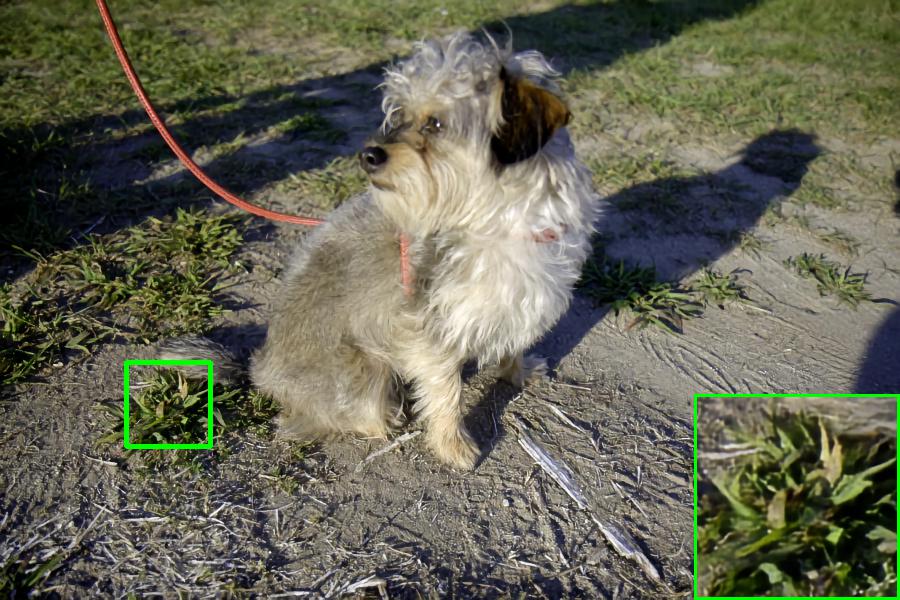} \\		
			\hspace{-4mm} 
			Output 1 &
			\hspace{-4mm} 
			Output 2 &
			\hspace{-4mm}
			Output 3 &
			\hspace{-4mm} 
			Output 4 &
			\hspace{-4mm} 
			Output 5 \\
		\end{tabular}
	\end{center}
	\vspace{-2mm}
\caption{Generalization of the proposed method to manually retouched color images from the MIT-Adobe 5K dataset~\cite{adobe-mit-5k}. The proposed method generates clear structures and high-fidelity colors.
	}
	\label{fig:retouching}
	\vspace{-4mm}
\end{figure*}

\begin{table}[t]
	\caption{\label{tab:compare_cvpr} Evaluations with our prior model \cite{rawSR} on the synthetic dataset. ``dense'' denotes the dense block. ``sFF'' and ``FF'' respectively represent the single-scale and multi-scale feature fusion. ``dense + sFF'' is the model of \cite{rawSR}, and ``DCA + FF'' is the full model of this work.
	}
	\vspace{-2mm}
	\footnotesize
	\centering
	\begin{tabular}{ccccc}
		\toprule
		\multicolumn{1}{c}{\multirow{2}*{Methods}}&\multicolumn{2}{c}{Blind}&\multicolumn{2}{c}{Non-blind}\\
		\cmidrule{2-5}
		& PSNR & SSIM & PSNR & SSIM \\    %
		\midrule
		dense + sFF & 30.79 & 0.8044 & 31.79 & 0.8272 \\
		DCA + sFF & 30.93 & 0.8068 & 31.97 & 0.8293  \\
		dense + FF & 30.90 & 0.8065 & 31.94 & 0.8294  \\
		DCA + FF & \bf 31.06 & \bf 0.8088 & \bf 32.07 & \bf 0.8314  \\
		\bottomrule
	\end{tabular}
	\vspace{-4mm}
\end{table}

\vspace{-2mm}
\subsection{Evaluation with our prior model}
There are also two main differences between the models in this work and \cite{rawSR}. 
First, we propose the DCA block (see Section~\ref{sec:effect_restoration}) which performs better than 
the dense blocks in \cite{rawSR}.
As shown in Figure~\ref{fig:compare_cvpr}(c) and (d) (the yellow arrow), the DCA blocks can reduce artifacts and generate clearer edges for image restoration.
Second, we apply the feature fusion for the color correction module in a multi-scale manner whereas \cite{rawSR} only fuses feature at one scale.
As shown in Figure~\ref{fig:compare_cvpr}(c) and (e) (the green arrow), the multi-scale feature fusion can generate more vivid colors closer to the ground-truth (Figure~\ref{fig:compare_cvpr}(g)) than single-scale feature fusion.
Finally, combining the above methods can further improve the performance qualitatively and quantitatively as shown in Figure~\ref{fig:compare_cvpr}(f) and Table~\ref{tab:compare_cvpr}.
In addition, we adapt the network structure for 4$\times$ super-resolution by changing the factor of the sub-pixel layer in the first module and adding a deconvolution layer to the second module.  
We also change the downsampling factor of the proposed data generation pipeline in \eqref{eq:data_generation} accordingly.
We train the proposed model with the same settings as described in Section~\ref{sec:implementation}. 
As shown in Table~\ref{tab:x4} and Figure~\ref{fig:x4}, our method can also generate high-quality results on 4$\times$ image upsampling both quantitatively and qualitatively.

\begin{table}[t]
	\caption{\label{tab:x4}Quantitative evaluations for 4$\times$ super-resolution on the synthetic dataset. ``Blind" represents the images with variable blur kernels, and ``Non-blind" denotes fixed kernel.
	}
	\vspace{-2mm}
	\footnotesize
	\centering
	\begin{tabular}{ccccc}
		\toprule
		\multicolumn{1}{c}{\multirow{2}*{Methods}}&\multicolumn{2}{c}{Blind}&\multicolumn{2}{c}{Non-blind}\\
		\cmidrule{2-5}
		& PSNR & SSIM & PSNR & SSIM \\    %
		\midrule
		SID~\cite{learning_to_see_in_the_dark} & 21.34 & 0.7043 & 21.35 & 0.6993  \\
		DeepISP~\cite{deepisp_tip}& 21.06 & 0.7015 & 21.22 & 0.7015  \\
		SRCNN~\cite{SRCNN}& 27.60 & 0.7293 & 28.29 & 0.7413  \\
		VDSR~\cite{SR-VDSR} & 27.98 & 0.7367 & 28.55 & 0.7500  \\
		SRDenseNet~\cite{SR-dense} & 28.20 & 0.7418 & 28.69 & 0.7536  \\
		RDN~\cite{SR-residual-dense}& 28.60 & 0.7481 & 29.08 & 0.7570  \\
		Ours & \bf 29.74 & \bf 0.7739 & \bf 30.27 & \bf 0.7871 \\
		\bottomrule
	\end{tabular}
	\vspace{-3mm}
\end{table}

\vspace{-2mm}
\subsection{Generalization to more complex color corrections}
While the ISP for our synthetic data is simulated with the Dcraw~\cite{dcraw} toolkit, the proposed super-resolution algorithm can be generalized to more complex color corrections. 
As one typical example, while the gamma correction is used for tone adjustment in the Dcraw toolkit, modern camera systems often use more complex algorithms~\cite{paris2011local} to deal with more challenging inputs, such as high dynamic range (HDR) images.
In Figure~\ref{fig:hdr}(a) and (b), we process the HDR image by gamma correction and the local Laplacian filter~\cite{paris2011local} separately.
We evaluate the proposed network with the same raw input but different reference images (see Figure~\ref{fig:hdr}(a) and (b)).
Despite the drastic color change between them, our super-resolution algorithm consistently generates clear results with high-fidelity color appearances as shown in Figure~\ref{fig:hdr}(c) and (d) without knowing the original ISP operations.

\vspace{1ex}
{\noindent \bf{Human retouched images.}} %
In addition to the HDR images,
we carry out experiments on images manually retouched by photographers from the MIT-Adobe 5K dataset~\cite{adobe-mit-5k} to simulate more diverse ISPs.
We evaluate our model on the same raw input with different reference images retouched by different photographers in Figure~\ref{fig:retouching}.
The proposed algorithm restores clear structures as well as high-fidelity colors, which shows that our method is able to generalize to more diverse and complex ISPs.

\begin{table}[t]
	\caption{\label{tab:time} Run-time of super-resolution algorithms on a 800$\times$600 image.
	}
	\vspace{-2mm}
	\footnotesize
	\centering
	\begin{tabular}{cccccc}
\toprule
		Method & DeepISP & VDSR & SRDenseNet & RDN & Ours \\
\midrule
		Time (s) & 0.2614 & 0.6892 & 0.9448 & 0.4987 & 0.647 \\
\bottomrule
	\end{tabular}
	\vspace{-1mm}
\end{table}

\begin{figure}[t]
	\newcommand{\thisheight}{0.28\linewidth}
	\footnotesize
	\begin{center}
		\begin{tabular}{c}
			\includegraphics[height = \thisheight]{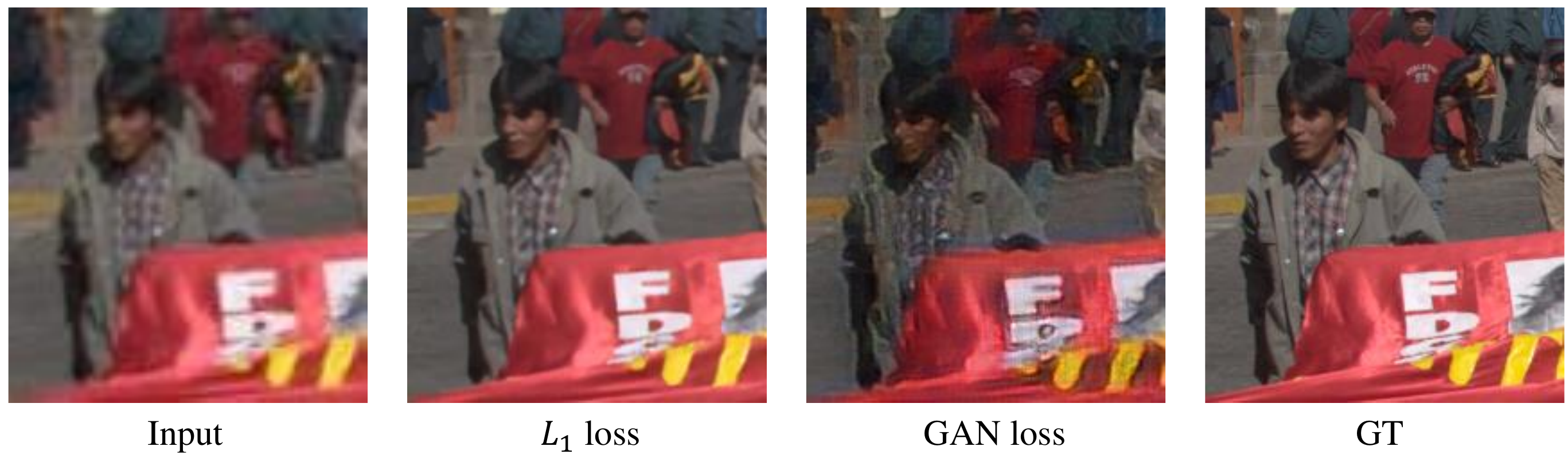} \\
		\end{tabular}
	\end{center}
	\vspace{-3mm}
	\caption{Training the super-resolution model with GAN loss.
	}
	\label{fig:GAN}
	\vspace{-4mm}
\end{figure}

\begin{figure}[t]
	\newcommand{\thisheight}{0.26\linewidth}
	\footnotesize
	\begin{center}
		\begin{tabular}{c}
			\includegraphics[height = \thisheight]{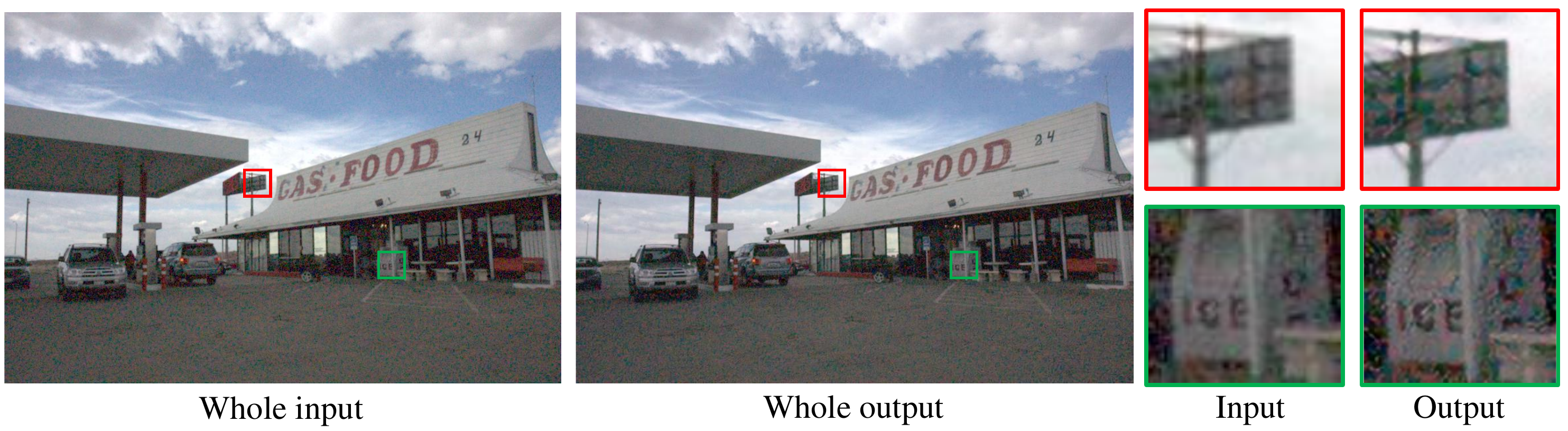} \\
		\end{tabular}
	\end{center}
	\vspace{-3mm}
	\caption{Failure case. The proposed method does not work well under severe noise.
	}
	\label{fig:failure}
	\vspace{-4mm}
\end{figure}

\vspace{-2mm}
\rev{\subsection{Adversarial training}
As introduced in Section~\ref{sec:implementation}, we use the $L_1$ loss for training the proposed network. 
A potential alternative is to use the adversarial training, for which we employ a combination of the perceptual loss~\cite{loss_perceptual} and GAN loss~\cite{gan} following the SRGAN~\cite{ganSR}. As shown in Figure~\ref{fig:GAN}, the generated image based on this training loss appears sharper at first glance.
However, upon close inspection, the generated image is of low quality with unpleasant artifacts and details not present in the ground truth.
Therefore, we do not use the GAN loss in our default settings.
}

\vspace{-2mm}
\subsection{Run-time}
We evaluate run-time of different super-resolution methods on the same machine with an Intel i7 CPU and a Nvidia GTX 1060 GPU.
Although there are two modules in the proposed model, our method has similar run-time as existing networks as shown in Table~\ref{tab:time}.
This is mainly due to the encoder-decoder formulation of the first module which processes the image features at a smaller scale, and the light-weight structure of the second module.

\vspace{-2mm}
\rev{\subsection{Practicality}
The proposed algorithm requires both the raw data and the processed color image as input. Most existing camera devices, such as SLRs and smartphones, can generate the color image in a considerably efficient way, which makes the proposed algorithm applicable for most users.
For the cases where the color input is not available, the users can process the raw data with their own ISP tools, \eg Dcraw~\cite{dcraw} and Photoshop, before applying the proposed super-resolution network. 
Further, it will be an interesting direction to explore the combination of the ISP system and the super-resolution pipeline for saving computation in future research.
}

\vspace{-2mm}
\rev{
\subsection{Limitations}
As introduced in Section~\ref{sec:implementation}, we use moderate noise parameters in our noise model~\eqref{eq:noise} to synthesize the training data.
When the input image has severe noise, the proposed method is likely to fail as shown in Figure~\ref{fig:failure}.
Our future work will focus on developing a network that explicitly considers noise reduction in the super-resolution process. 
}

\vspace{-2mm}
\section{Conclusions}

In this work, we propose a novel pipeline to generate realistic training data by simulating the real imaging process of digital cameras.
We develop a two-branch CNN to exploit the radiance information recorded in raw data.
The proposed algorithm performs favorably against the state-of-the-arts on the synthetic dataset and enables high-quality super-resolution results in real scenarios.
This work shows the effectiveness of learning with raw data, and we expect more applications of it in other related image processing
problems.
%
%
%
%
%
%
%
%
%
%
%
%
%
%
%
%
%

%
%

%
%
%

%

%

%
%
%
%
%
%
%
%
%
%
%
%
%
%
%
%
%
%
%
%
%
%
%
%
%
%
%
%
%
%
%
%
%
%
%
%
%
%
%
%
%
%
%
%
%
%
%
%
%
%
%
%
%
%
%
%
%
%
%
%
%
%
%
%
%
%
%
%
%
%
%
%
%
%
%
%
%
%
%
%
%
%
%
%
%
%
%
%

%
%
%
%
%
%
%
%
%
%
%
%
%
%
%
%
%
%
%
%
%

%

%

%
%

%
%
%
%
%
%

%
%
%
%
%
%

%
%
%
%
%
%
%
%
%
%
%
%\ifCLASSOPTIONcompsoc
%  %
%  \section*{Acknowledgments}
%  {
%  	Ming-Hsuan Yang is supported in part by the NSF CAREER Grant $\#1149783$.
%  }
%\else
%  %
%  \section*{Acknowledgment}
%\fi

%
%
\ifCLASSOPTIONcaptionsoff
  \newpage
\fi

\vspace{-2mm}
\bibliographystyle{IEEEtran}
\bibliography{rawSR}

\end{document}